\documentclass[journal]{IEEEtran}
\usepackage{epsfig,subfigure,graphics,float,amsmath}
\usepackage{cite}

\begin{document}

\title{Electric-field-coupled oscillators for collective electrochemical perception in underwater robotics}
\author{Serge Kernbach\\[3mm]
\small CYBRES GmbH, Research Center of Advanced Robotics and Environmental Science, \\
\small Melunerstr. 40, 70569 Stuttgart, Germany, \emph{serge.kernbach@cybertronica.de.com}
\vspace{-5mm}
}

\date{}
\maketitle

\begin{abstract}
This work explores the application of nonlinear oscillators coupled by electric field in water for collective tasks in underwater robotics. Such coupled oscillators operate in clear and colloidal (mud, bottom silt) water and represent a collective electrochemical sensor that is sensitive to global environmental parameters, geometry of common electric field and spatial dynamics of autonomous underwater vehicles (AUVs). Implemented in hardware and software, this approach can be used to create global awareness in the group of robots, which possess limited sensing and communication capabilities. Using oscillators from different AUVs enables extending the range limitations related to electric dipole of a single AUV. Applications of this technique are demonstrated for detecting the number of AUVs, distances between them, perception of dielectric objects, synchronization of behavior and discrimination between 'collective self' and 'collective non-self' through an 'electrical mirror'. These approaches have been implemented in several research projects with AUVs in fresh and salt water.
\end{abstract}


\section{Introduction}
\label{sec:intro}

Underwater exploration became recently an important economic, ecologic and social concept. Possible tasks for autonomous underwater vehicles (AUVs) include monitoring ecologically sensitive areas, exploration of the seabed, safety inspections, and finding objects of interest~\cite{TinyFish10}. Challenges of underwater robotics are related, among others, to a high damping factor of water, which limits sensing and communication capabilities of AUVs~\cite{KernbachDipperSutantyo11}.

To extend these limitations, the state of the art approach is to use multiple AUVs~\cite{Bingham02}. These are small robot platforms, designed to operate in collective way. This approach has several advantages such as exploration of large areas, increased functional operability through heterogeneity, enhanced reliability and a lower cost of platforms. Research and technological developments concern collective cognitive capabilities,  coordination of the whole robot group, platform design and other issues~\cite{Kernbach11-HCR}.

Interesting bio-inspired solution observed in weakly electric fish~\cite{Emde98}, \cite{Sim01062011} consists in using underwater electric fields for collective sensing and communication. This approach implemented for different purposes in potential and current modes \cite{Boyer15}, {\cite{Chevallereau14}, \cite{Shang20} demonstrated strong and weak sides of such technology. For example, some difficulties are related to electric dipole of a single AUV limited by the body length of AUV \cite{Baffet08}. Due to small dipole size, the effective range of electric-field-based interactions is also limited. Improvement of this technology can be achieved by using multiple oscillating dipoles coupled by electric field. Such coupled oscillators are sensitive to spatial distribution of dipoles, their orientation and presence of obstacles in a common field. Variation of these parameters influences common synchronization patterns, amplitude/phase of signals. Individual robots, despite limited sensing capabilities, can become aware of a global environmental state by measuring parameters of local oscillations.

Specific frequency-current dynamics of oscillating dipoles can be used in a manner of electrochemical impedance spectroscopy \cite{Kernbach17water} with different oscillating frequencies. Such collective electrochemical sensor is used for environmental sensing, detection of conducting/dielectric objects and sensing in colloidal (mud, bottom silt) solutions. Different interference patterns enable identification of AUVs and underwater beacons.

In this paper, nonlinear effects of electric-field-coupled oscillators and their applicability to AUVs \cite{ANGELS}, \cite{Thenius16subCULT} are explored. It is demonstrated that changes in amplitude, phase and temporal patterns of signals can be attributed to various group-internal/-external events. Global synchronization effects can also be used for behavioral coordination. In particular, we explore the capability to discriminate between collectively self- and nonself- generated signals by a so-called 'electrical mirror' \cite{CoCoRo}. Similarly to an optical mirror, the electrical mirror reflects received electrical signals and allows a robot to receive its own electrical reflection. The correlation between emitted and received signals enables robots to determine group affiliation or to discriminate between 'collective self' and 'collective non-self'. Electric fish recognize self-/nonself-generated signals in a similar way~\cite{Caputi11}. These experiments represent one of developmental lines for designing collective cognitive capabilities in a group of AUVs, which were implemented in several research projects with underwater robots \cite{AquaJelly}, \cite{CoCoRo}, \cite{ANGELS}, \cite{Thenius16subCULT} operating in fresh and salt water.

\section{Oscillators with Embodied Coupling}
\label{sec:coordination}

Coupled oscillators are well known in theoretical physics \cite{Atmanspachera05} and nonlinear control~\cite{Konishi99}, where different spatio-temporal effects are observed \cite{Chate92}. Both low-dimensional \cite{Maistrenko98}, \cite{Levi99} and high-dimensional coupled map lattice (CML) \cite{Kaneko93} have applications in secure communication and functional control. The CML approach represents a useful modeling tool for studying the spatio-temporal dynamics of complex nonlinear systems. Examples are the concepts of algorithmic and analytic agents~\cite{Kernbach08}, \cite{Kornienko_S04} decision-making processes \cite{Kornienko_OS01}, \cite{Kornienko_S03A} and homeostatic regulation \cite{Kornienko_S06b} on a collective level. Due to disembodiment of state variables and couplings, the dynamics of CMLs can be investigated analytically as well as numerically. However, such a disembodiment imposes several limitations on applications of this technique in robotics \cite{Christensen09}.

Another approach for coupled oscillators is represented by central pattern generators (CPG)~\cite{Yuste:2005p45387}, which are investigated e.g. in neuroscience, biology \cite{hamann2017flora} and in locomotive control~\cite{Endo08}. The state variables of CPG are embodied (for example the amplitude of a biped movement); the disembodied 'virtual channels' are used for information exchange between oscillators. Due to embodiment, CPG-based coupled oscillators are widely used in robot controllers \cite{Meister11}, \cite{kernbach09-2adaptive-short}. The CPG approach represents a balance between using embodied processes as state variables and the possibility of analytical treatments for complex biological or technological processes.

This work proposes an approach that differs from the CML and CPG-based methodologies. The idea is to involve physical embodiment for coupling instead of state variables. Such an approach is more natural for operations in liquid media with a high damping factor. For instance, interferences between electric fields emitted by oscillating dipoles can represent an embodied coupling mechanism, see Fig.~\ref{fig:DelayedMirror1}. This coupling depends on parameters of water (e.g. salinity, geometry of sensing area, presence of obstacles), the frequency of individual oscillations, the number of dipoles and distances between them, relatives positions and orientations of dipoles in three-dimensional space. All these parameters reflect the global state of the system. The perception of this information is strongly local. Analyzing perturbations of oscillations \cite{Nayfeh93}, each robot can become aware of the global situation and can undertake corresponding activities.

\begin{figure}[ht]
\centering
\subfigure[\label{fig:DelayedMirror1}]{\includegraphics[width=0.245\textwidth]{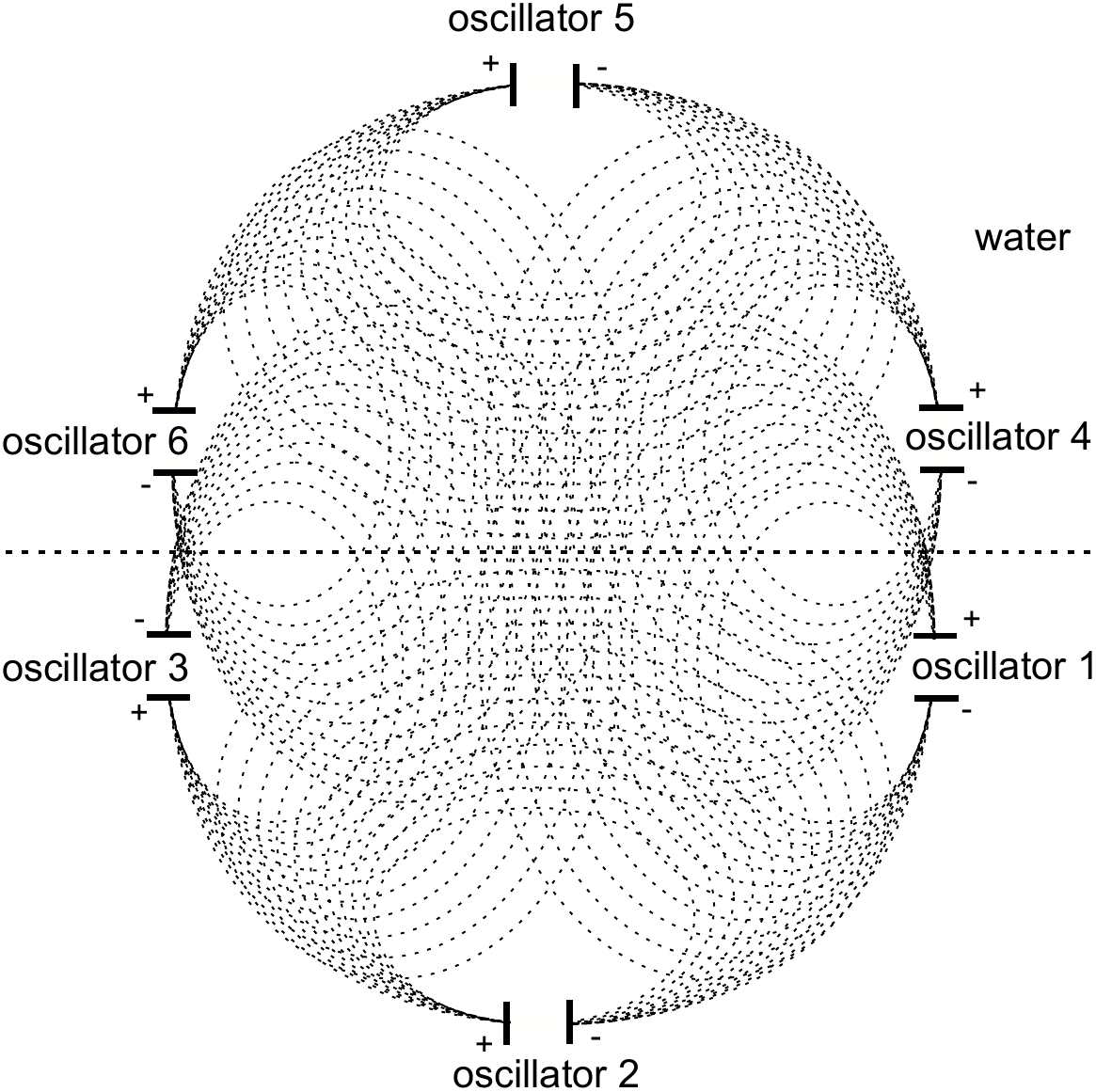}}~
\subfigure[\label{fig:DelayedMirror2}]{\includegraphics[width=0.245\textwidth]{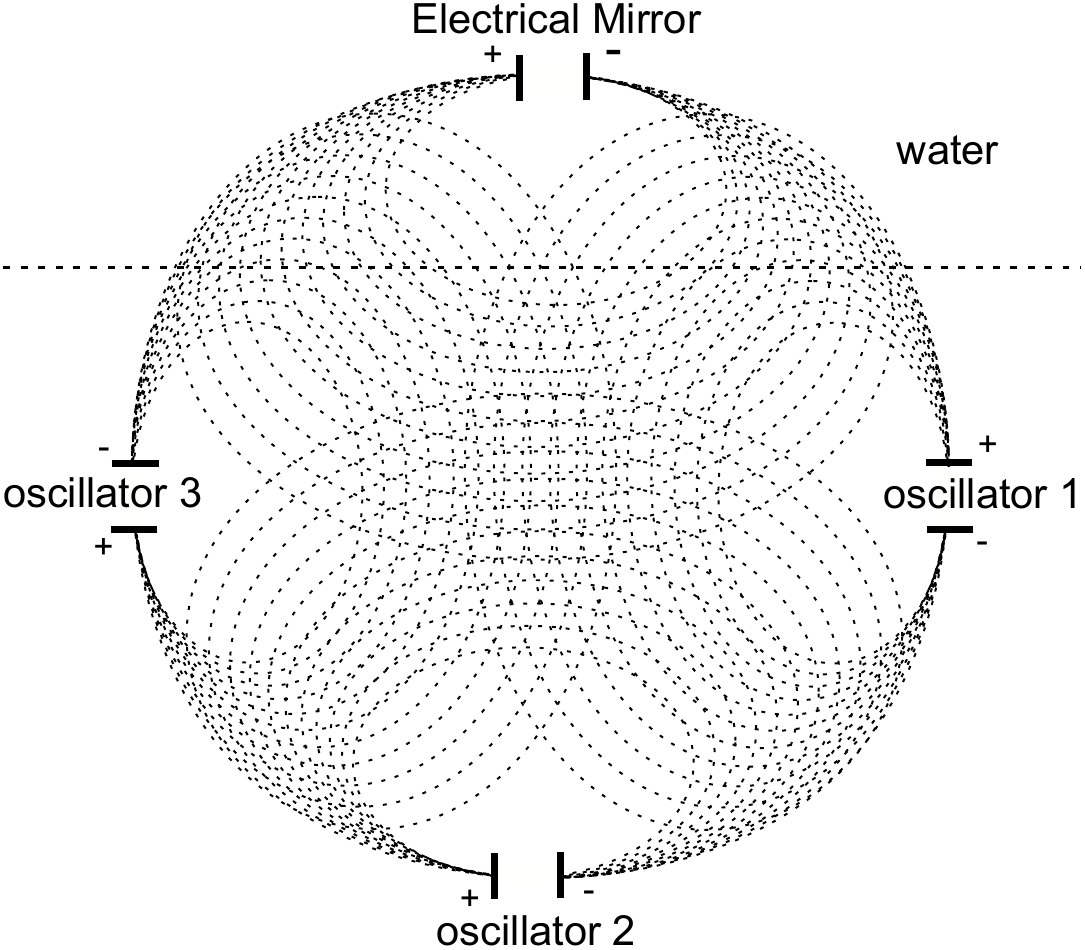}}
\caption{\small Physical embodiment of coupled oscillators by interferences of electric fields in water in current mode \textbf{(a)} without electrical mirror and \textbf{(b)} with electrical mirror. \label{fig:DelayedMirror}}
\end{figure}

Structures of oscillating devices are shown in Fig.~\ref{fig:structureDevices}. For modelling we use time-discrete dynamical systems~\cite{Sandefur90}, \cite{Levi99}, because they are suitable for implementation in a microcontroller. The signal from the oscillator ${\xi_n^o}$ at each time step $n$ is converted into an analog signal by the DAC, which is connected to the positive- $E^+$ and negative- $E^-$ emitting electrodes. In potential mode, the positive- $R^+$ and negative- $R^-$ receiving electrodes are connected to a high-pass filter, amplifier and ADC and represent the signal ${\xi_n^w}$ from water at the time step $n$, see Fig.~\ref{fig:ExperimentSchemeSingle}. In the current mode, a single pair of electrodes is used to produce the potential and to sense a current between electrodes, see Fig. \ref{fig:CurrentMode}. Analytical considerations, represented below, can be applied to both modes of operations.

\begin{figure}[ht]
\centering
\subfigure[\label{fig:ExperimentSchemeSingle}]{\includegraphics[width=0.45\textwidth]{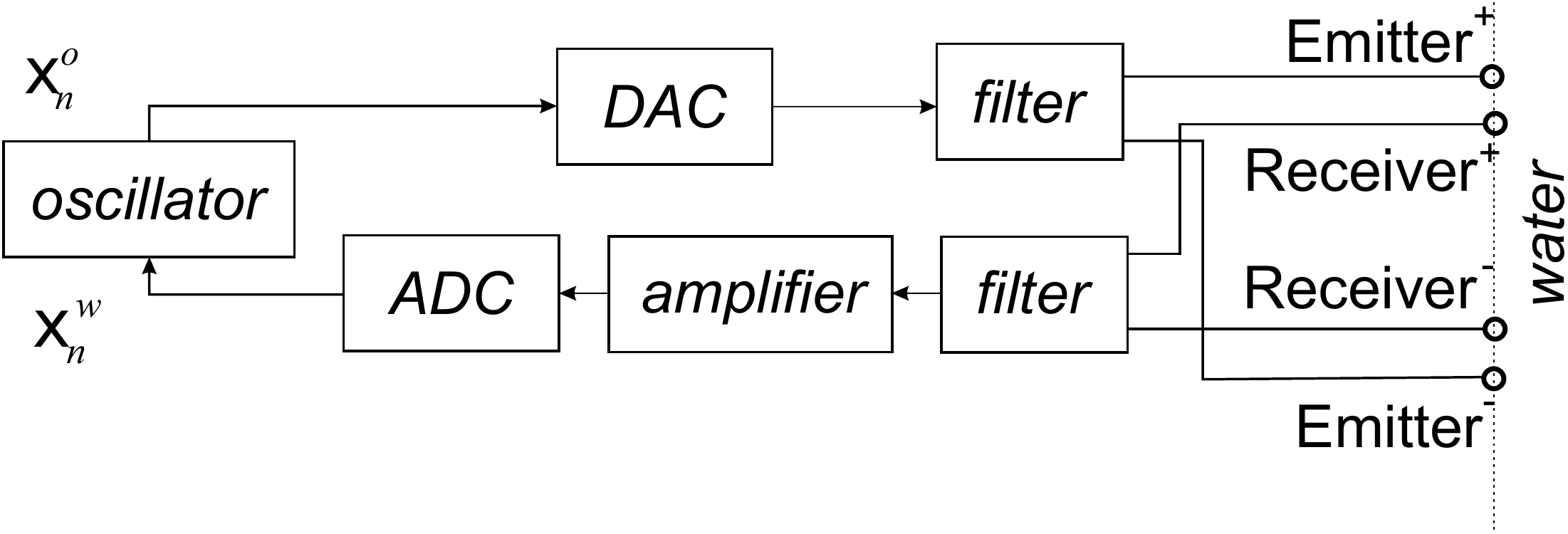}}
\subfigure[\label{fig:CurrentMode}]{\includegraphics[width=0.45\textwidth]{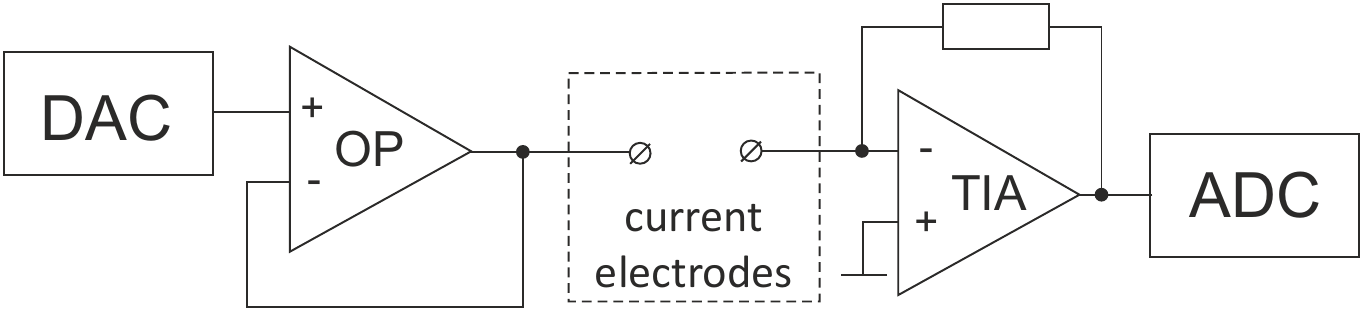}}
\caption{\small Structure of oscillating devices in \textbf{(a)} potential mode and \textbf{(b)} current mode, notations: DAC (digital-to-analog converter), ADC (analog-to-digital converter), OP (operational amplifier), TIA (transimpedance amplifier).
\label{fig:structureDevices}}
\end{figure}

Let $x_n$ be a state variable and $f(x_n, \alpha)$ -- a function of $x_n$ and the control parameter $\alpha$. The dynamical system can be written in the form
\begin{eqnarray}
{\xi_n^o} &=& f(x_n, \alpha), \\
\label{eq:gen2}
x_{n+1} &=& f(x_n, \alpha) + k_2{\xi_n^w},
\end{eqnarray}
where $k_2$ is a coefficient. The variable ${\xi_n^w}$ contains own signal sent by the emitting electrodes as well as signals from other emitting devices. Since own signal has an essential impact on the dynamics of the system (\ref{eq:gen2}), we can follow two strategies to remove it from $\xi_n^w$. First, we can set $k_2 =0$ and investigate the dynamics of $\xi_n^w$ without modifying the dynamics of $x_n$. Since the coupling is embodied, the dynamics of $\xi_n^w$ can provide enough information about the global state of the system. In the second strategy, the signal is subtracted from the equation ~(\ref{eq:gen2}). The term $(\xi_n^w - k_1 {\xi_n^o})$ represents the coupling part from other oscillating devices, where $k_1$ is the coefficient reflecting the damping factor of water as well as the amplification in electronics. The following condition
\begin{equation}
\label{eq:balancing condition}
\xi_n^w - k_1 \xi_n^o=0
\end{equation}
should be satisfied when no other oscillating devices are in the water. This is achieved by calibrating the device and setting the coefficient $k_1$. The following system
\begin{eqnarray}
\xi_n^o &=& f(x_n, \alpha), \\
\label{eq:gen3}
x_{n+1} &=& f(x_n, \alpha) + k_2 (\xi_n^w - k_1 \xi_n^o)
\end{eqnarray}
represents the model of an oscillating device. In the following we use only Eq.~(\ref{eq:gen3}), bearing in mind $\xi_n^o=f(x_n, \alpha)$.

In further treatment of $f(x_n, \alpha)$ we involve the well-known logistic map $x_{n+1} =  \alpha x_n (1-x_n)$, where $x \in R$ is the state variable and $\alpha$ is the control parameter. Since the input from water does not include a constant potential (which is removed by a high-pass filter), the logistic map must also be transformed to remove such a potential. This is achieved by eliminating the non-periodic stationary states. Solving $x_{n} =  \alpha x_n (1-x_n)$, we receive $x_{st_{1,2}}=\{0, \frac{\alpha-1}{\alpha}\}$. From the linear stability analysis we know that $x_{st_{1}}=0$ is stable in the region $\alpha=(-1..1)$ and $x_{st_{2}}=\frac{\alpha-1}{\alpha}$ is stable in the region $\alpha=(1..3)$. By inserting a new variable $y_n=x_n+\frac{\alpha-1}{\alpha}$ and $y_{n+1}=x_{n+1}+\frac{\alpha-1}{\alpha}$ and rewriting the systems in terms of the variables $x_n$ and $x_{n+1}$, we obtain
\begin{equation}
\label{eq:transformedSystem}
x_{n+1} =  x_n (2-\alpha x_n - \alpha).
\end{equation}
The stationary state $x_{st_{1}}=-\frac{\alpha-1}{\alpha}$ is stable in the region $\alpha=(-1..1)$ and $x_{st_{2}}=0$ in the region $\alpha=(1..3)$.

For further tests in Sec.~\ref{sec:setup} we use Eq.(\ref{eq:transformedSystem}) in the form of Eq.(\ref{eq:gen3}); thus
\begin{equation}
\label{eq:coupledSystem}
x_{n+1} =  x_n (2-\alpha x_n - \alpha) + k_2 (\xi_n^w - k_1 \xi_n^o)
\end{equation}
is used in the microcontroller for generating oscillations. Coupling through an electric field in water for low frequencies can be approximated by $\sum_{j=1}^m g^j  x_n^{j}$, where $g$ represents the distance from $j-$device to a mutual center of the system~\cite{Alamir10}. For the case shown in Fig.~\ref{fig:DelayedMirror1}, we assume that the coefficient $k_1$ is calibrated so that its signal is filtered from the received signal. Since potential fields have properties of superposition, the system (\ref{eq:coupledSystem}) for several devices can be represented as
\begin{equation}
x_{n+1}^i = x_n^i (2-\alpha x_n^i - \alpha) + \sum_{j=1}^m g^j  x_n^{j},~~~~~i=1,...m,
\label{eq:coupl121}
\end{equation}
where $m$ is a dimension of the system (\ref{eq:coupl121}) and the coefficients $g^j$ reflect factors such as amplifications of $\xi_n^o$, losses in the high-pass filters, water damping and spatial conditions between emitting devices. Since only periodic signals are taken into consideration (see Sec.~\ref{sec:setup}), we can use only a period-two motion of (\ref{eq:coupl121}) and corresponding global synchronization effects as reported in \cite{Wu98}, \cite{Lu07}.

\subsection{Coupled Oscillators Mode with $k_2=0$}
\label{sec:linCoupledModeZero}

The coupled oscillators mode corresponds to the case in which each of devices implements Eq.(\ref{eq:coupledSystem}), emits $\uparrow{\xi^i_n}$ and receives $\downarrow{\xi^i_n}$ from the water. Since coupling is performed using an electric field with additive properties, the term $\downarrow{\xi^i_n}$ contains all signals from other oscillators. In the case of $k_2=0$, the term $\downarrow{\xi^i_n}$ does not influence the dynamics of Eq.(\ref{eq:coupledSystem}). The dynamics of $\downarrow{\xi^i_n}$ is determined by two factors: different $g^j$ caused by the distance between devices and phase desynchronization of oscillators, see more in Sec.~\ref{sec:desynchronization}.

To exemplify the effect of phase desynchronization, in Fig.~\ref{fig:TeorDesynch1} we plot the behavior of two equations (\ref{eq:coupledSystem}), which have slightly different resolutions of step $n$ (this is performed in numerical simulation) and, in Fig.~\ref{fig:TeorDesynch2}, the term $\sum_{j=1}^m g_2^j  x_n^{j}$. The resolution of step $n$ in Fig.~\ref{fig:TeorDesynch2} is ten times higher than in Fig.~\ref{fig:TeorDesynch1}, i.e., one step $n$ is represented by 10 points.
\begin{figure}[ht]
\centering
\subfigure[\label{fig:TeorDesynch1}]{\includegraphics[width=0.245\textwidth]{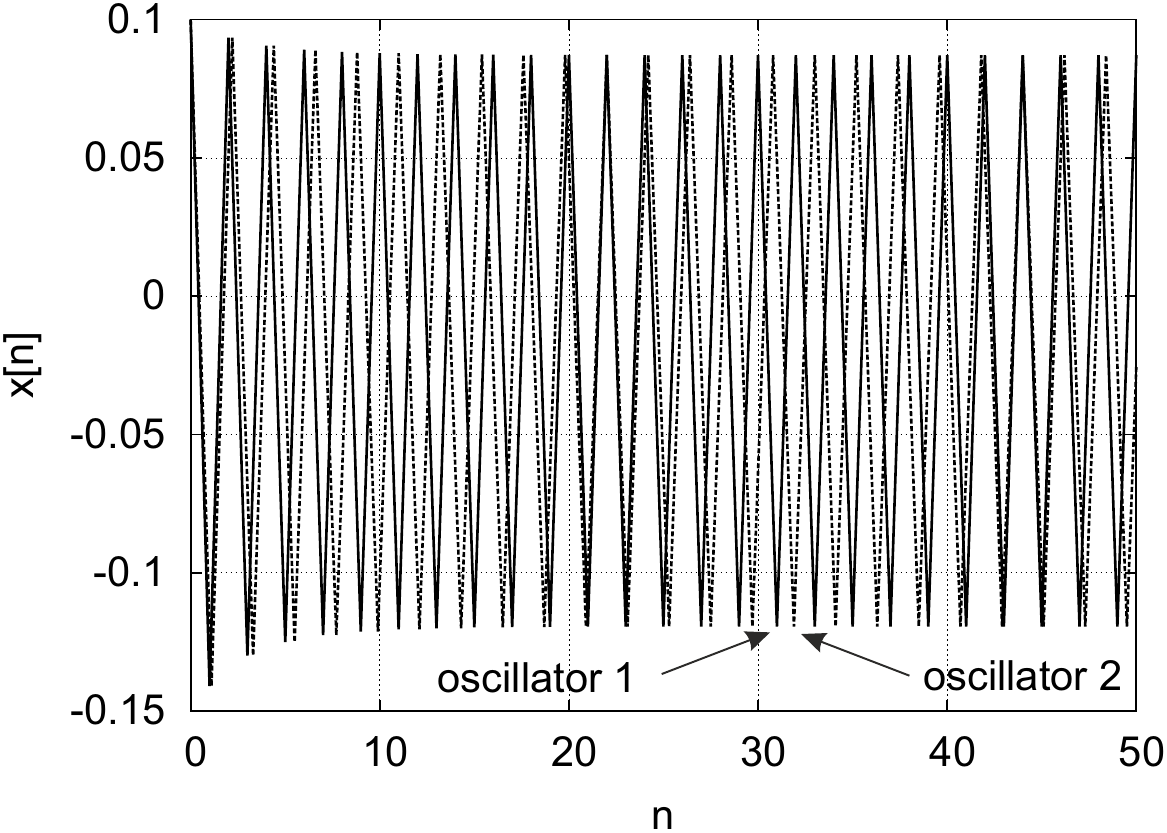}}~
\subfigure[\label{fig:TeorDesynch2}]{\includegraphics[width=0.245\textwidth]{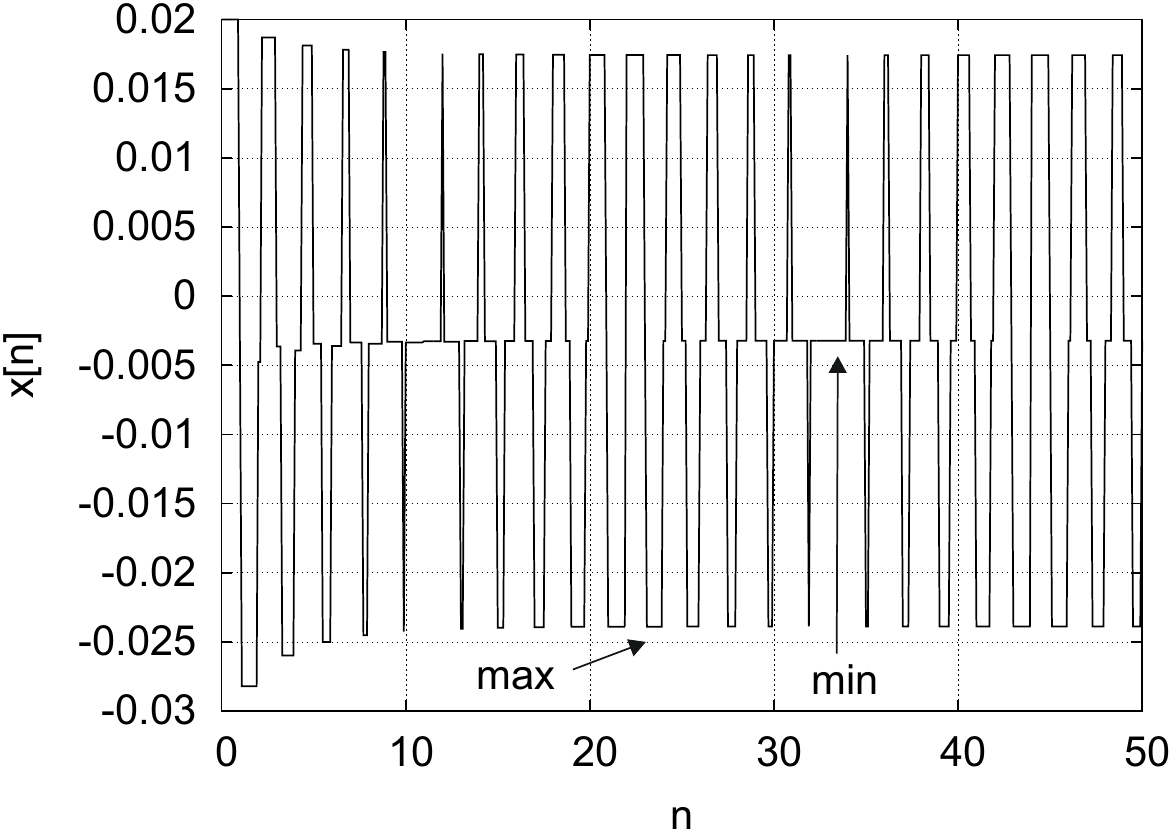}}
\caption{\small \textbf{(a)} Behavior of phase-desynchronized Eq.~(\ref{eq:coupledSystem}), $\alpha=3.1$, $k_2=0$, initial condition 0.1 for both equations; \textbf{(b)} The term $\sum_{j=1}^m g^j  x_n^{j}$, which corresponds to the value $\downarrow{\xi^i_n}$ from the water, $g^1=g^2=0.1$ (resolution of step $n$ is ten times higher). \label{fig:TeorDesynch}}
\end{figure}
We can see that the maximum and minimum values are generated by $x_n (g^1+g^2)$ and $x_n (g^1-g^2)$. Since $g^j$ is primarily related to distances between oscillators, the maximum value of the signal provides information about  $\sum_{j=1}^m g^j$, that is, the sum of all distances between devices.

\subsection{Coupled Oscillators Mode with $k_2\neq0$}
\label{sec:linCoupledMode}

For the case $k_2\neq0$ it is assumed that $k_1$ is calibrated so that the influence of own variables in $\downarrow{\xi^i_n}$ is neglected. Thus, the system (\ref{eq:coupl121}) has only one-way coupling with another variable, denoted as $x_n^{i+1}$. To summarize the changes caused by $g^j$ and by $k_2$, we consider only the coefficient $k_2$. For simplification, it is assumed that the distance between devices is equal. Nonlinear effects related to asynchronous updates~\cite{mehta:350} and desynchronization are neglected. To analyze the periodical behavior, we consider the second-iteration of the map, i.e. $f(f(x))$, which takes the following form for the system (\ref{eq:coupl121}) with $m=2$: 
\begin{eqnarray}
x_{n+1}^i &=& (x_n^i (2 - \alpha x_n^i - \alpha ) + k_2 x_n^{i+1}) (2- \alpha (x_n^i(2 -  \alpha x_n^i \nonumber\\&&
- \alpha ) + k_2 y_n^{i+1}) - \alpha) + k_2 (x_n^{i+1} (2 - \alpha x_n^{i+1} - \alpha) \nonumber\\&&+ k_2 x_n^i),
\label{eq:couplSecond}
\end{eqnarray}
considering boundary conditions for $x_n^{i+1}$. This is a $4^{th}$ order system with 16 stationary states $x_{st}^i$. Here, period-two stationary states from (\ref{eq:coupl121}) become non-periodic and can be analyzed. The Jacobian of the system (\ref{eq:couplSecond}) has two eigenvalues $\lambda_{1,2}$. From the stability conditions $|\lambda_{1,2}|\leq1$ evaluated on each $x_{st}, y_{st}$, we can estimate which stationary states are stable. This step is done by combining analytic and numeric approaches. The stable $x_{st}\in R$ (without complex conjugated $x_{st}$) are shown in Fig.~\ref{fig:eigenVCoupl3} and corresponding eigenvalues in Fig.~\ref{fig:eigenVCoupl2}.
\begin{figure}[ht]
\centering
\subfigure[\label{fig:eigenVCoupl2}]{\includegraphics[width=0.245\textwidth]{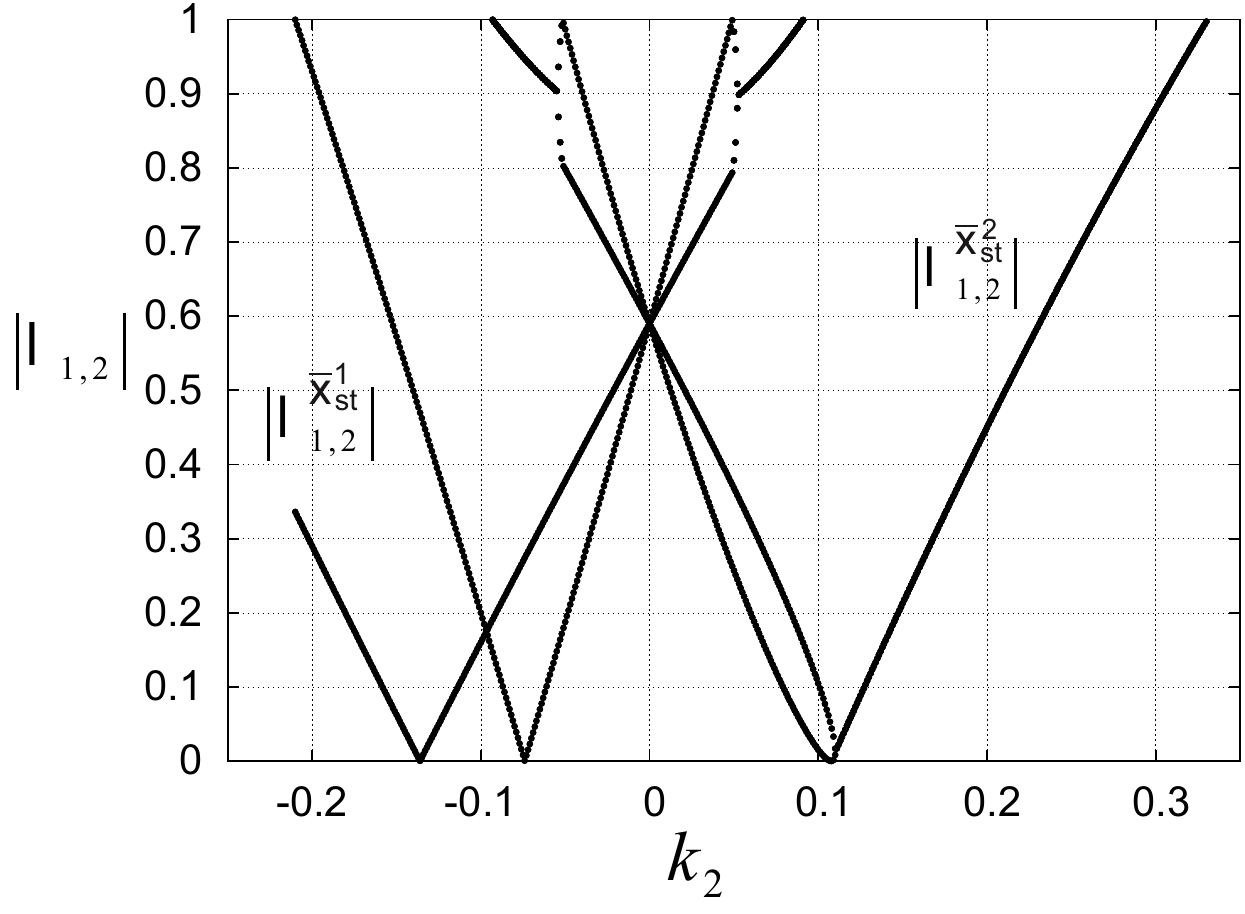}}~
\subfigure[\label{fig:eigenVCoupl3}]{\includegraphics[width=0.245\textwidth]{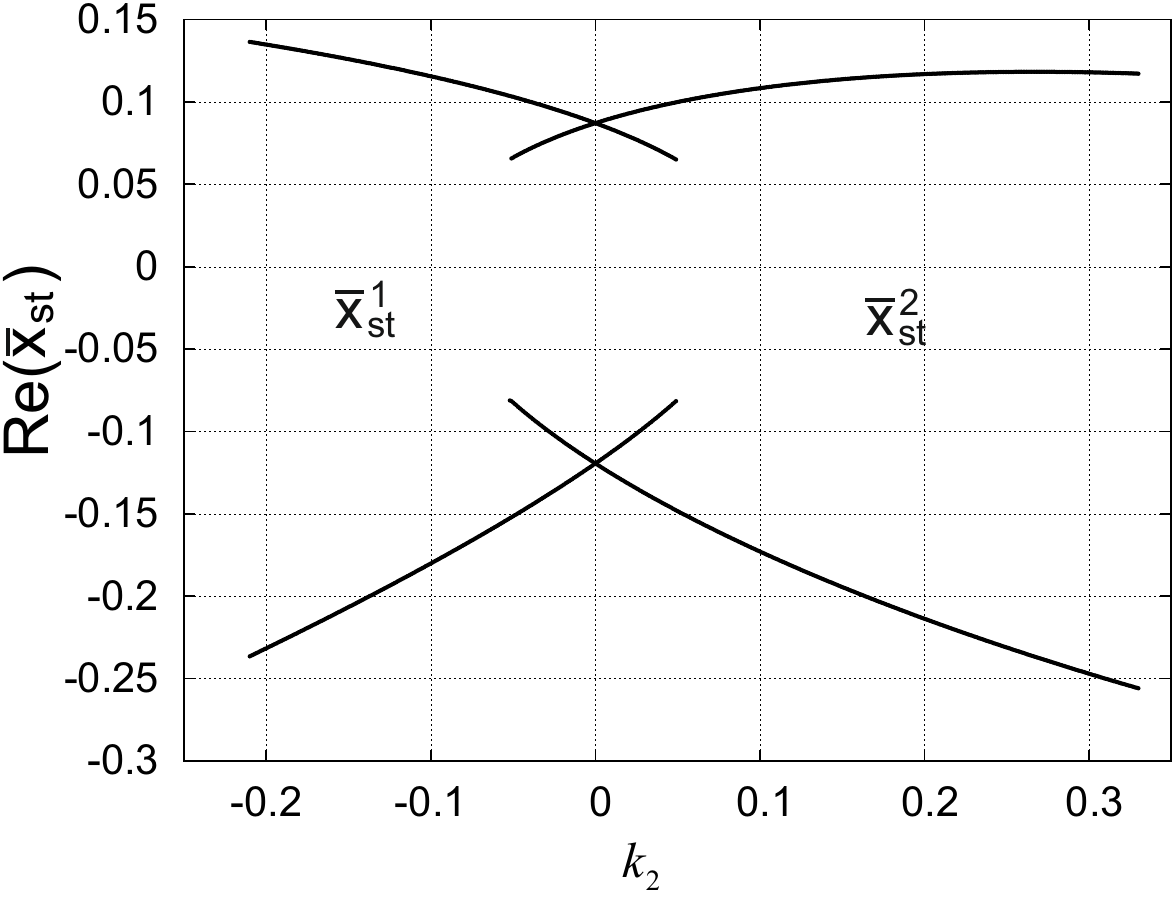}}\\
\subfigure[\label{fig:BFe4}]{\includegraphics[width=0.245\textwidth]{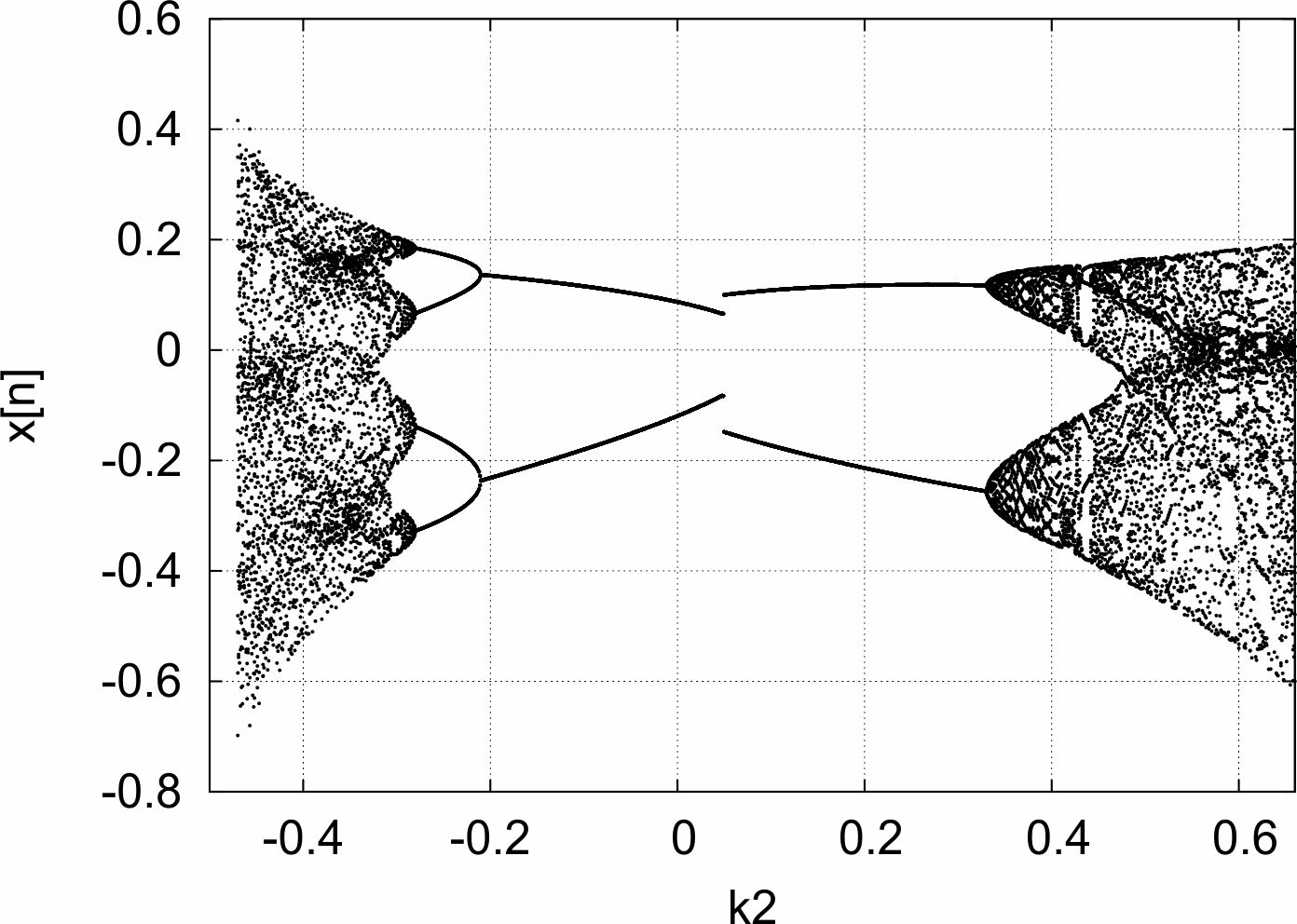}}~
\subfigure[\label{fig:BFe3}]{\includegraphics[width=0.245\textwidth]{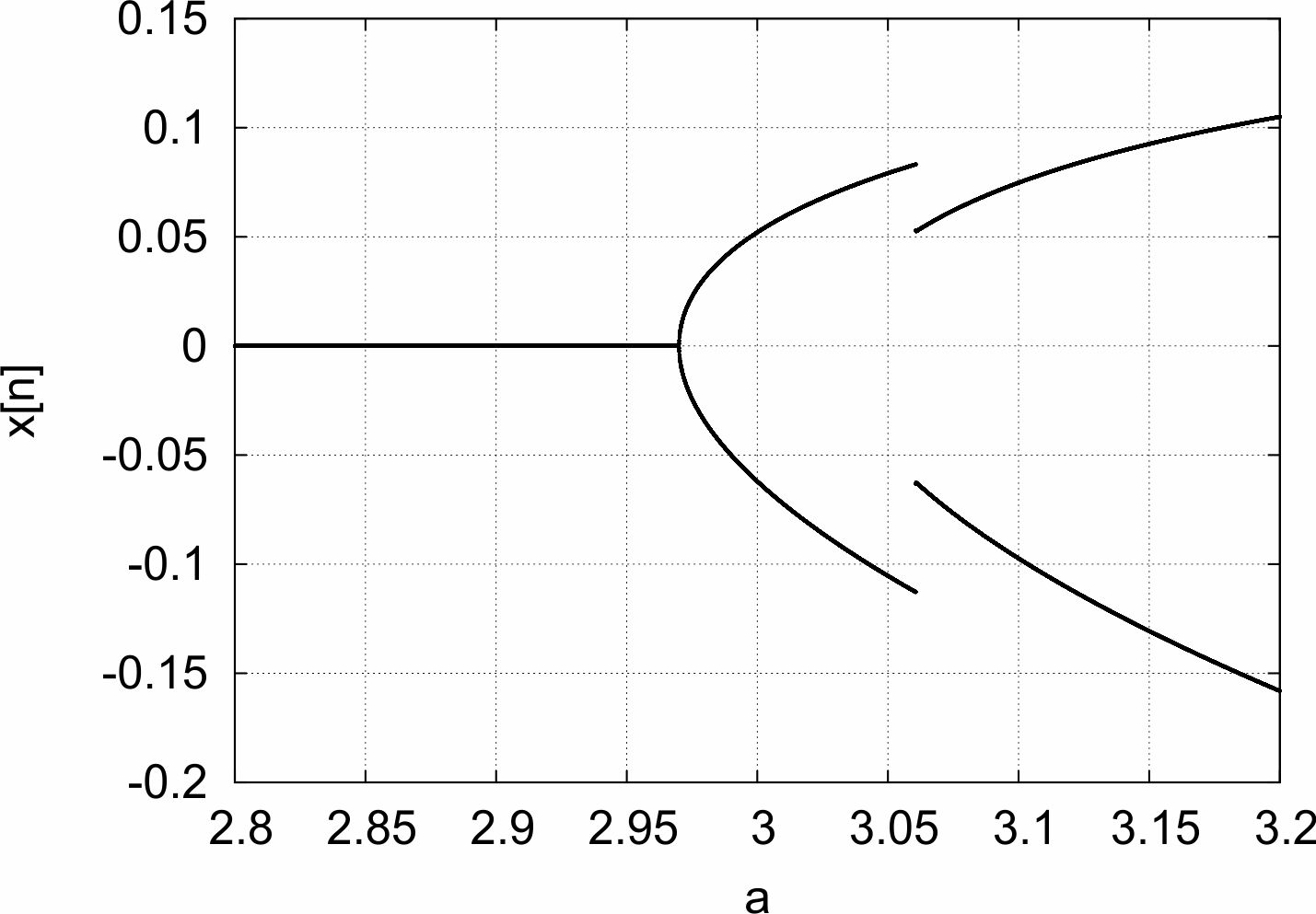}}
\caption{\small \textbf{(a)} Eigenvalues of the system (\ref{eq:couplSecond}) at $\alpha=3.1$; \textbf{(b)} Stable stationary states $\overline{x}_{st}\in R$ at $\alpha=3.1$; \textbf{(c)} Period-doubling ($k_2<0$) and Neimark-Sacker ($k_2>0$) bifurcation scenarios, $\alpha=3.1$, initial conditions randomly selected in the range [0 ... 1]; \textbf{(d)} Existence of multiple stable stationary states in the coupled system (\ref{eq:coupl121}), $m=2$, $k_2=0.05$, initial conditions randomly selected in the range [0 ... 0.1].
\label{fig:eigenVCoupl}\label{fig:BFe}}
\end{figure}
We observe two bifurcations at $k_2=-0.21$ and $k_2=0.33$; further, the system (\ref{eq:couplSecond}) has multiple stable stationary states. Since complex conjugated stationary states are also stable in this region of the parameter $k_2$, the linear stability analysis points to two bifurcation scenarios: first, a period-doubling scenario for $k_2<0$, second, a Neimark-Sacker bifurcation for $k_2>0$. Since nonlinear analysis is outside the scope of this work, we investigate the behavior of (\ref{eq:coupl121}) numerically. The change of bifurcation scenarios is demonstrated in Fig.~\ref{fig:BFe4}. We also observe a switch between stable stationary states by varying initial conditions $x_0^i$ in Fig.~\ref{fig:BFe3}. Thus, a variation of the coefficient $k_2$ (the distance between oscillating devices) primarily leads to changes in the amplitude of period-two signal or to an appearance of qualitatively different types of behavior.

Considering the problem of $m>2$, we note that values of $k_2<0$ lead to synchronization of amplitudes in period-two behavior (that is, all $x_n^i$ are the same in stationary state, see Fig.~\ref{fig:CollMirror}). We introduce
\begin{equation}
\label{eq:couplingTermRandom}
g^j=e-Random(),
\end{equation}
where \emph{Random()} is a function delivering random values between $0$ and $3 \cdot 10^{-3}$. Thus, we explore the behavior of the coupled system (\ref{eq:coupl121}) with random distances between AUVs and without calibration of $k_1$. In Fig.~\ref{fig:Coll1} we plot the dependence between $\frac{x_{n-1}}{x_{n}}$ and $m$.
\begin{figure}[h!]
\centering
\subfigure[\label{fig:Coll1}]{\includegraphics[width=0.245\textwidth]{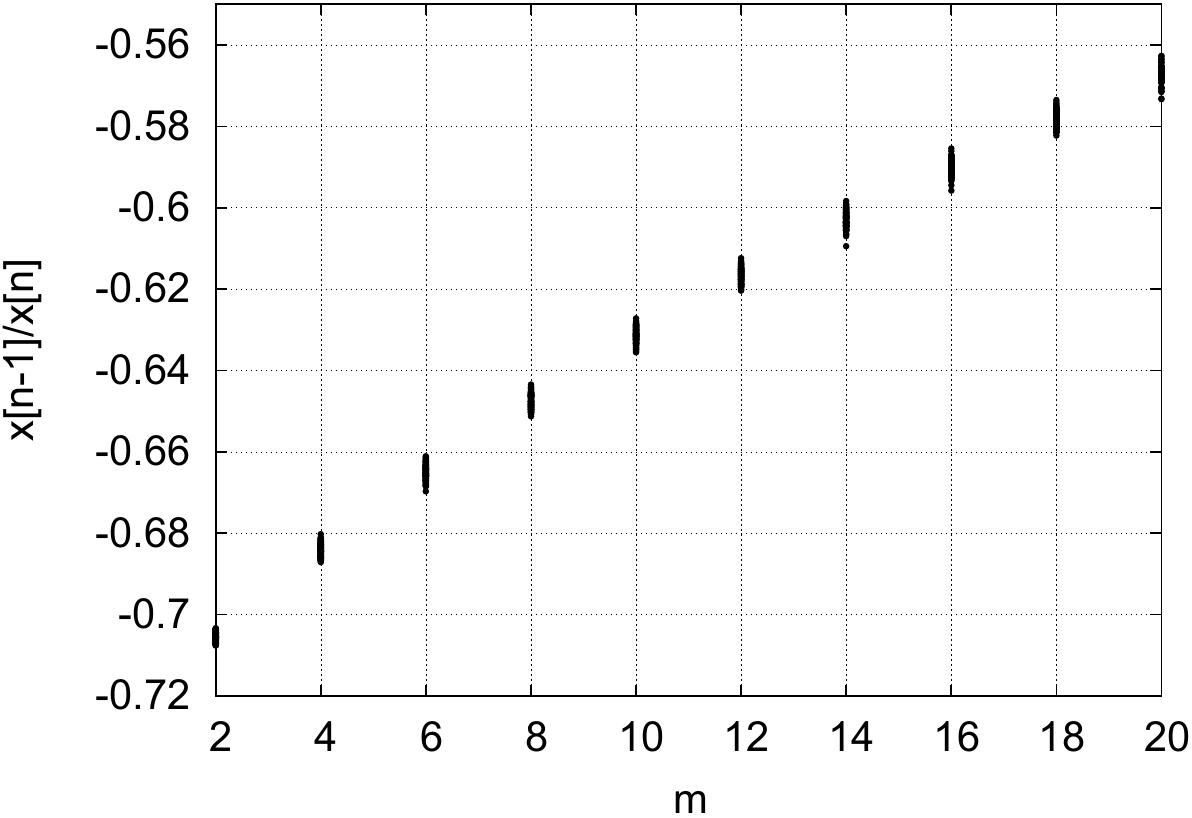}}~
\subfigure[\label{fig:Coll2}]{\includegraphics[width=0.245\textwidth]{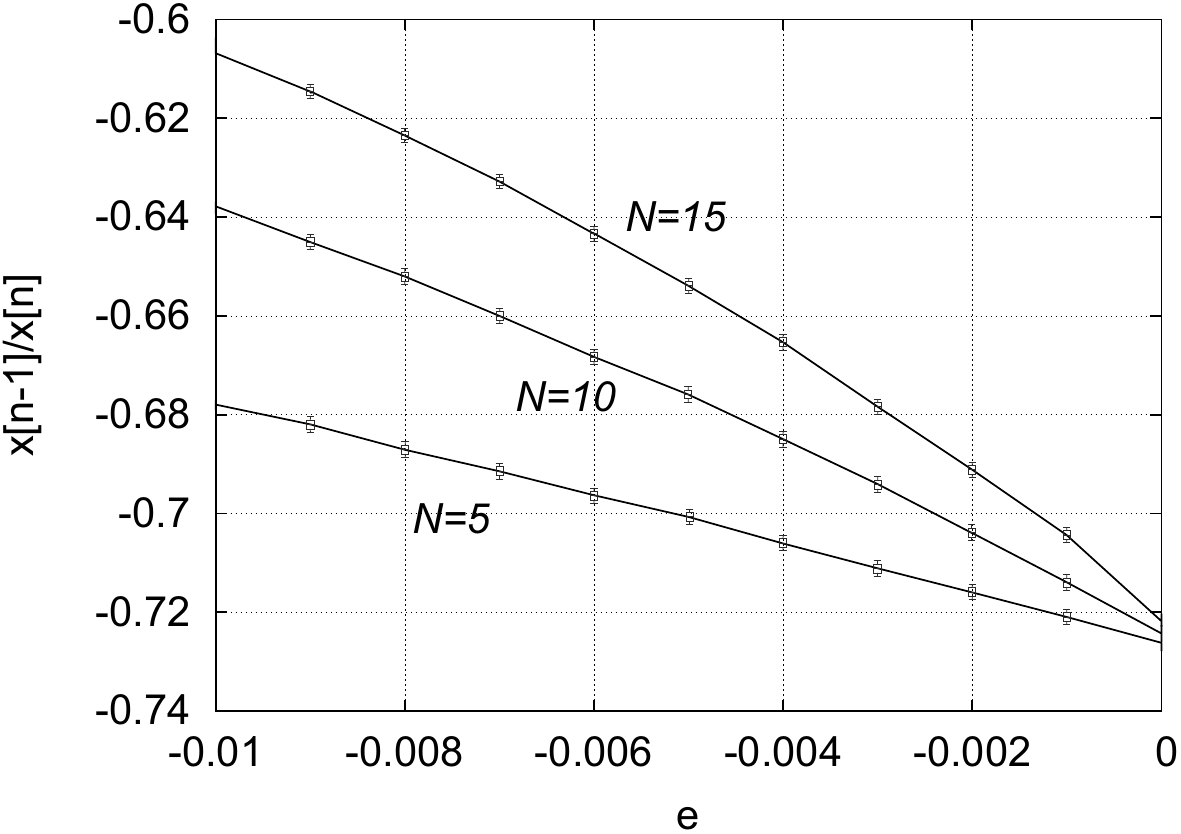}}
\caption{\small Strategies for the coupled system (\ref{eq:coupl12}) at $m>2$, initial conditions randomly selected in the range [0 ... 0.1], $\alpha=3.1$. At each $N$ 100 attempts are performed. \textbf{(a)} Dependence between $\frac{x_{n-1}}{x_{n}}$ and $m$, $k_2$ is selected as (\ref{eq:couplingTermRandom}), $e=-0.01$  \textbf{(b)} Dependence between $\frac{x_{n-1}}{x_{n}}$ and $e$ from (\ref{eq:couplingTermRandom}) for $m=5,10,15$.\label{fig:Coll}}
\end{figure}
This curve demonstrates a clear dependence on $m$, even when the distances between robots are different. Since the synchronous update strategy is used, the term $\sum_{j=1}^m g^j  x_n^{j}$ is calculated before evaluation of all $x_{n+1}^i$. As an alternative strategy, the number of devices can be fixed and a spatial distribution of the group can be explored. In Fig.~\ref{fig:Coll2} we plot the values of $\frac{x_{n-1}}{x_{n}}$ independent of $e$ from (\ref{eq:couplingTermRandom}) at fixed $m=5,10,15$. The value $e$ determines all $g^j$ from $\sum_{j=1}^m g^j  x_n^{j}$ and points to spatial configuration as the sum of all distances between the AUVs. We see that the local value $\frac{x_{n-1}}{x_{n}}$ can be used to estimate these global values.

\subsection{Electrical Mirror Mode}
\label{sec:DelyedFeedbackMode}

A swarm of AUVs in coupled oscillator mode with Eq.~(\ref{eq:coupledSystem}) can interact either with another swarm of AUVs or with the electrical mirror, as shown in Fig.~\ref{fig:DelayedMirror}. In the second case shown in Fig.~\ref{fig:DelayedMirror2}, the AUV's individual oscillations will be reflected by the mirror. Since the correlation between self- and reflected signals dramatically changes the dynamics, the swarm can collectively recognize this global situation.

The mirror device stores the common signal (i.e. the term $\sum_{j=1}^m g^j  x_n^{j}$) at each time step and sends it back with delay $\gamma$. This delayed signal creates
\begin{equation}
x_{n+1}^i = x_n^i (2-\alpha x_n^i - \alpha) + \sum_{j=1}^m g^j  x_n^{j}+\sum_{j=1}^m \tilde{g}^j  x_{n-\gamma}^{j},
\label{eq:coupl12}
\end{equation}
where $\tilde{g}^j$ at the delayed terms represents the previous state of the system. For a system with no spatial changes between emitting devices we can set $\tilde{k}_2^j={k}_2^j$.

The electrical mirror corresponds to the delayed feedback produced when the same signal with delayed $\gamma$ steps $x_{n-\gamma}$ is added to the system (for $\gamma=0$ we add a non-delayed signal). Assuming the coefficient $k_1$ is calibrated and the self-signal is filtered out, we rewrite Eq.(\ref{eq:transformedSystem}) in the following form
\begin{equation}
\label{eq:coupledSystemDF}
x_{n+1} =  \alpha x_n (2-\alpha x_n - \alpha) + k_3 x_{n-1} + k_4 x_{n-2}.
\end{equation}
Equation (\ref{eq:coupledSystemDF}) can be considered as a model of time-delayed feedback for two emitting devices (one oscillator and one electrical mirror). Each of the delayed terms has a specific influence on the dynamics: they shift the corresponding bifurcation and to some extent 'stretch' or 'squeeze' the whole dynamics. In Fig.~\ref{fig:BFDelay1} we demonstrate the term $0.45 x_{n-1}-0 x_{n-2}$, which shifts the first period-doubling bifurcation backwards with regard to parameter $\alpha$. Fig.~\ref{fig:BFDelay2} shows the term $-0.45x_{n-1}-0.45x_{n-2}$, which shifts the first bifurcation forwards and the second bifurcation backwards.
\begin{figure}[ht]
\centering
\subfigure[\label{fig:BFDelay1}]{\includegraphics[width=0.245\textwidth]{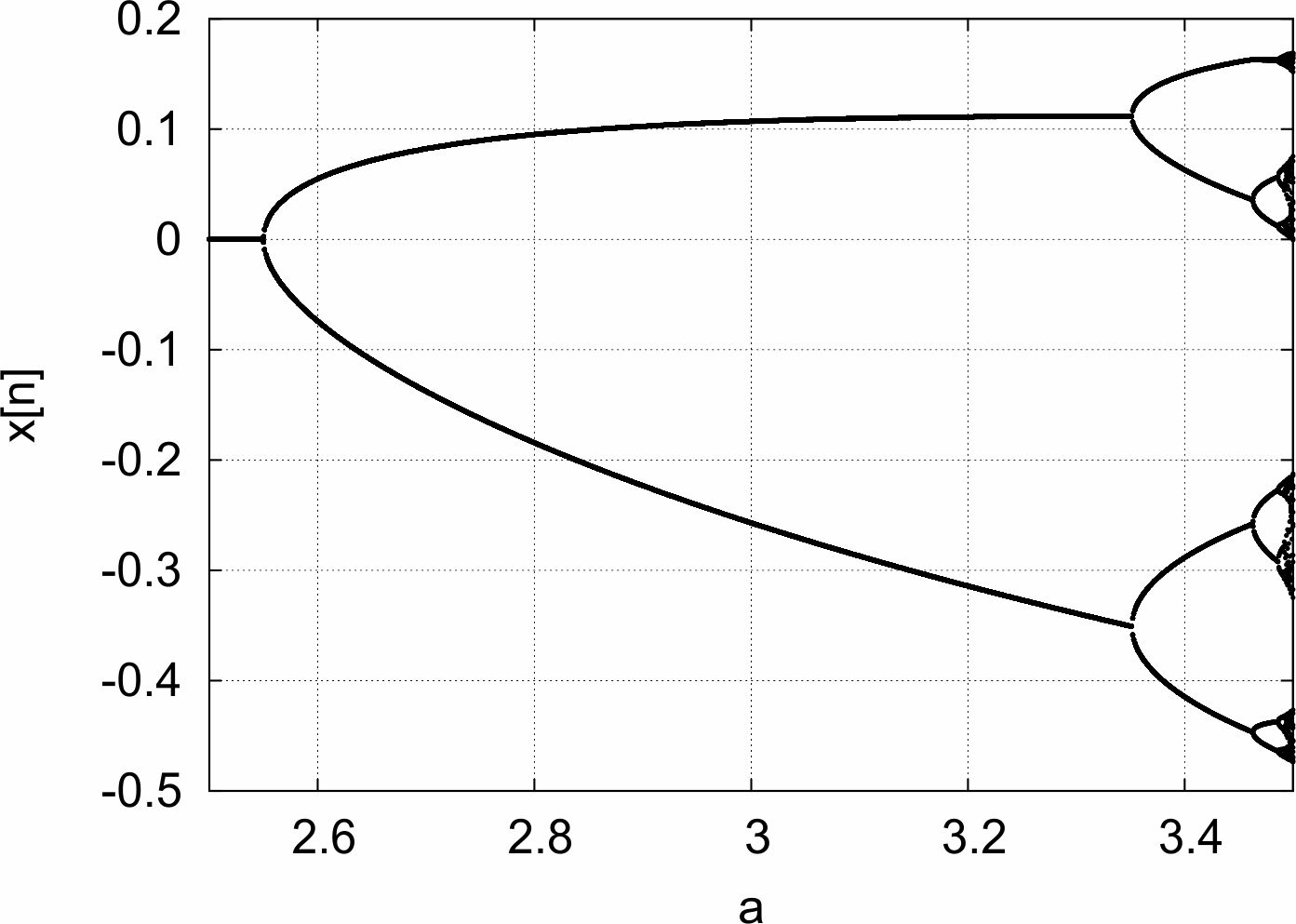}}~
\subfigure[\label{fig:BFDelay2}]{\includegraphics[width=0.245\textwidth]{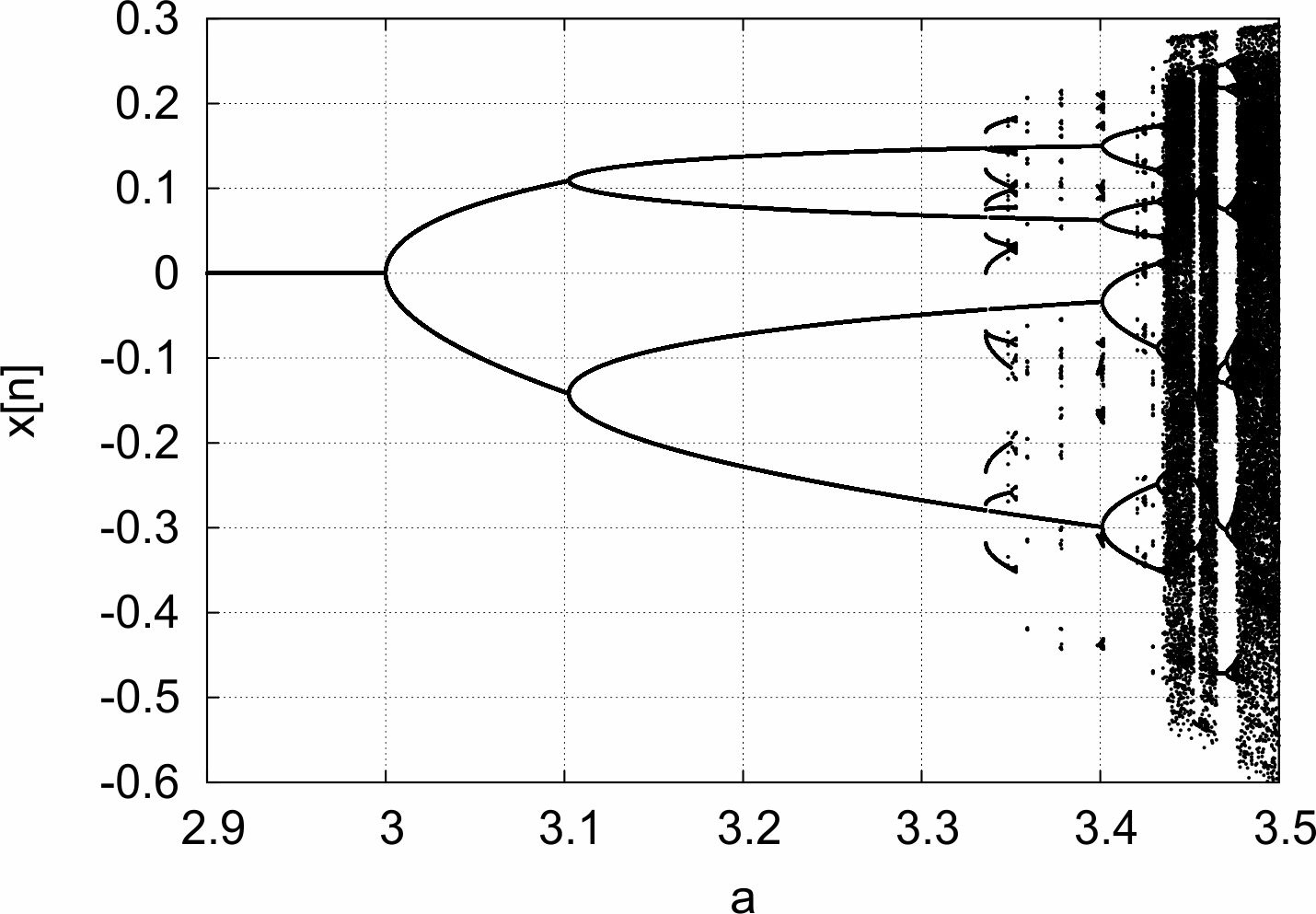}}
\caption{\small Bifurcation diagrams with different delayed feedback: \textbf{(a)} $0.45 x_{n-1}-0 x_{n-2}$; \textbf{(b)} $-0.45 x_{n-1}-0.45 x_{n-2}$.
\label{fig:BFDelay}}
\end{figure}
Since the delayed feedback produces behavior qualitatively different from the coupled mode, a robot can recognize whether another emitting device is a "robot" or a "mirror". The approach considered in Sec.~\ref{sec:linCoupledMode} uses different amplitudes of the period-two oscillation, here a transition to non-periodic or to period-four oscillation by means of delayed feedback is implemented. These two possibilities are shown in Fig.~\ref{fig:BFDelayK21}, where the values of $k_3$ in the range  $[-1...-0.0333]$ at $\alpha=3.1$ lead to non-periodic behavior and in Fig.~\ref{fig:BFDelayK31} the values of $k_4$ in the range $[0...0.3]$ at $k_2=0.138$ and $\alpha=3.1$ lead to the period-four behaviour (this case is sensitive to the choice of initial conditions).
\begin{figure}[ht]
\centering
\subfigure[\label{fig:BFDelayK21}]{\includegraphics[width=0.245\textwidth]{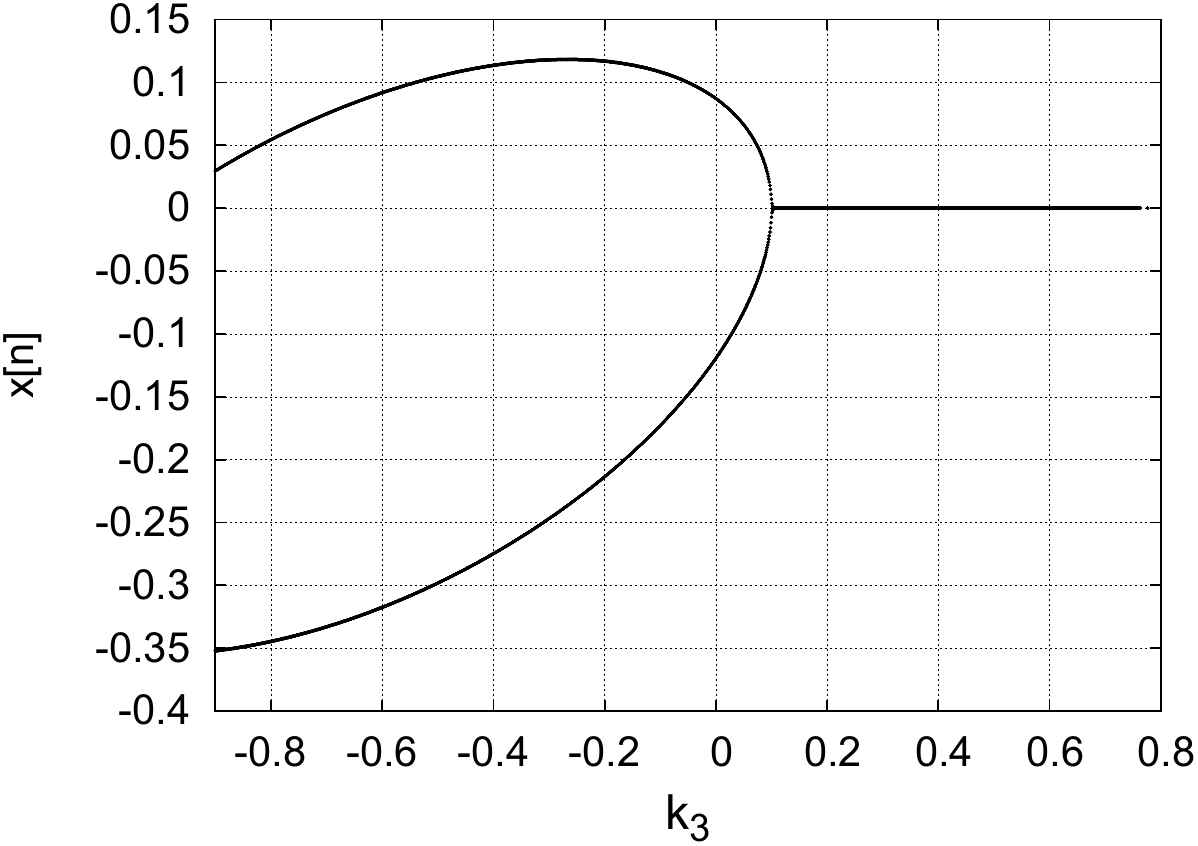}}~
\subfigure[\label{fig:BFDelayK31}]{\includegraphics[width=0.245\textwidth]{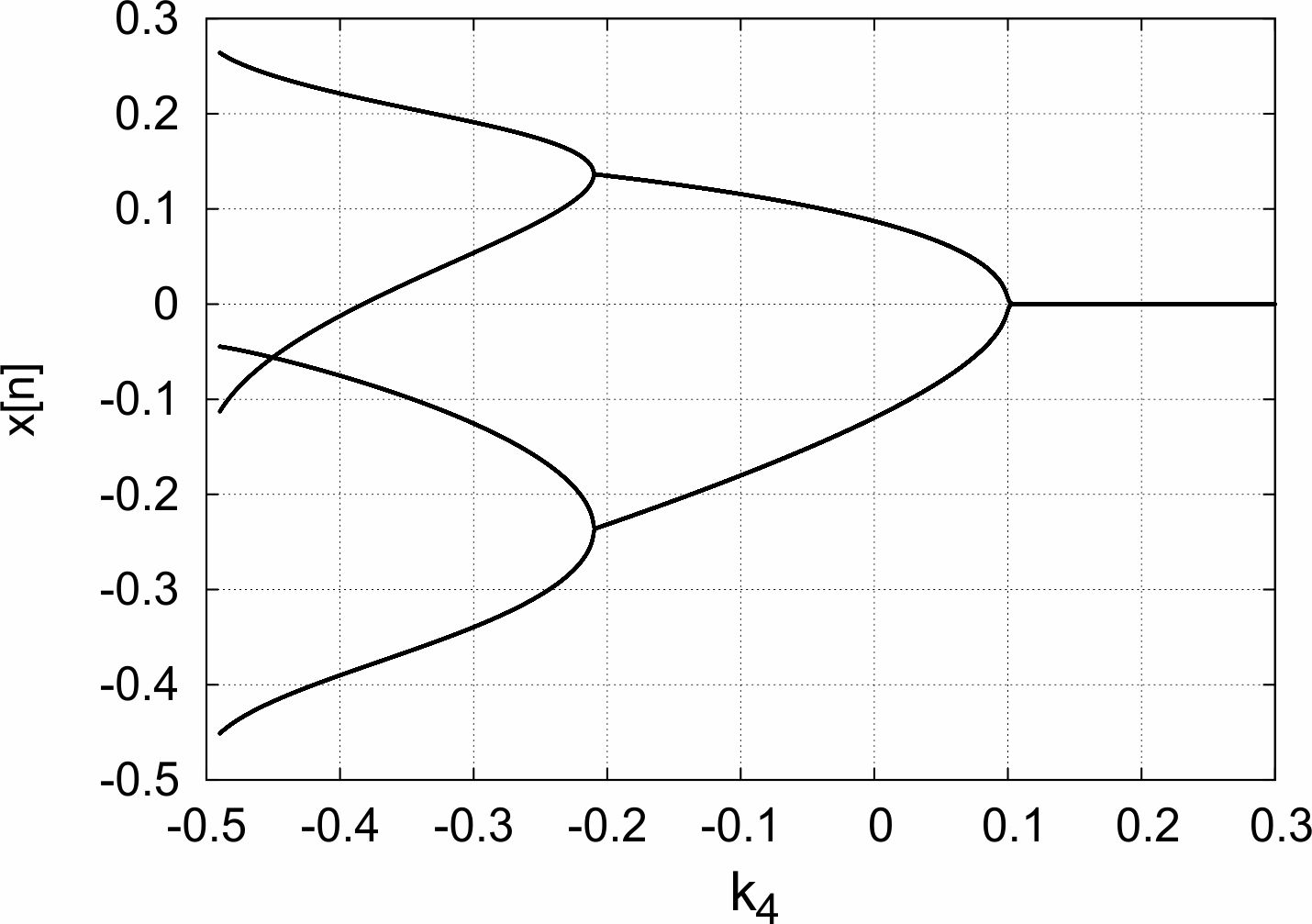}}
\caption{\small Bifurcation diagrams with different delayed feedback at $\alpha=3.1$: \textbf{(a)} $k_3=0$; initial conditions are random between the intervals [0...0.1]; \textbf{(b)} $k_2=0$, initial conditions are random between the intervals [0...0.1].
\label{fig:BFDelayK2}}
\end{figure}

To demonstrate analysis of Eq.(\ref{eq:coupledSystemDF}), we rewrite it as
\begin{eqnarray}
x_{n+1} &=& x_n (2-\alpha x_n - \alpha) + k_3 y_n+k_4 z_n,\nonumber\\
y_{n+1} &=& x_n,\nonumber\\
z_{n+1} &=& y_n.
\label{eq:delayed3eqs}
\end{eqnarray}
The non-periodical stationary states are $x_{st}=y_{st}=z_{st}=\{0,\frac{a-1+k_3+k_4}{a}\}$. The Jacobian of the system (\ref{eq:delayed3eqs}) is given by
\begin{equation}
\label{eq:delyedJacobian}
\left[
\begin {array}{ccc}
2 - 2 x \alpha - \alpha & k_3&k_4\\
1&0&0\\
0&1&0
\end {array}
 \right].
\end{equation}
We plot eigenvalues $|\lambda_{1,2,3}|$ for two cases of $k_4=0$ and $k_3=0$ in Fig.~\ref{fig:EigDel}.
\begin{figure}[ht]
\centering
\subfigure[\label{fig:EigDel1}]{\includegraphics[width=0.245\textwidth]{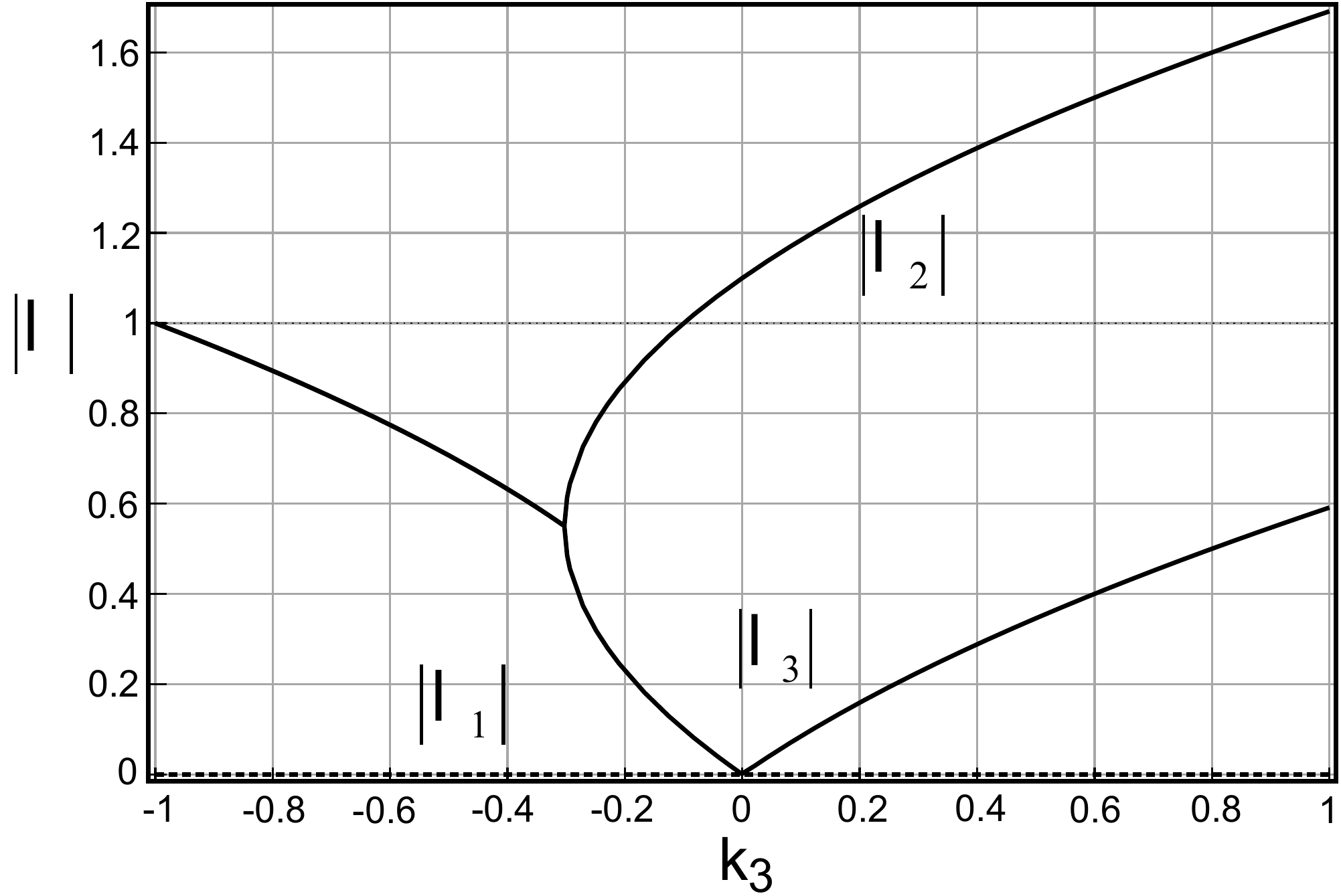}}~
\subfigure[\label{fig:EigDel2}]{\includegraphics[width=0.245\textwidth]{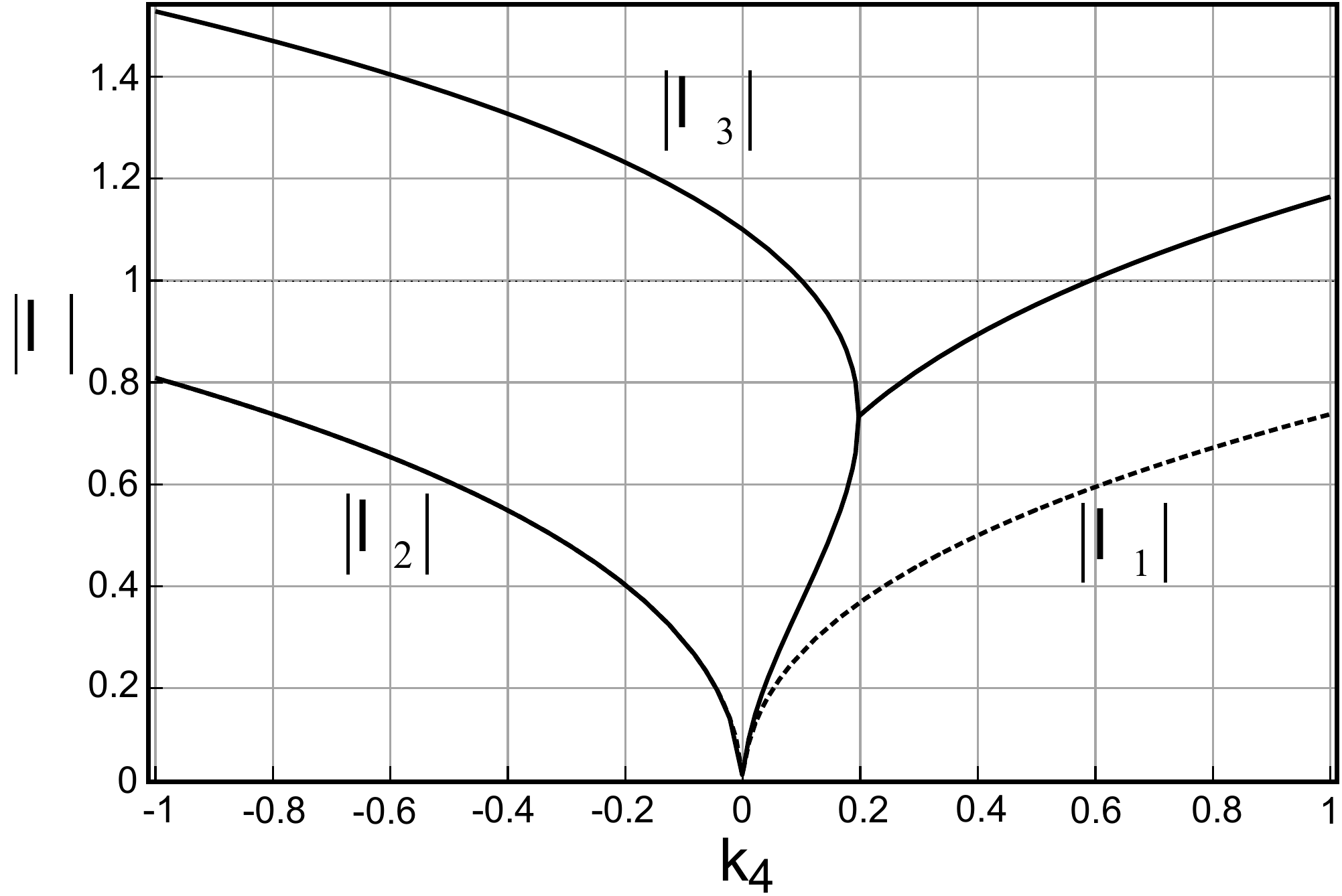}}
\caption{\small Eigenvalues of (\ref{eq:delyedJacobian}) evaluated on non-periodical stationary states at $\alpha=3.1$ for \textbf{(a)} $k_4=0$ and \textbf{(b)} $k_3=0$.
\label{fig:EigDel}}
\end{figure}
As we can see, both the terms $k_3 x^i_{n-1}$ and $k_4 x^i_{n-2}$ influence the stability of eigenvalues and thus can shift the first period-doubling bifurcation forwards and backwards. Similarly, we can investigate the linear stability of the second period-doubling bifurcation by using the second-iterated map obtained from (\ref{eq:delayed3eqs}). By varying $k_3$ and $k_4$, the first and second bifurcations can be simultaneously shifted, as shown in Fig.~\ref{fig:BFDelay2}. Thus, the electrical mirror will produce period-four motions, whereas coupled oscillators without an electrical mirror will produce only period-two motions. Based on this qualitatively different behavior, all AUVs can locally recognize a self-reflection.

In the case of several emitting devices (i.e. $m>2$), we perform a numerical simulation of the system (\ref{eq:coupl12}). Fig.~\ref{fig:CollMirror} demonstrates the temporal behavior of this system with and without the delayed term.
\begin{figure}[ht]
\centering
\subfigure[\label{fig:CollMirror1}]{\includegraphics[width=0.245\textwidth]{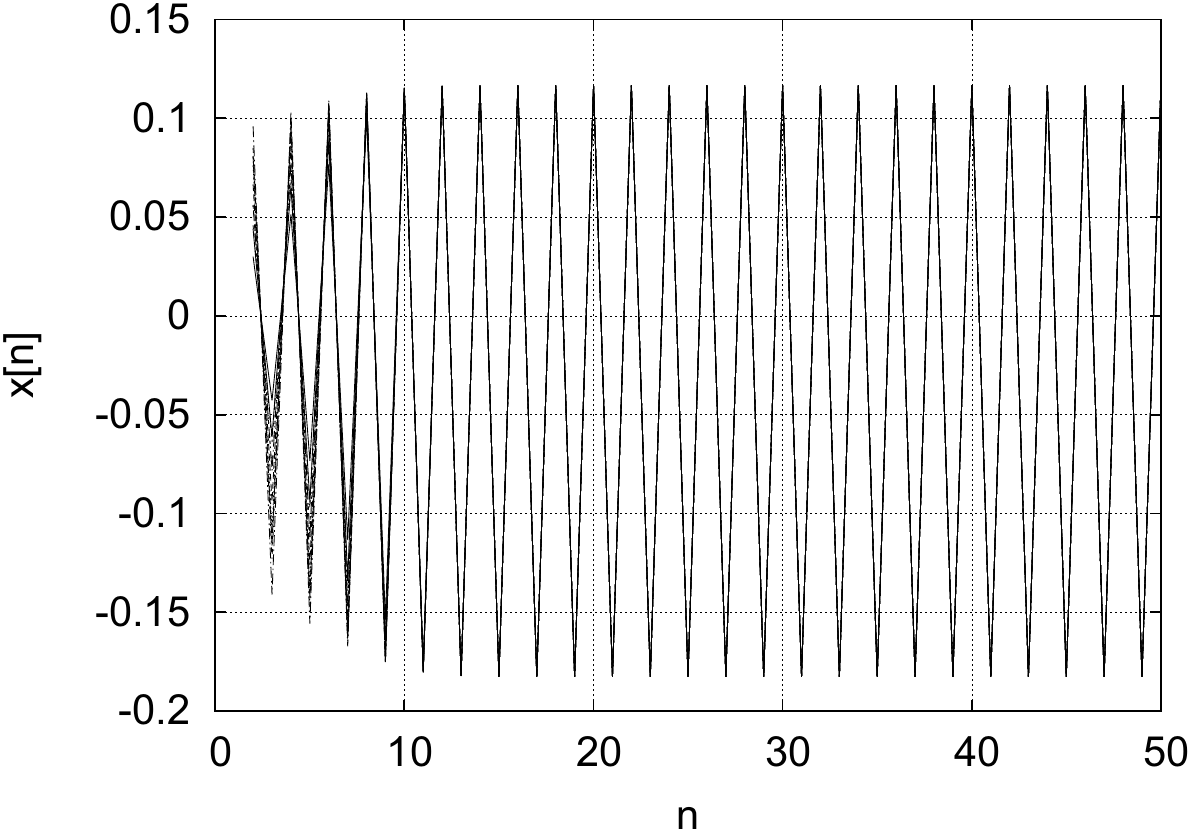}}~
\subfigure[\label{fig:CollMirror2}]{\includegraphics[width=0.245\textwidth]{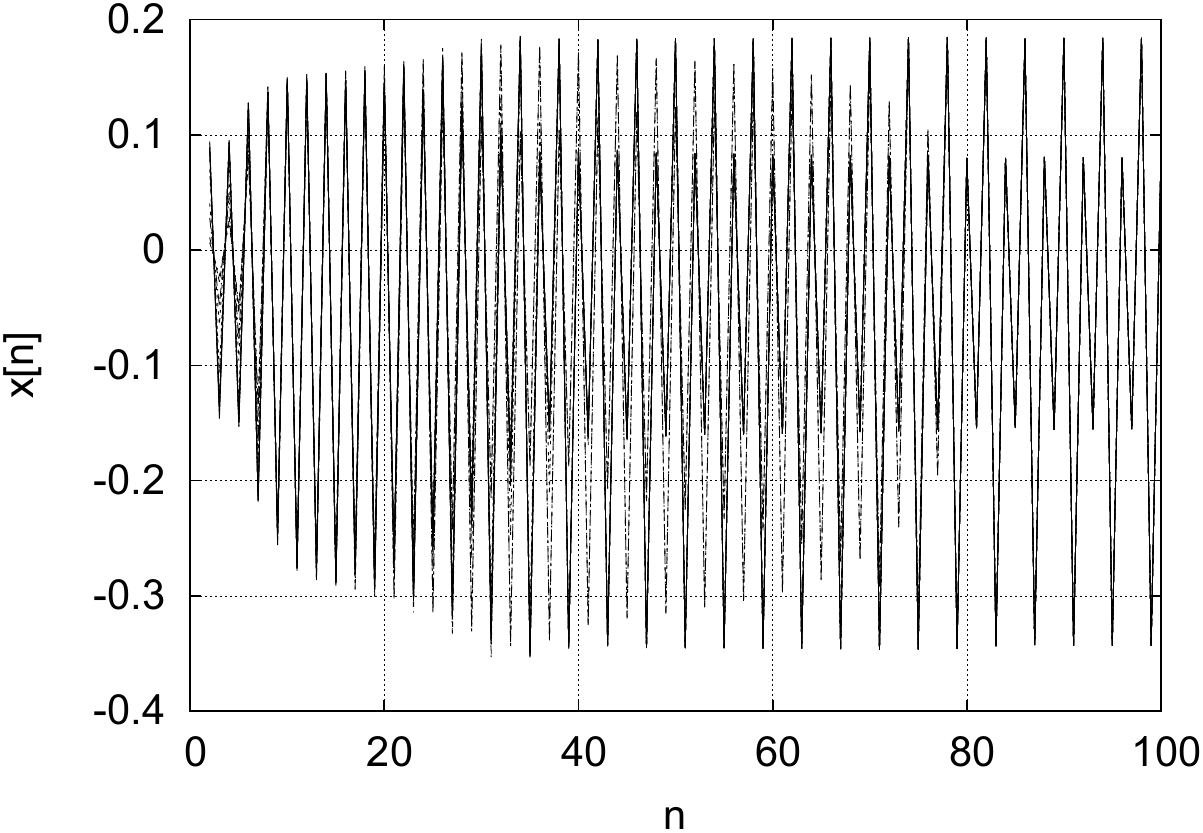}}
\caption{\small Temporal behavior of the system (\ref{eq:coupl12}), $m=10$, $g^j$ is defined by (\ref{eq:couplingTermRandom}), $e=-0.01$; \textbf{(a)} period-two motion without electrical mirror, we observe global synchronization within the first ten steps; \textbf{(b)} period-four motion with electrical mirror $2\sum_{j=1}^m \tilde{g}^j  x_{n-2}^{j}$, we observe global synchronization within the first 80 steps.
\label{fig:CollMirror}}
\end{figure}
The numerical value of non-delayed and delayed term $\sum_{j=1}^m {g}^j  x_{n}^{j}$ is small; thus an amplification of the delayed signal is needed to shift the dynamics into the period-four region. In both cases a global synchronization is observed.

\section{Experimental Setup}
\label{sec:setup}

The experimental setup includes both the potential mode operation, shown in Fig. \ref{fig:ExperimentSchemeSingle} and \ref{fig:InputFilter}, and the current mode operation, shown in Fig. \ref{fig:CurrentMode}. Experiments have been performed with 2-8 electric sensors. Sending and receiving electrodes are placed in plastic tubes on a movable platform in aquarium, see Fig. \ref{fig:experiment11} for potential mode and Fig. \ref{fig:experimentCurrent} for current mode. Electrodes are connected to electronic boards with analog filters, amplifiers, DAC and ADC. During all experiments, the water in aquarium was grounded and all electronic devices, for example the water pump, were switched off. The laptops used for data collection ran on batteries for each of the emitting/receiving devices separately. Both potential and current mode systems have been implemented in AUVs from CoCoRo \cite{CoCoRo}, subCULTron \cite{Loncar19}, \cite{Thenius16subCULT} projects, shown in Figs. \ref{fig:lili} and \ref{fig:amussel}, and several other developments \cite{AquaJelly}, \cite{ANGELS}, such as active electric mirror, see Fig. \ref{fig:mirror}.

\begin{figure}[ht]
\centering
\subfigure[\label{fig:setupTwo}]{\includegraphics[width=0.245\textwidth]{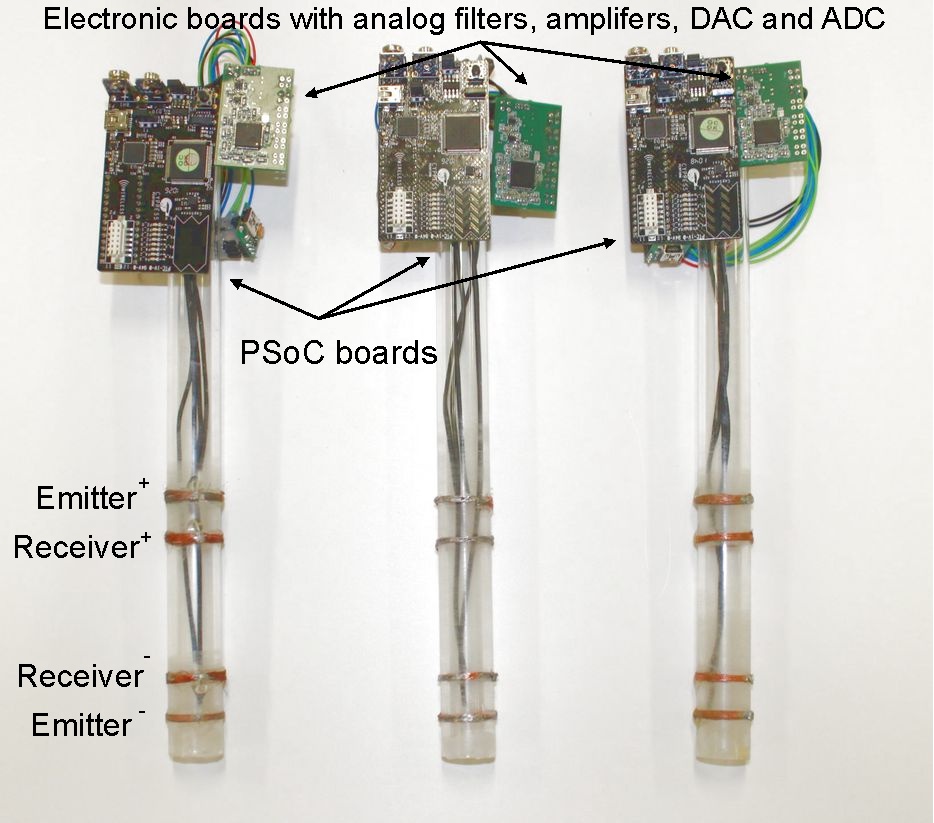}}~
\subfigure[\label{fig:experiment11}]{\includegraphics[width=0.245\textwidth]{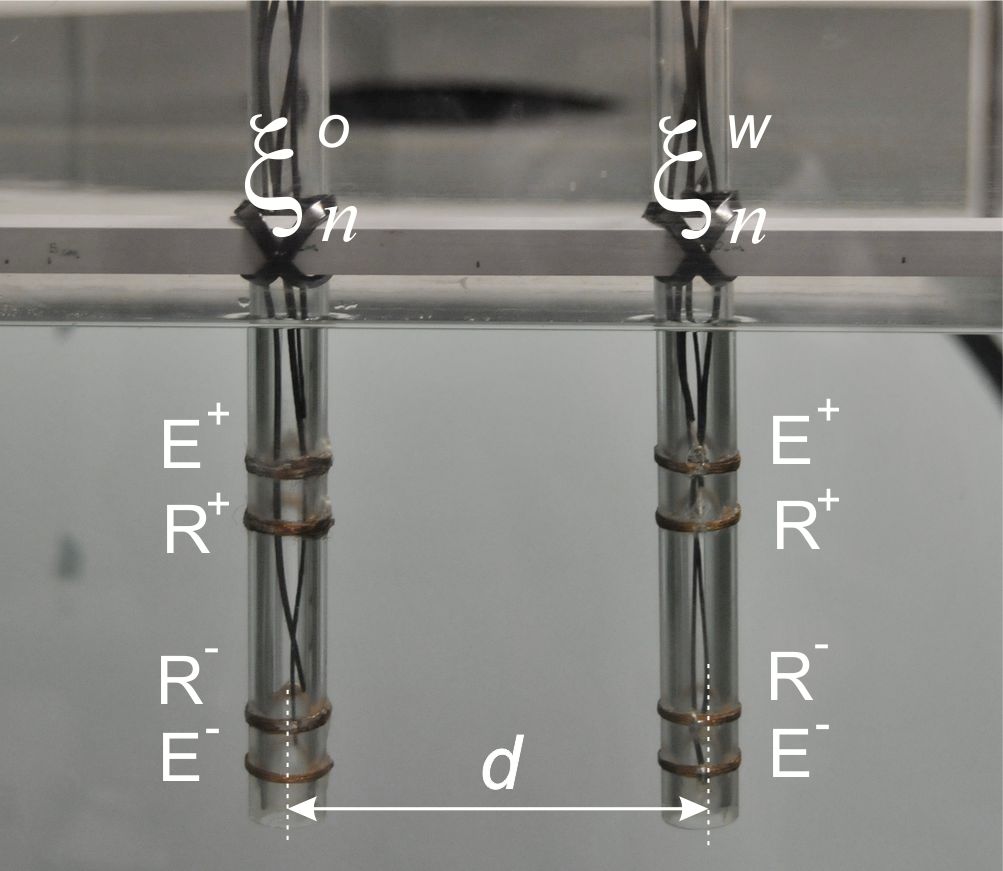}}
\subfigure[\label{fig:experimentCurrent}]{\includegraphics[width=0.45\textwidth]{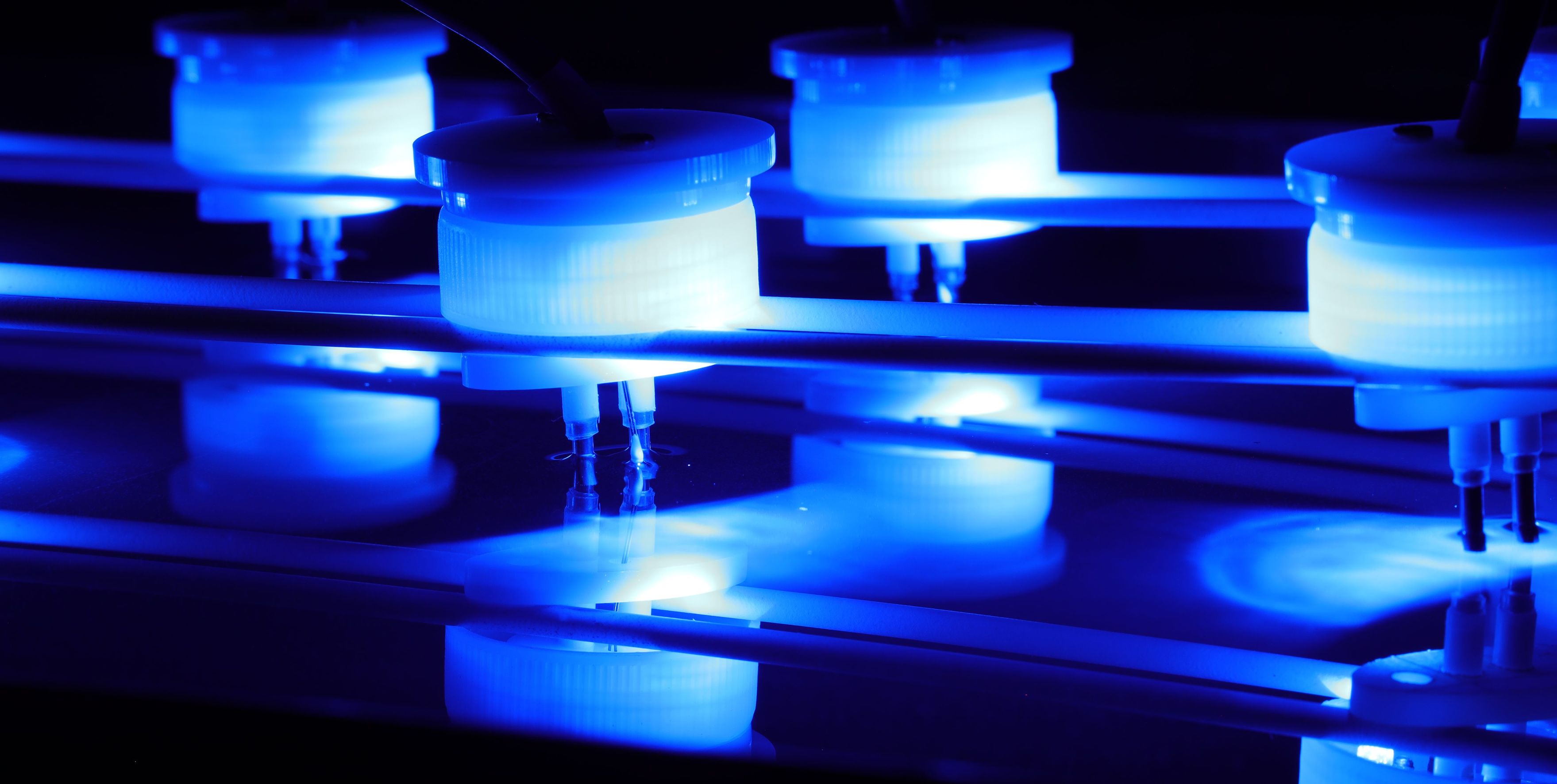}}
\caption{\small \textbf{(a,b)} Experimental set-up with the potential mode electrodes (four electrodes) and \textbf{(c)} current mode electrodes (two electrodes are slightly removed from water for demonstration purposes).
\label{fig:commonSetup}}
\end{figure}

Two receiving electrodes in potential mode are connected to a high-pass filter with the cut-off frequency $16$~Hz, see Fig.~\ref{fig:InputFilter}. 
\begin{figure}[ht]
\centering
\subfigure{\includegraphics[width=0.245\textwidth]{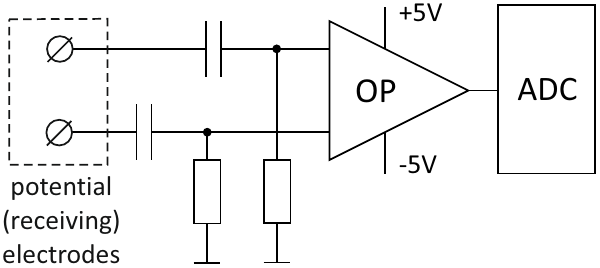}}~
\caption{\small Input high-pass filters in potential mode system.\label{fig:InputFilter}}
\end{figure}
The high-pass filter is needed to filter any unwanted constant electrical fields that may be present in the experimental environment. The output of these filters is connected to the differential inputs of an instrumental OpAmp. With a supply voltage of $\pm 5$~V, the amplifier is able to process input voltages between $\pm 10$~V without saturation. The emitting electrodes are connected to the outputs of a 2-channel DAC. This DAC can generate an analog voltage between $0$~V and $+5$~V on both outputs independently, which leads to a relative voltage of $\pm 5$~V between the two electrodes. Additionally the outputs can be deactivated (high impedance) to reduce unwanted influence on ADC.

\begin{figure}[ht]
\centering
\subfigure[\label{fig:timeOsciDiagram1}]{\includegraphics[width=0.245\textwidth]{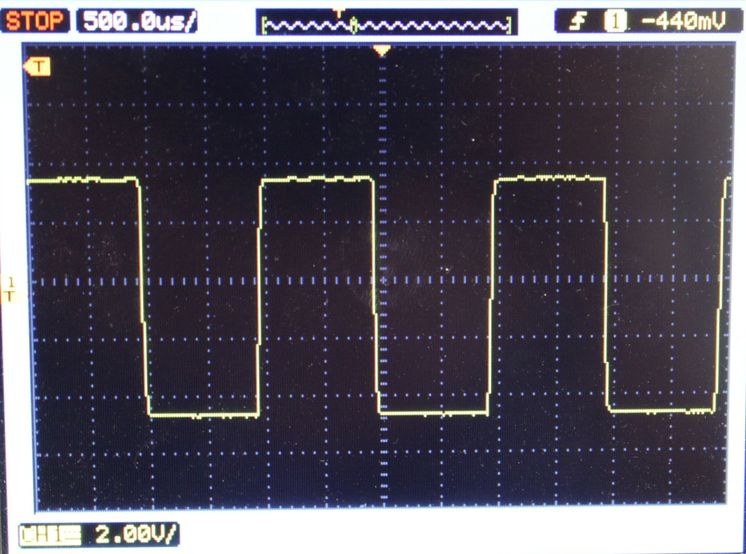}}~
\subfigure[\label{fig:timeOsciDiagram2}]{\includegraphics[width=0.245\textwidth]{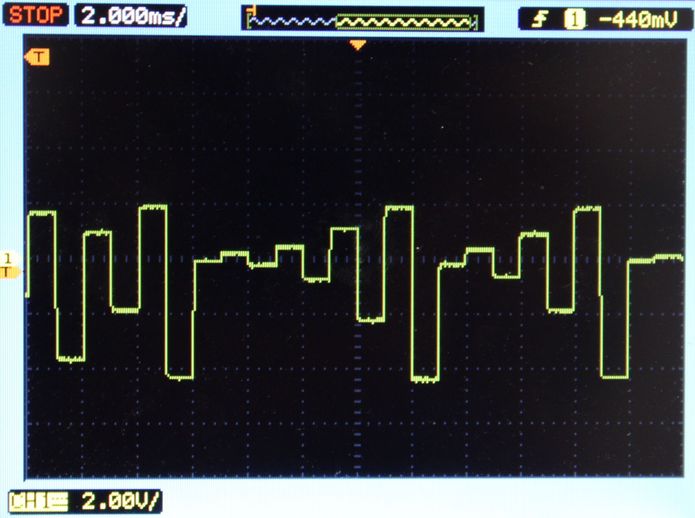}}\\
\subfigure[\label{fig:timeOsciDiagram3}]{\includegraphics[width=0.245\textwidth]{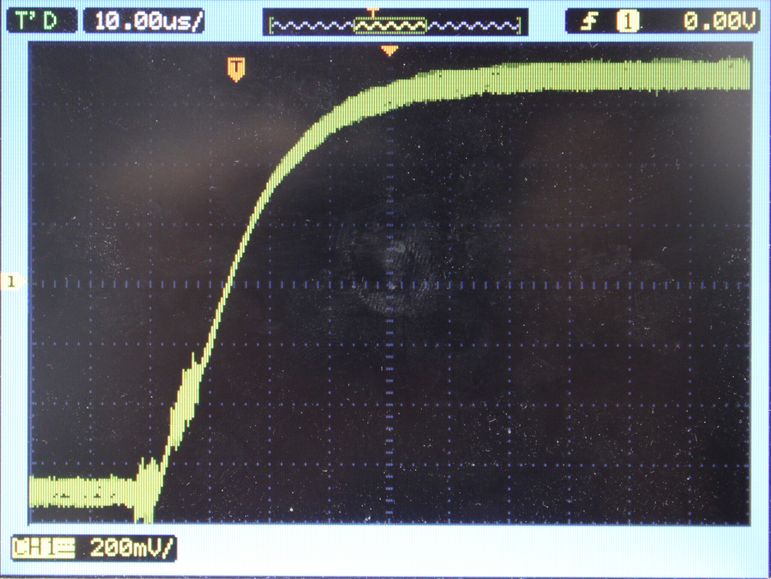}}~
\subfigure[\label{fig:timeOsciDiagram4}]{\includegraphics[width=0.245\textwidth]{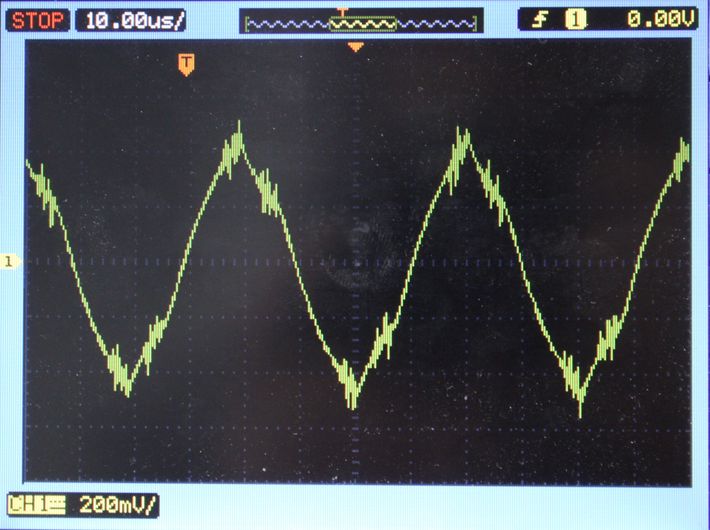}}
\caption{\small Measured amplitudes of signals on emitting electrodes: \textbf{(a)} period-two signal of the system (\ref{eq:transformedSystem}) at $\alpha=3.1$, $x_0=0.1$, $k_{DAC}=2000*6$; \textbf{(b)} chaotic signal in the system (\ref{eq:transformedSystem}) at $\alpha=3.7$, $x_0=0.1$, $k_{DAC}=2000*0.175$; \textbf{(c)} time needed to raise the voltage between levels, about $50\mu s$; \textbf{(d)} minimal delay from software and SPI communication to switch between voltage levels, about $20\mu s$. \label{fig:timeOsciDiagram}}
\end{figure}

For mapping between the values of $x_n$ and the output voltage $\pm 5$~V we use the coefficient $k_{DAC}$, calculated from the condition $|x_n| k_{DAC} < 2048$ for the maximum positive and negative values of $x_n$. In contrast to simulations of coupled maps, where all maps are iterated sequentially, real experimental hardware iterates all maps in parallel. This imposes several conditions on the timing of $x$, ${\xi^o_n}$ and ${\xi^w_n}$ . Since all variables of the RHS of equation (\ref{eq:coupledSystem}) are of time $n$, we first write $x$ into the DAC, read $\downarrow{\xi}$ from the ADC and then iterate the map. The measured output amplitudes of the signals for the period two ($x_n=0.08714, -0.11940$) and chaotic ($x_n=[0.25,.., -0.75]$) modes are shown in Figs.~\ref{fig:timeOsciDiagram1} and \ref{fig:timeOsciDiagram2}. We measured the time needed to raise the level of voltage, as shown in Fig.~\ref{fig:timeOsciDiagram3}. Switching between $\pm0.7V$, takes about $50\mu s$. The minimum delay from the software and SPI communication for switching between two levels of voltage is about $20\mu s$, see Fig.~\ref{fig:timeOsciDiagram4}.

\begin{figure}[ht]
\centering
\subfigure[\label{fig:lili}]{\includegraphics[width=0.25\textwidth]{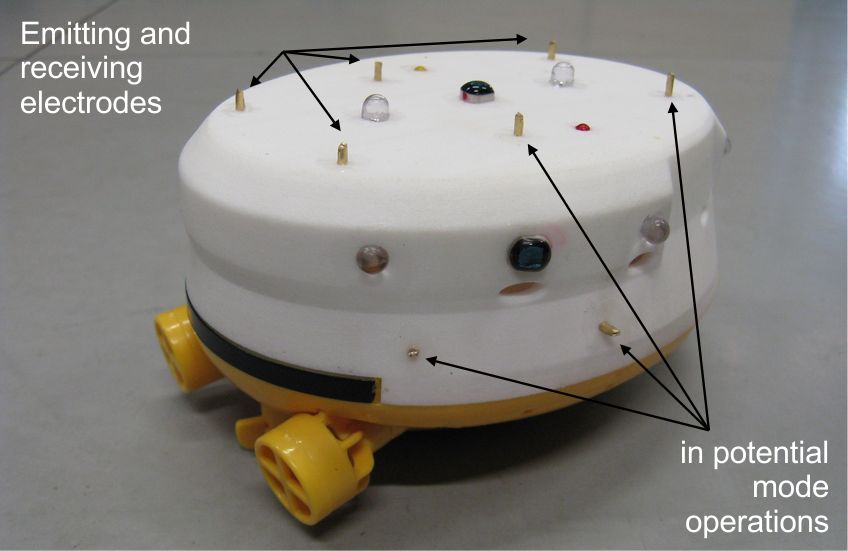}}~
\subfigure[\label{fig:amussel}]{\includegraphics[width=0.24\textwidth]{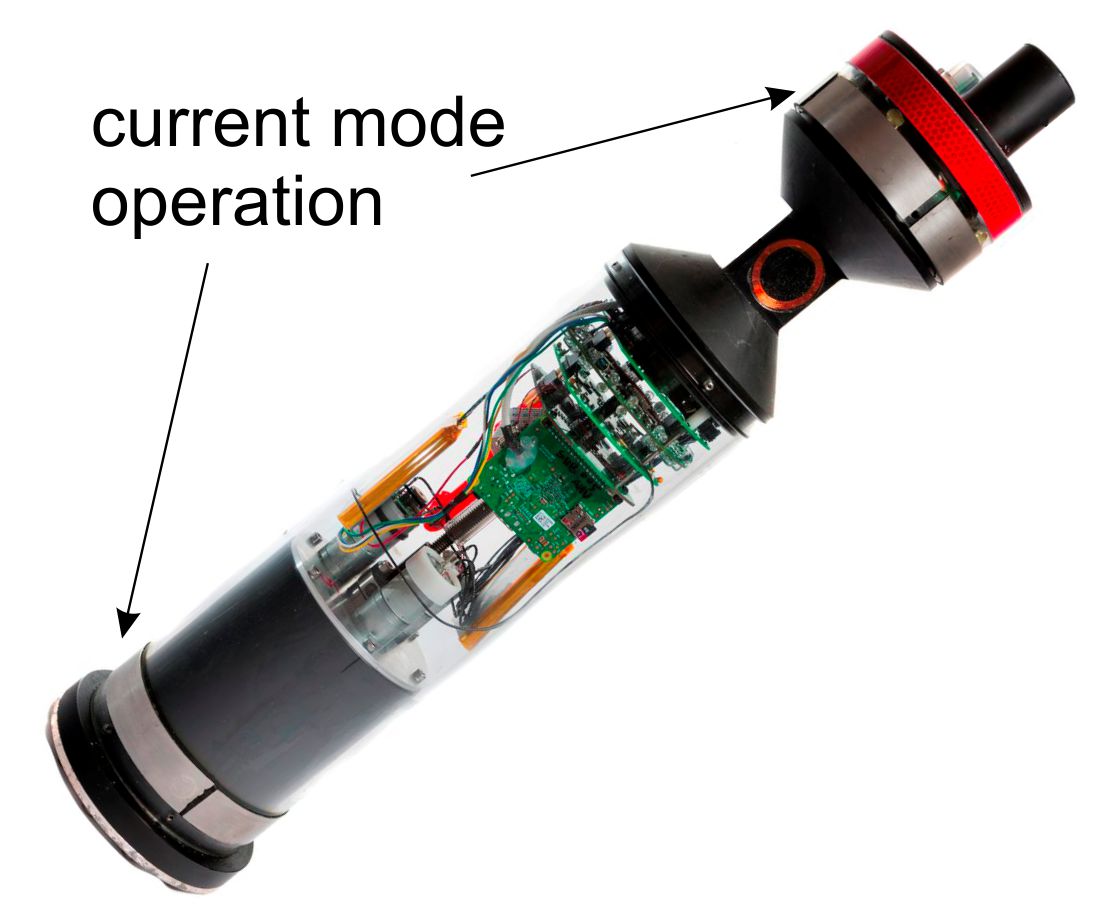}}
\caption{\small Examples of AUVs that use the development with electrical sensors: \textbf{(a)} the potential mode in AUVs from the CoCoRo project \cite{CoCoRo} and \textbf{(b)} the current mode in AUVs from the subCULTron project \cite{Loncar19}, \cite{Thenius16subCULT}. \label{fig:commonSetupAUV}}
\end{figure}

\section{Experiments}
\label{sec:experiments}

\subsection{Part 1: Dependency Between Emitted and Received Signals}
\label{sec:exp1}

In the first experiment we investigated the dependency between $\uparrow{\xi^i_n}$ and $\downarrow{\xi^i_n}$ in potential mode by varying the distance $d$ between two oscillating devices, see Fig.~\ref{fig:experiment12}. The first device sends the generated signal $\uparrow{\xi^i_n}$ using Eq.~(\ref{eq:coupledSystem}) without receiving $\downarrow{\xi^i_n}$; the second device receives $\downarrow{\xi^i_n}$ without sending $\uparrow{\xi^i_n}$.
\begin{figure}[ht]
\centering
\subfigure{\includegraphics[width=0.45\textwidth]{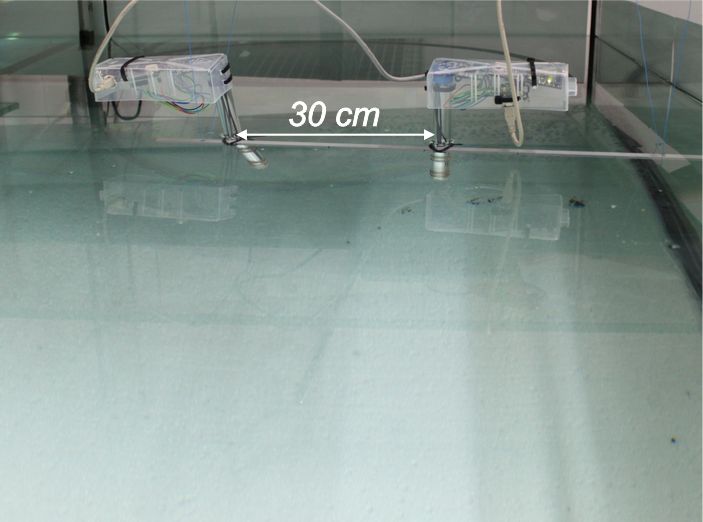}}
\caption{\small Experimental setup for the first experiment, the position 2 in the middle of aquarium. \label{fig:experiment12}}
\end{figure}

We expect to see two different values of signals depending on obstacles \cite{Baffet08}. To explore this effect, two experiments are performed: one with the electrodes installed on the glass wand of the aquarium, Fig.~\ref{fig:experiment11}, and one with the electrodes placed in the middle of the aquarium, Fig.~\ref{fig:experiment12}, where the distance to the closest obstacle is twice the maximum distance between electrodes. In both experiments the electrodes are placed at the same depth. We used a period-two behavior at $\alpha=3.1$, mapped to the output voltage of -4.7V ... +4.4V. The measured level of noise is shown in Fig.~\ref{fig:experiment131}. The maximum amplitude of noise varied between +0.03V and -0.05V; we observed a smaller level of noise (+0.01V and -0.02V) in the middle of the aquarium and a bias of the average value to negative (around -0.007V). This bias arises from the non-symmetrical amplification of the signal from the electrodes.
\begin{figure}[h!]
\centering
\subfigure[\label{fig:Exp1Diagr1}]{\includegraphics[width=0.245\textwidth]{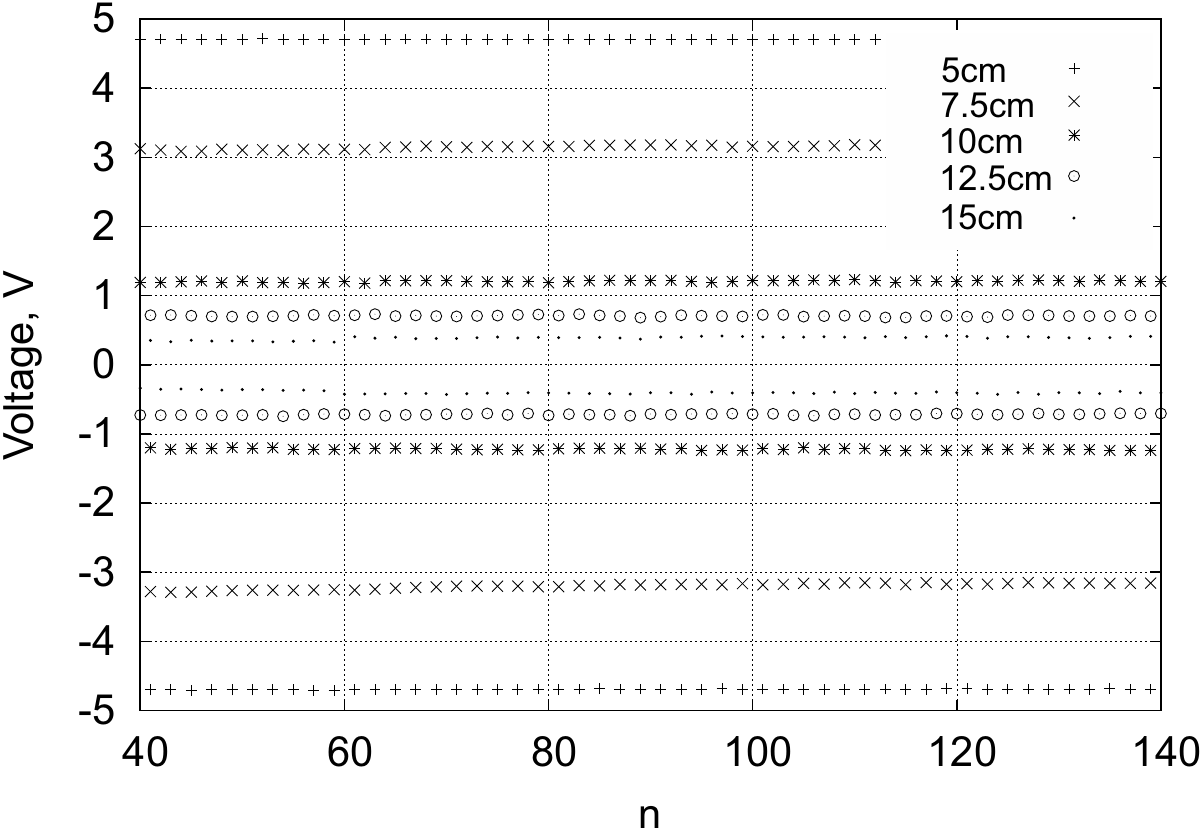}}~
\subfigure[\label{fig:Exp1Diagr2}]{\includegraphics[width=0.245\textwidth]{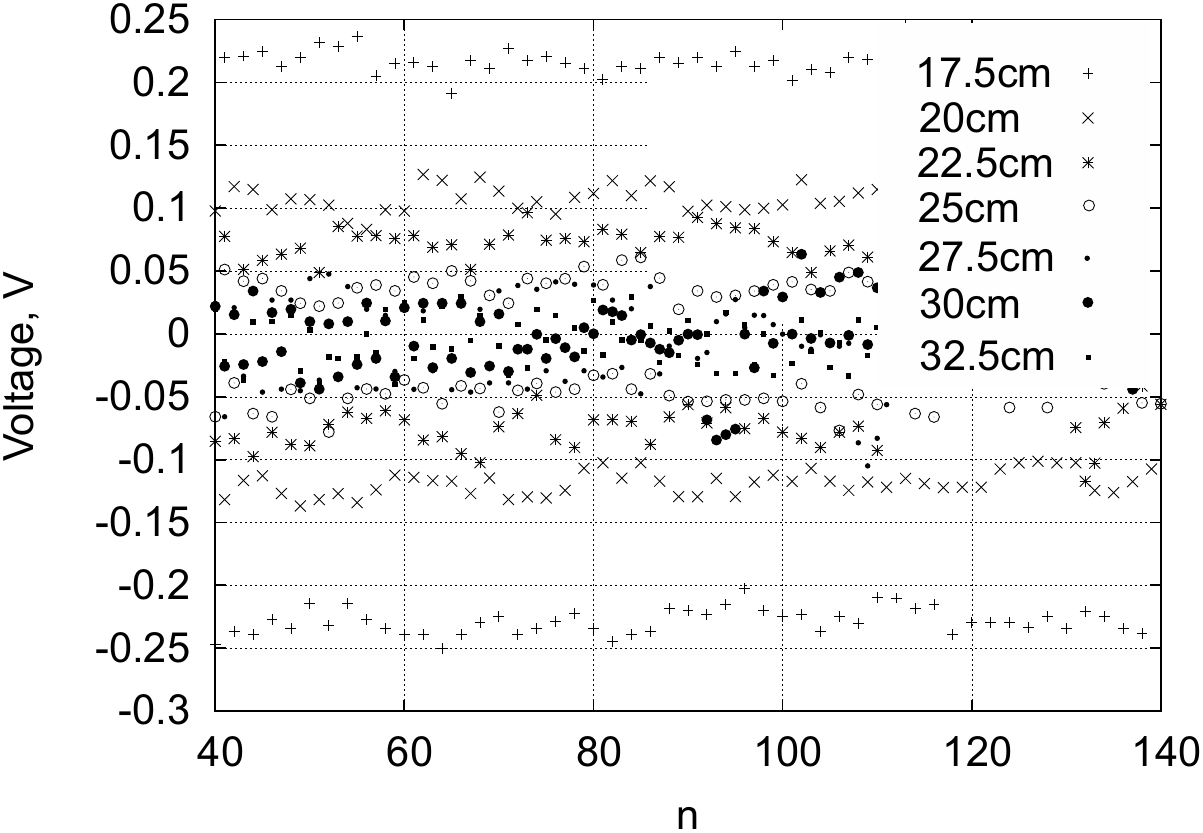}}
\caption{\small Experimental data for the first position of electrodes for d=5cm,...,32.5cm.\label{fig:experimentalDataP1E1}}
\end{figure}
Fig.~\ref{fig:experimentalDataP1E1} shows a plot of experimental data for the first position of electrodes. Due to greater noise, we considered this as the worst case for communication. The distance between electrodes was increased from 5cm to 32.5cm in steps of 2.5cm. At $d<5cm$, the input voltage is around the saturation level of the amplifier. Values up to 25cm are above the noise level. The voltage values for $d>25cm$ are within the noise areas and need statistical processing. Thus, d=5,..., 25cm can be considered as the working range of the system. At the distance between emitting electrodes of 7.5cm, the communication distance is about 3.3 body lengths in the worst case for potential mode.

Fig.~\ref{fig:meanBothCases} shows the decay of periodic signals for all measured values of $d$ for positions 1 and 2. We observe a larger variation in the signal for position 1, moreover signal decay is higher in the middle of aquarium.
\begin{figure}[ht]
\centering
\subfigure[\label{fig:experiment131}]{\includegraphics[width=0.245\textwidth]{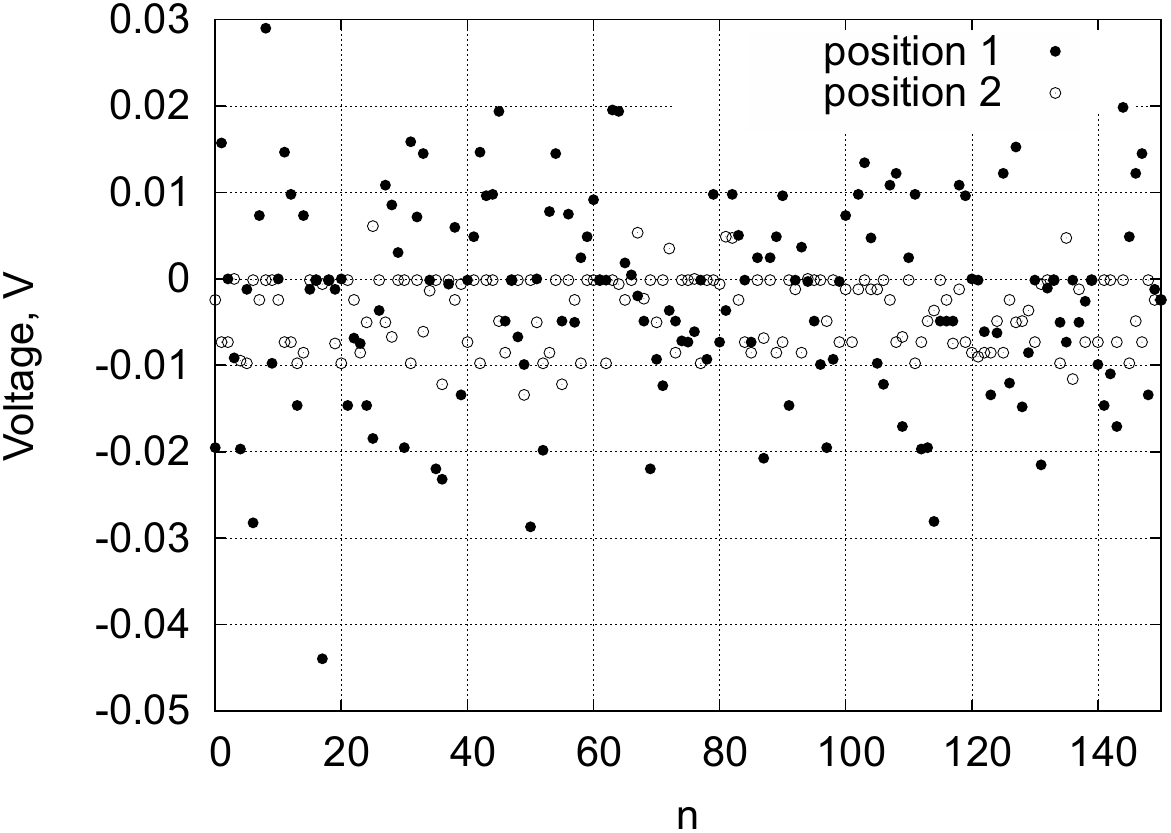}}~
\subfigure[\label{fig:experiment132}]{\includegraphics[width=0.245\textwidth]{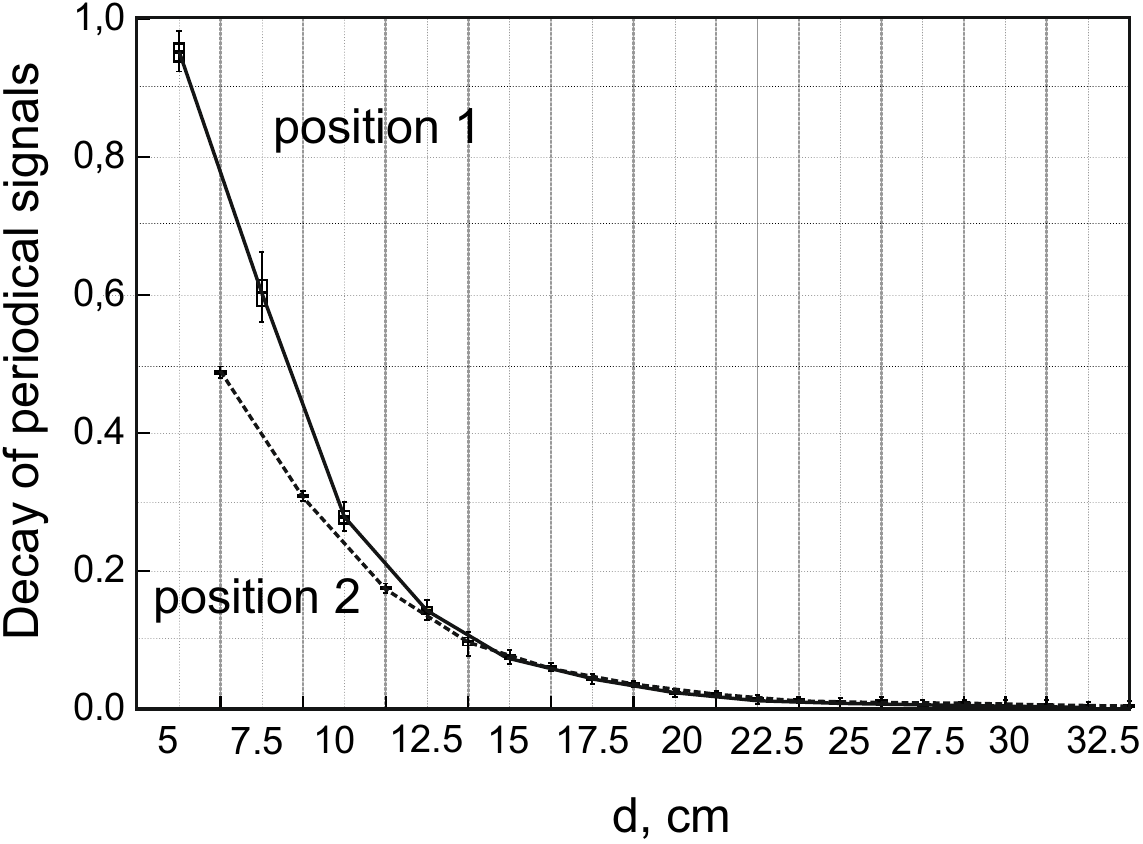}}
\caption{\small \textbf{(a)} Noise measured for the first and second position of the electrodes; \textbf{(b)} Signal decay for distances between 5cm and 32.5cm for the first and second positions of the emitting devices.\label{fig:meanBothCases}}
\end{figure}

\subsection{Part 2: Calibration of The Coefficient $k_1$}
\label{sec:exp2}

The need for calibration is explained by the expression (\ref{eq:balancing condition}). The self-signal in the term $\xi_n^o$ should be compensated, so that $\xi_n^w - k_1 \xi_n^o=0$ in the absence of any other emitting devices. In this experiment the position of electrodes is first fixed in relation to the level of water and $k_1$ is calculated for this setup. For later experiments, the device auto-calibrates.
\begin{figure}[ht]
\centering
\subfigure[\label{fig:coeffk1}]{\includegraphics[width=0.245\textwidth]{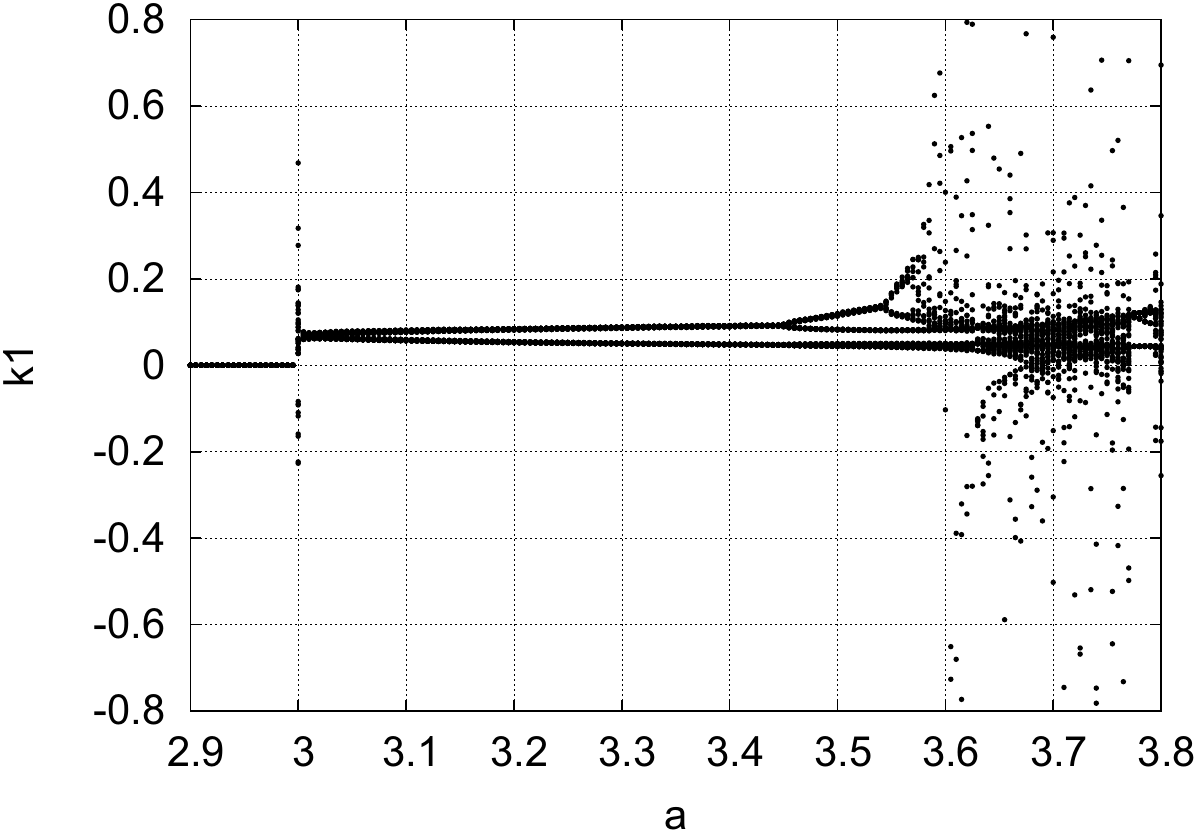}}~
\subfigure[\label{fig:coeffk10}]{\includegraphics[width=0.245\textwidth]{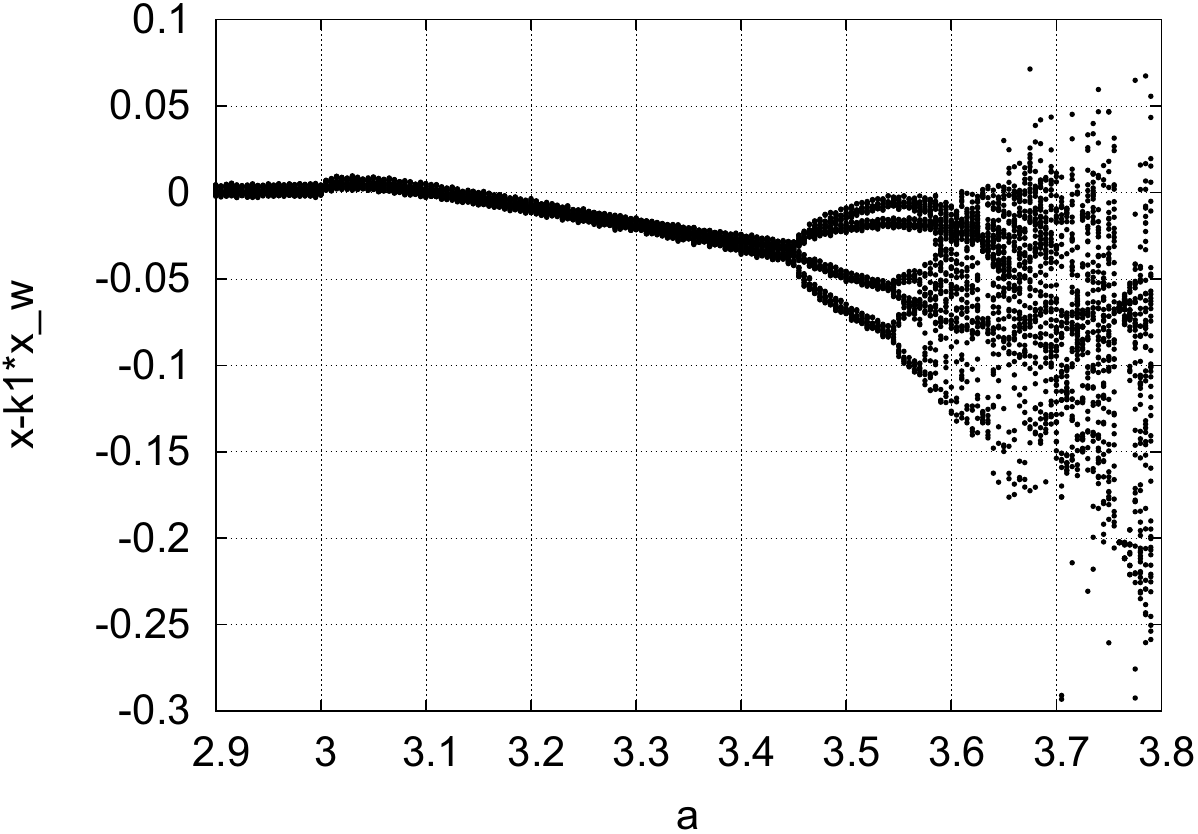}}
\caption{\small \textbf{(a)} Experimental estimation of the coefficient $k_1$, calculated as $\frac{\protect\xi_n^w}{\protect\xi_n^o}$ in the microcontroller, $k_{DAC}=2000*0.175$; \textbf{(b)} The term $\protect\xi_n^w - k_1 \protect\xi_n^o$.}
\end{figure}

In Fig.~\ref{fig:coeffk1} we plot $\frac{\protect\xi_n^w}{\protect\xi_n^o}$, which is calculated directly in the microcontroller. We observe different values of $k_1$ in Fig.~\ref{fig:coeffk1} for each type of behavior. For the period-two motion of the system (\ref{eq:coupledSystem}) at $\alpha=3.1$ we find an oscillating behavior of $\frac{\protect\xi_n^w}{\protect\xi_n^o}$ between $k_1^h=0.13$ for positive and $k_1^l=0.03$ for negative values of the $x_n$. This difference can be explained by nonsymmetrical gain in the operational amplifier. Thus, in the expression (\ref{eq:balancing condition}) we must consider the sign of $x_n$. Iteration from such initial conditions, close to the value of the long-term dynamics, is recommended. Moreover, it depends also on the delay between sending a value to the DAC and reading the voltage from the ADC. Fig.~\ref{fig:coeffk10} shows the values of $\xi_n^w - k_1 \protect\xi_n^o$ with the calculated coefficient $k_1$ for $\alpha=3.1$. We observe the appearance of structures for $\alpha>3.4$, which will change the dynamics of original system in this area. Fig.~\ref{fig:compensatedDiagram} demonstrates two bifurcation diagrams with the term $\pm0.5 (\protect\xi_n^w - k_1 \protect\xi_n^o)$ and changes in the high-periodical and chaotic regions of the parameter $\alpha$.
\begin{figure}[ht]
\centering
\subfigure[\label{fig:compensatedDiagram1}]{\includegraphics[width=0.245\textwidth]{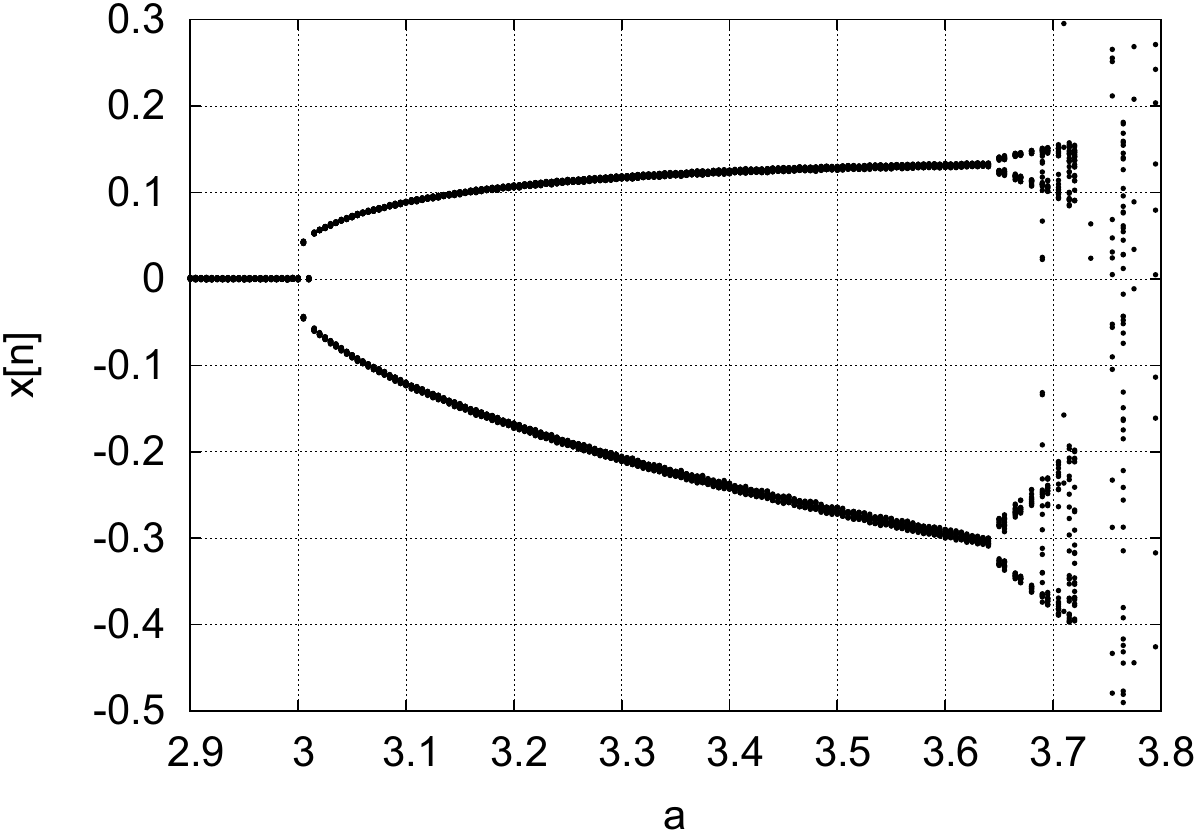}}~
\subfigure[\label{fig:compensatedDiagram2}]{\includegraphics[width=0.245\textwidth]{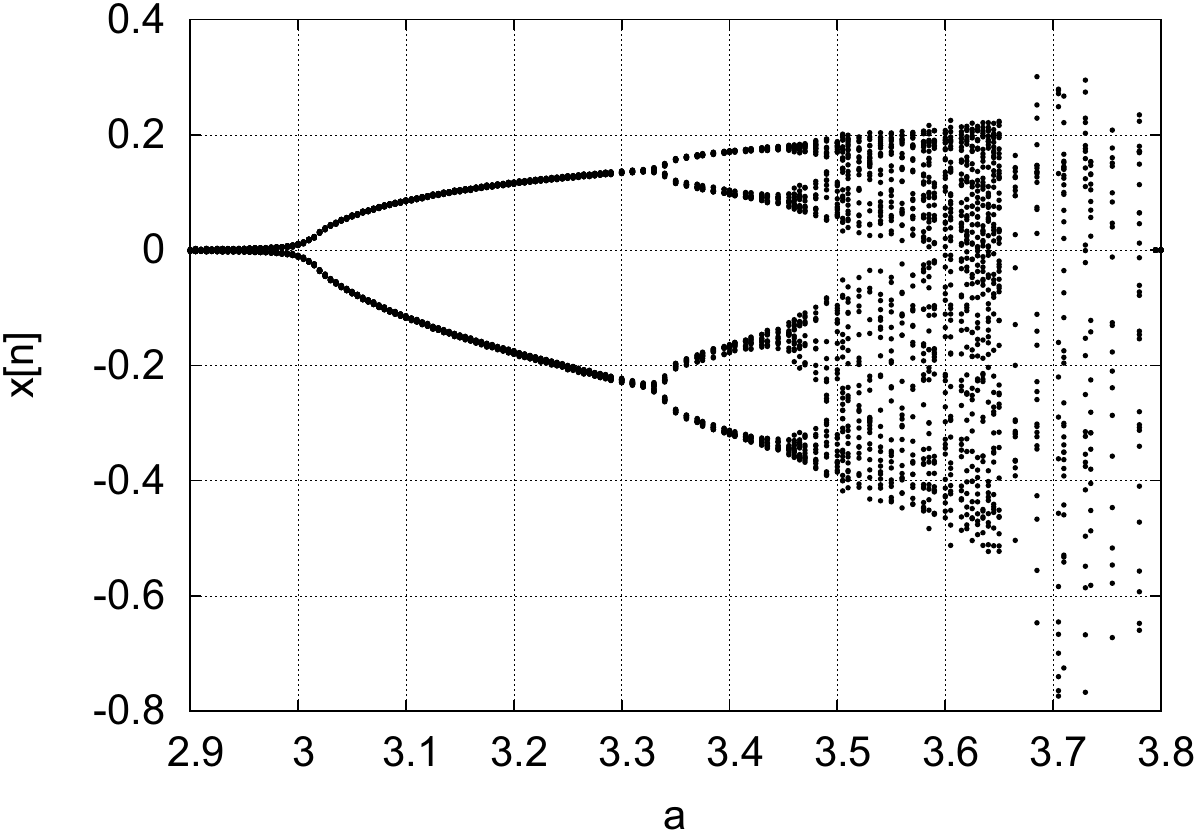}}
\caption{\small Bifurcation diagrams of the system (\ref{eq:coupledSystem}) from the experiment with two emitting/receiving devices, $k_2=\pm0.5$, $\alpha$ is iterated with the interval of $0.005$. \textbf{(a)} $+0.5 (\protect\xi_n^w - k_1 \protect\xi_n^o)$, $x_0=0.01$; \textbf{(b)} $-0.5 (\protect\xi_n^w - k_1 \protect\xi_n^o)$, $x_0=0.1$.  \label{fig:compensatedDiagram}}
\end{figure}

However, for the calculated values of $\alpha=3.1$, we observe a compensation of the self-signal up to the level of noise (the maximum amplitude of this term depends on calibration, it is about $\pm 5 \cdot 10^{-3}$), see Fig.~\ref{fig:coeffk10}. This allows working with even larger values of $k_2$ in further experiments.

There are two strategies for using the coefficient $k_1$. Since it is calculated only for some regions of $x_n$, values outside this region will increase noise and can change the dynamics of the system, see Fig.~\ref{fig:compensatedDiagram2}. Thus, the first strategy is to limit the working region of $x_n$. The second strategy is to calibrate $k_1$ each time with values taken from, for example, a look-up table. In further experiments the coefficient $k_1$ is auto-calibrated each time when the device is switched on.

\subsection{Part 3: Coupled Oscillators Mode}
\label{sec:exp3}

For tests of the coupled oscillator mode, all devices are programmed with the same program and $x_n$ values are read from them. We performed three experiments: \textbf{(1)} synchronization and desynchronization of oscillations, \textbf{(2)} different distances between oscillating devices, \textbf{(3)} different numbers of oscillating devices. The experimental setup for these experiments is shown in Fig.~\ref{fig:setup3Devices}. All three devices $D_1$-$D_3$ were placed either in a triangle or in a line at distance of $d_{12}$, $d_{23}$, $d_{13}$ from each other.
\begin{figure}[ht]
\centering
\subfigure{\includegraphics[width=0.4\textwidth]{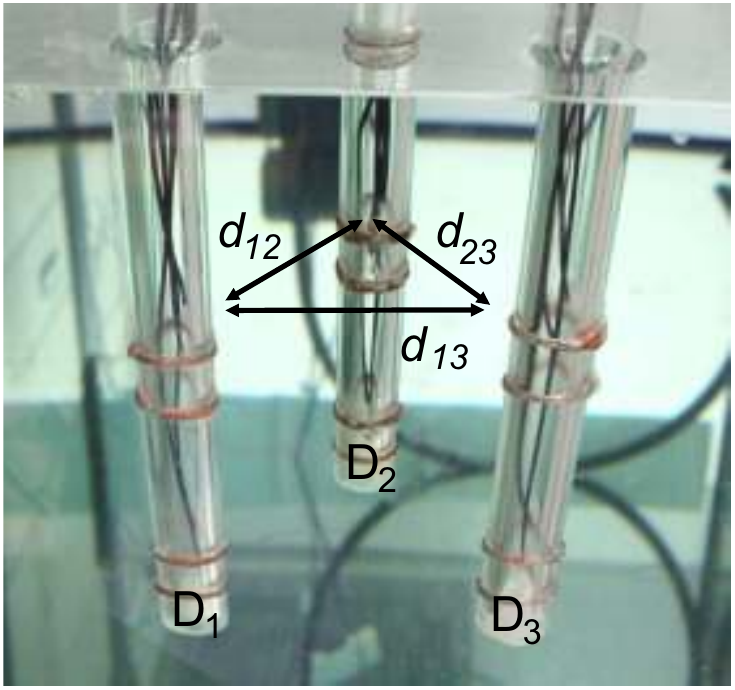}}
\caption{\small Setup for the experiments with coupled oscillators. \label{fig:setup3Devices}}
\end{figure}
Before each experiment, all devices self-calibrate to estimate values of the coefficient $k_1$.

\subsubsection{Synchronization and Desynchronization of Oscillations}
\label{sec:desynchronization}

A desynchronization effect appears for two reasons: first, not all the main clocks of the devices have exactly the same frequency; second, the positive and negative pulses initially overlap, since the oscillators start asynchronously. Potentially, desynchronization of coupled oscillators is undesirable, however it can lead to several interesting effects. In Fig.~\ref{fig:Desynchro} we show two oscillograms of three devices, running Eq.(\ref{eq:coupledSystem}) at $\alpha=3.1$ and pulses of duration $2$ms.
\begin{figure}[h!]
\centering
\subfigure[\label{fig:Desynchro2}]{\includegraphics[width=0.245\textwidth]{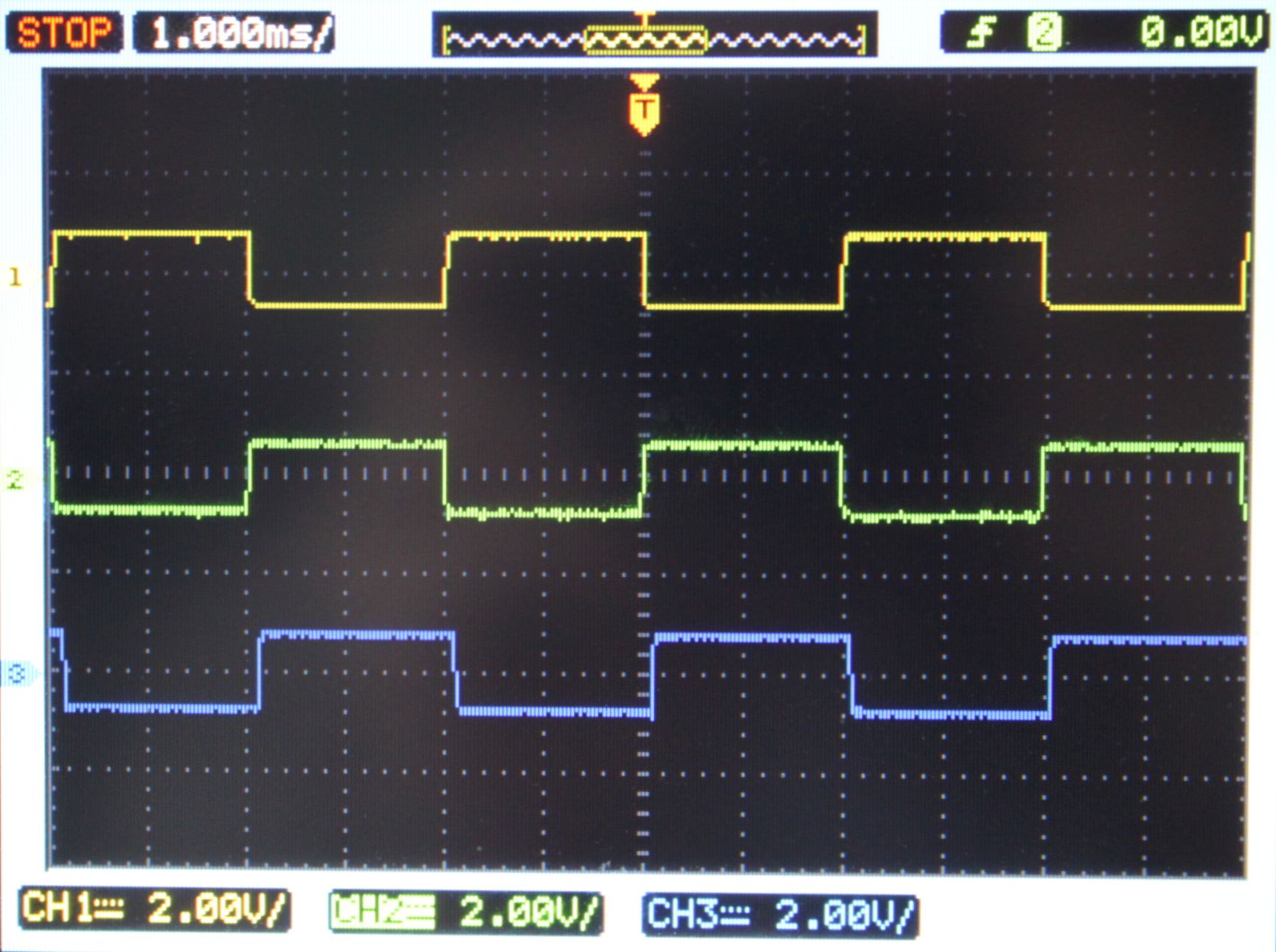}}~
\subfigure[\label{fig:Desynchro3}]{\includegraphics[width=0.245\textwidth]{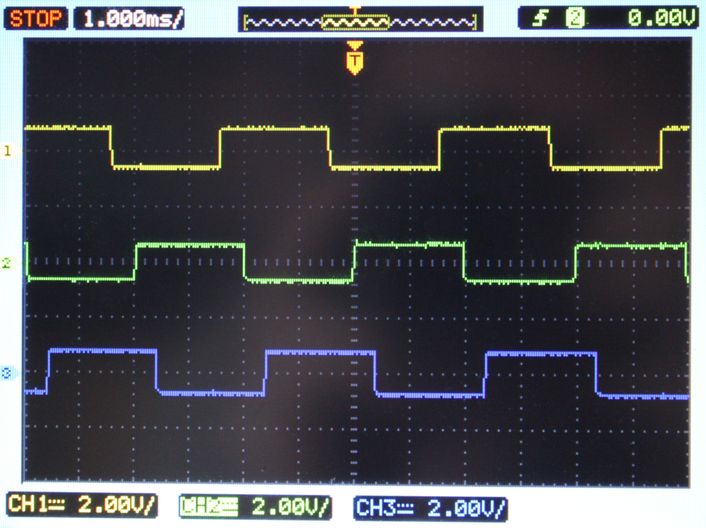}}~
\caption{\small Different stages of phase desynchronization between emitting devices. \label{fig:Desynchro}}
\end{figure}
We changed the clock divider of each device to exhibit an approximately $0.2$Hz shift of frequencies. Thus, the periods of oscillations in different devices can exactly overlap, can have inverse phases, see Fig.~\ref{fig:Desynchro2}, and can be shifted differently from each other, see Fig.~\ref{fig:Desynchro3}.

To confirm the model (\ref{eq:coupl121}) in Sec.~\ref{sec:linCoupledModeZero} and our ideas regarding adding and subtracting amplitudes in asynchronous mode, two devices ran Eq.(\ref{eq:coupledSystem}) with $k_2=0$ (that is, without reading values from the water). The third device reads the ADC and measures the intensity of electric field without emitting functionality. We performed three measurements, shown in Fig.~\ref{fig:measurementField1}.
\begin{figure}[h!]
\centering
\subfigure[\label{fig:measurementField1}]{\includegraphics[width=0.245\textwidth]{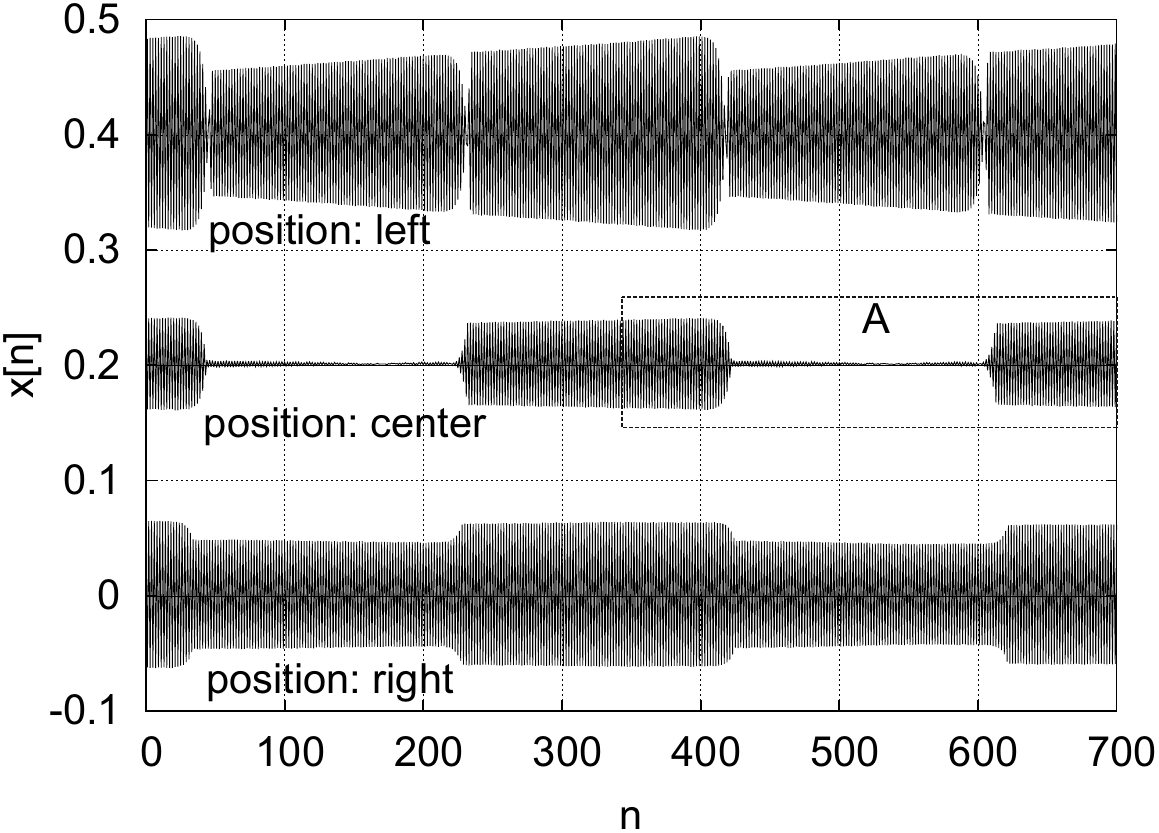}}~
\subfigure[\label{fig:measurementField2}]{\includegraphics[width=0.245\textwidth]{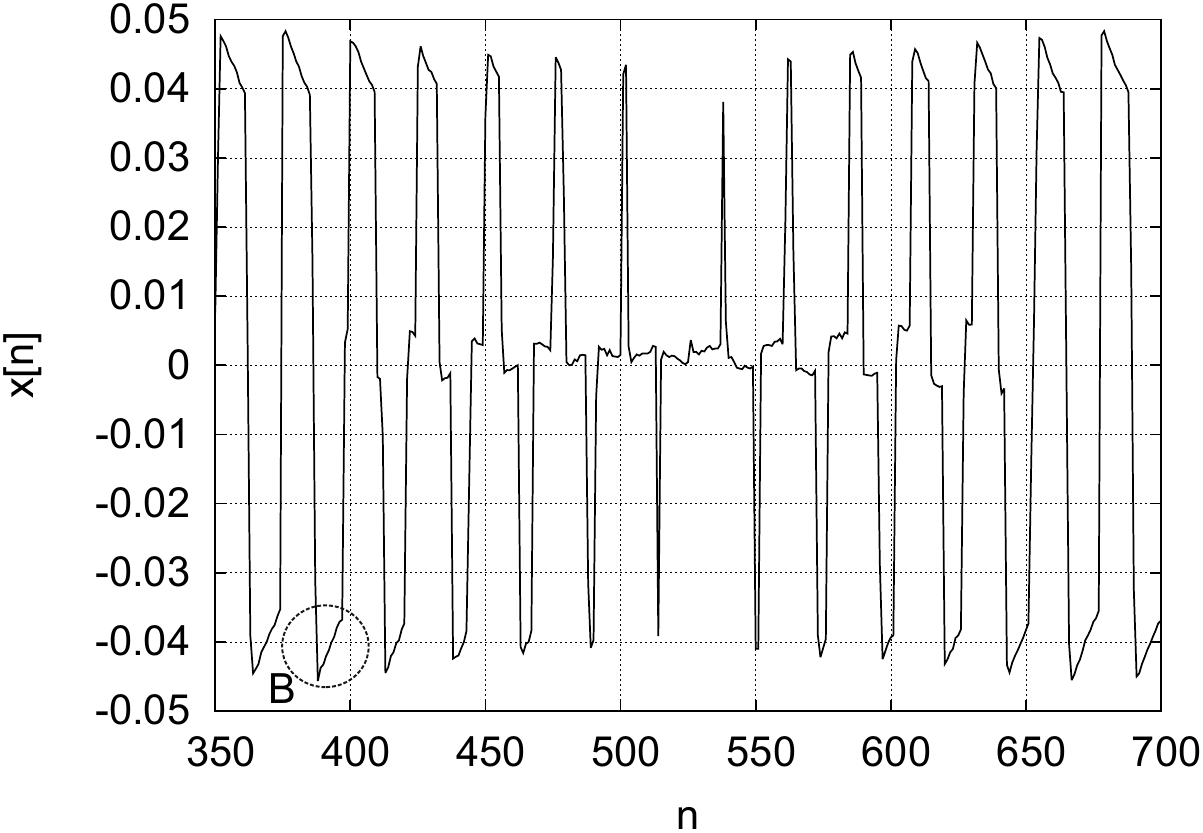}}
\caption{\small Measuring intensity of electrical field, both emitting devices $D_1$ and $D_3$ run Eq.~(\ref{eq:coupledSystem}), $\alpha=3.1$, $k_2=0$, $d_{13}=15cm$. \textbf{(a)} 'position:left' -- the measuring device is close to the left-hand emitting device (0.4 is added to all values), 'position:center' -- the position of measuring device is between the two emitting devices (0.2 is added to all values), 'position:right' -- the measuring device is close to the right-hand emitting device. The sampling frequency of the measuring device is 0.5kHz; \textbf{(b)} Region 'A' with a higher sampling frequency of 20kHz. \label{fig:measurementField}}
\end{figure}
As expected, a spatially-dependent adding and substraction of phases is observed. Due to a low sampling frequency in Fig.~\ref{fig:measurementField1} we see a step-wise phase exchange. Fig.~\ref{fig:measurementField2} shows the region 'A' with a higher sampling rate, where adding and subtracting of phases is readily visible. The decrease of potential (the region 'B') in Fig.~\ref{fig:measurementField2} is explained by the high-pass filter. Another interesting effect is also visible in Fig.~\ref{fig:measurementField1}; the left and right emitting devices have slightly different waveforms, resulting in the symmetry breaking when approaching the left or right sides of aquarium. Based on this effect, a robot can distinguish between emitting devices and thus navigate.

For $k_2\neq0$ the phase desynchronization appears as a switch between two amplitudes of period-two motion. To make this effect more visible, we increase the duration of pulses to 10ms and plot the values $\downarrow{\xi^i_n}$ from the water for two oscillating devices, as shown in Fig.~\ref{fig:deSynchro10ms}.
\begin{figure}[ht]
\centering
\subfigure[\label{fig:desynchro10ms1}]{\includegraphics[width=0.245\textwidth]{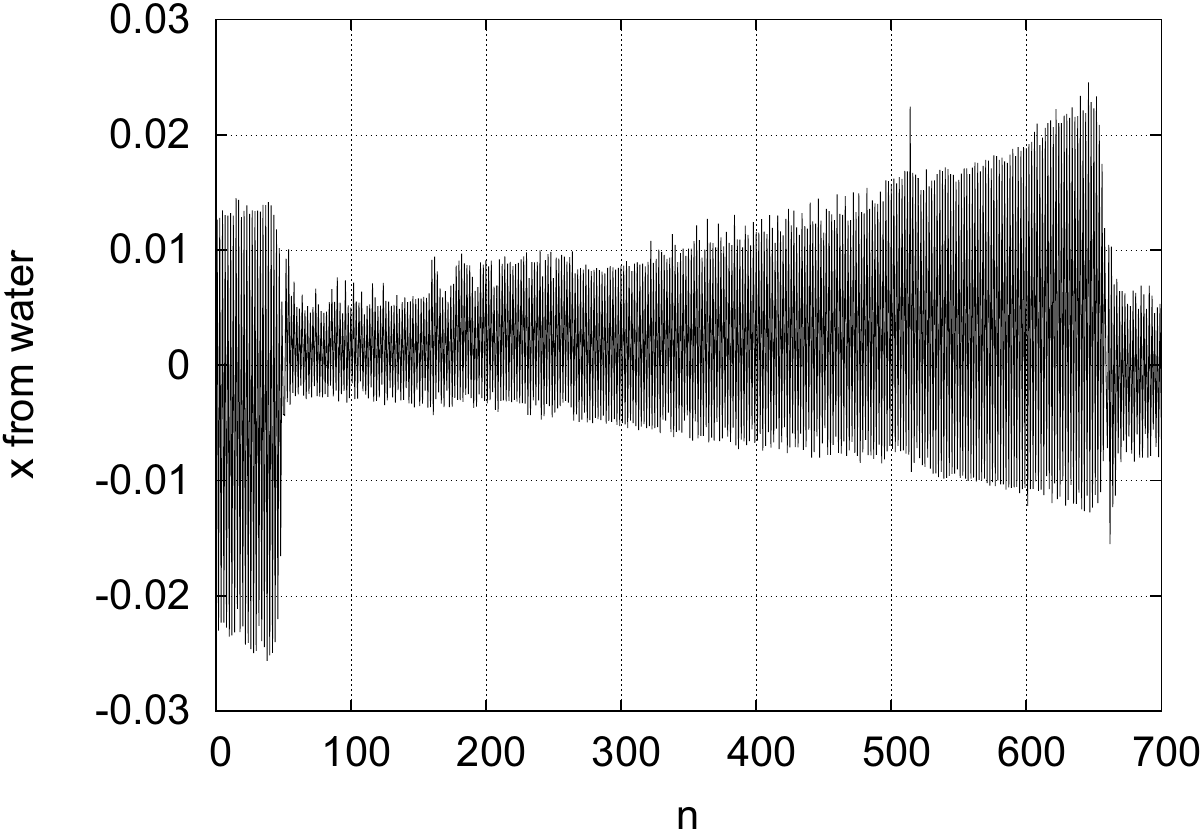}}~
\subfigure[\label{fig:desynchro10ms2}]{\includegraphics[width=0.245\textwidth]{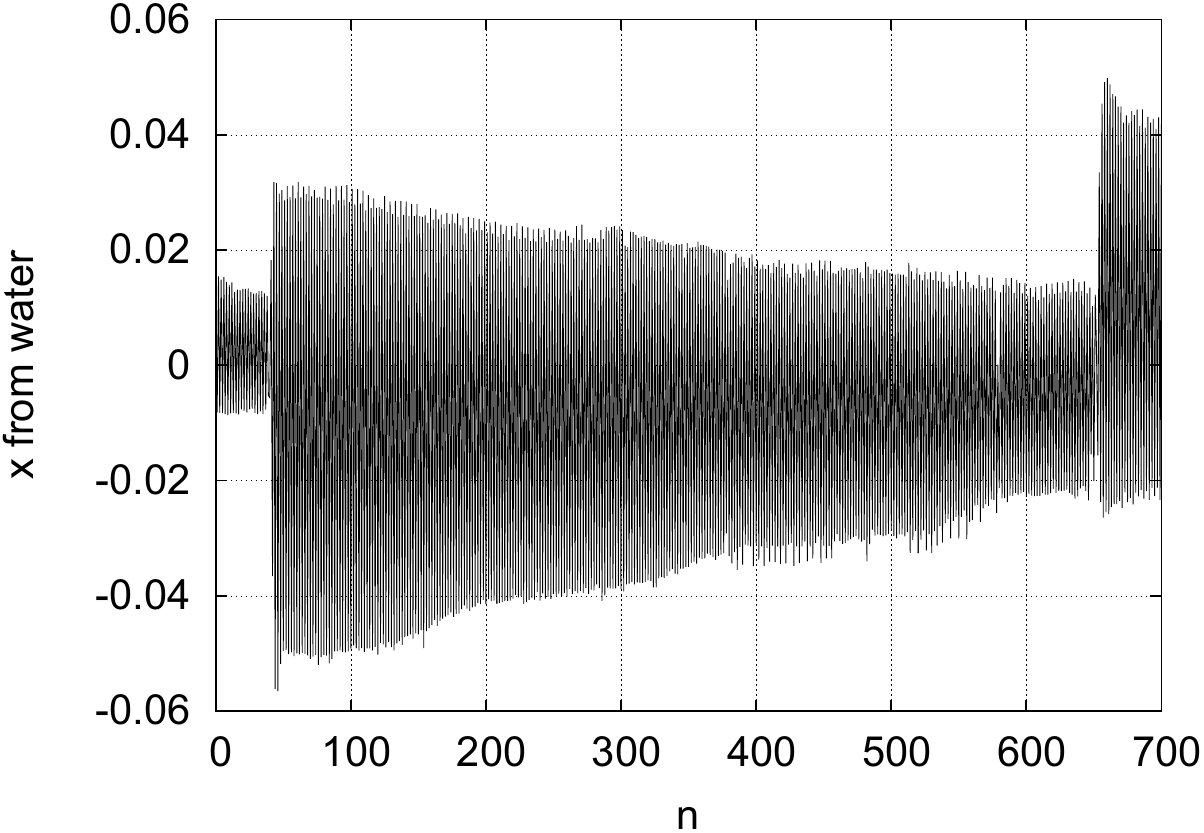}}
\caption{\small Desynchronization effects at $\alpha=3.1$, $k_2=-0.1$, $x_0^i=0.1$, $k_1$ is autocalibrated before experiments, $d=15cm$, duration of pulses $10ms$. \textbf{(a)} Device $D_1$; \textbf{(b)} Device $D_2$; \label{fig:deSynchro10ms}}
\end{figure}
Increasing of amplitudes in the first device, and at the same time, decreasing of amplitudes in the second device are clearly visible.

\subsubsection{Different Distances Between Oscillating Devices}

To measure distances between emitting devices, we can apply two approaches: first, one based on a phase desynchronization for $k_2=0$ and second, one based on measuring the amplitude of $x_n$, as described in Sec.~\ref{sec:linCoupledMode}. Both approaches worked well in the test environment. Fig.~\ref{fig:waterSynchro} shows the difference between the \emph{max} and \emph{min} of the amplitudes for $k_2=0$.
\begin{figure}[ht]
\centering
\subfigure{\includegraphics[width=0.49\textwidth]{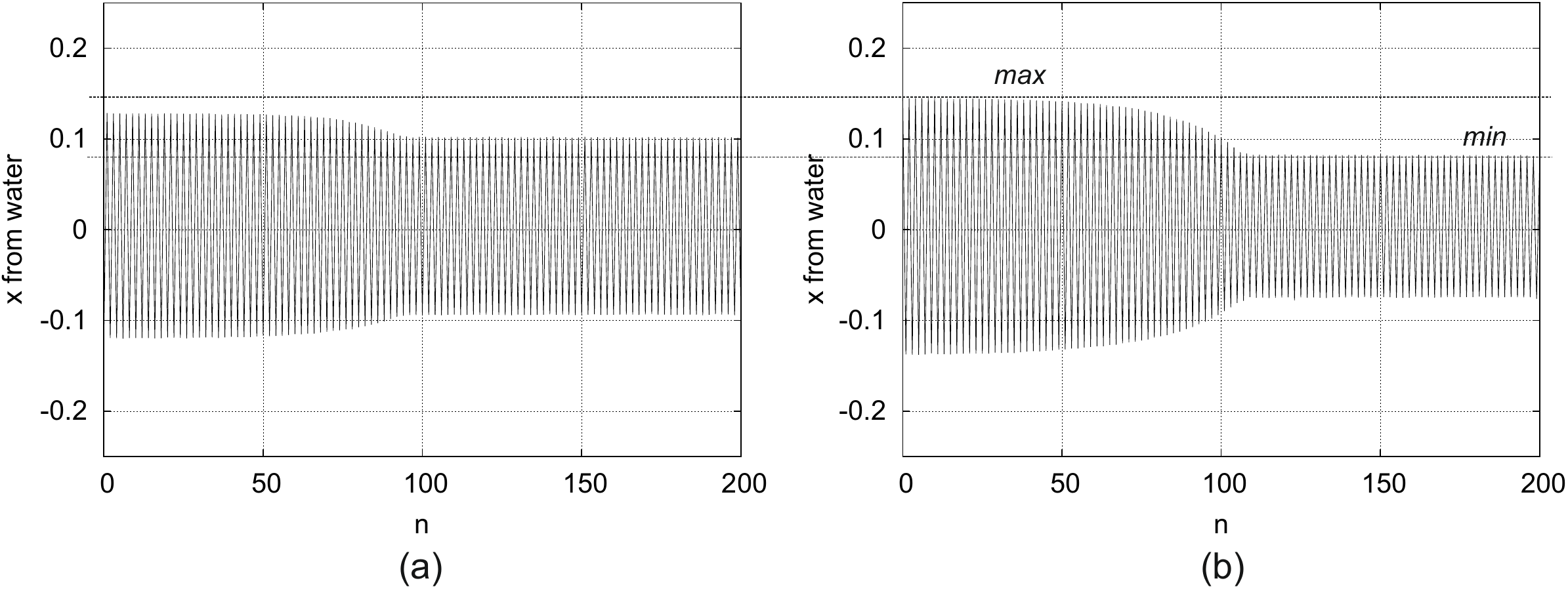}}
\caption{\small Measuring distances between emitting devices by measuring \emph{max} and \emph{min} amplitudes, $\alpha=3.1$, $k_2=0$. \textbf{(a)} 10cm between devices; \textbf{(b)} 5cm between devices. \label{fig:waterSynchro}}
\end{figure}
In contrast to the measurement shown in Fig.~\ref{fig:measurementField}, in this experiment we measured the intensity of field simultaneously with the self-emitted signals. The resolution of this approach is primarily defined by measurement accuracy and the amplification factor of input electronics.

To demonstrate the approach with $k_2\neq0$, we used a strong positive coupling coefficient $k_2$, which leads to global synchronization and changes in qualitative behavior. Three experiments for $d_{12}=d_{23}=d_{13}=25$cm (period-two behavior), for $d_{12}=d_{23}=d_{13}=15$ and for $d_{12}=d_{23}=d_{13}=10$cm have been performed, see Fig.~\ref{fig:fullSynchro}.
\begin{figure}[ht]
\centering
\subfigure[\label{fig:compensatedDiagram11}]{\includegraphics[width=0.245\textwidth]{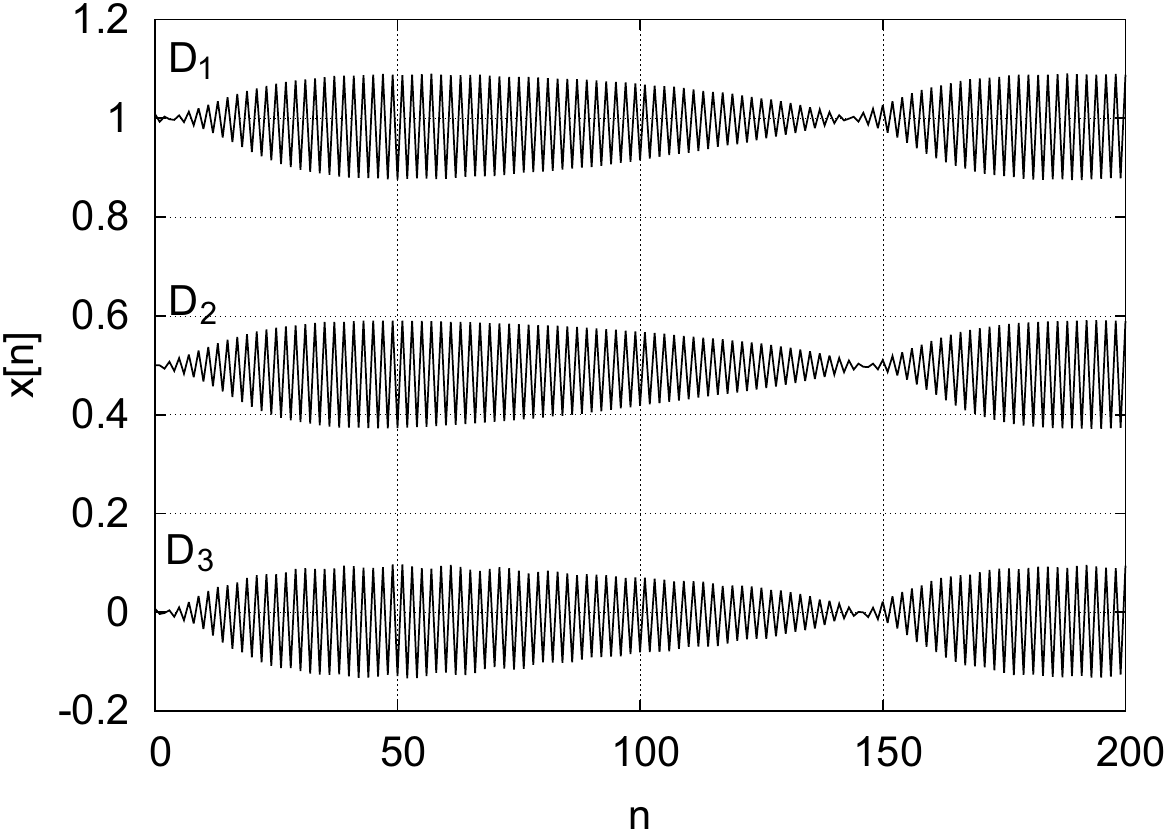}}~
\subfigure[\label{fig:compensatedDiagram21}]{\includegraphics[width=0.245\textwidth]{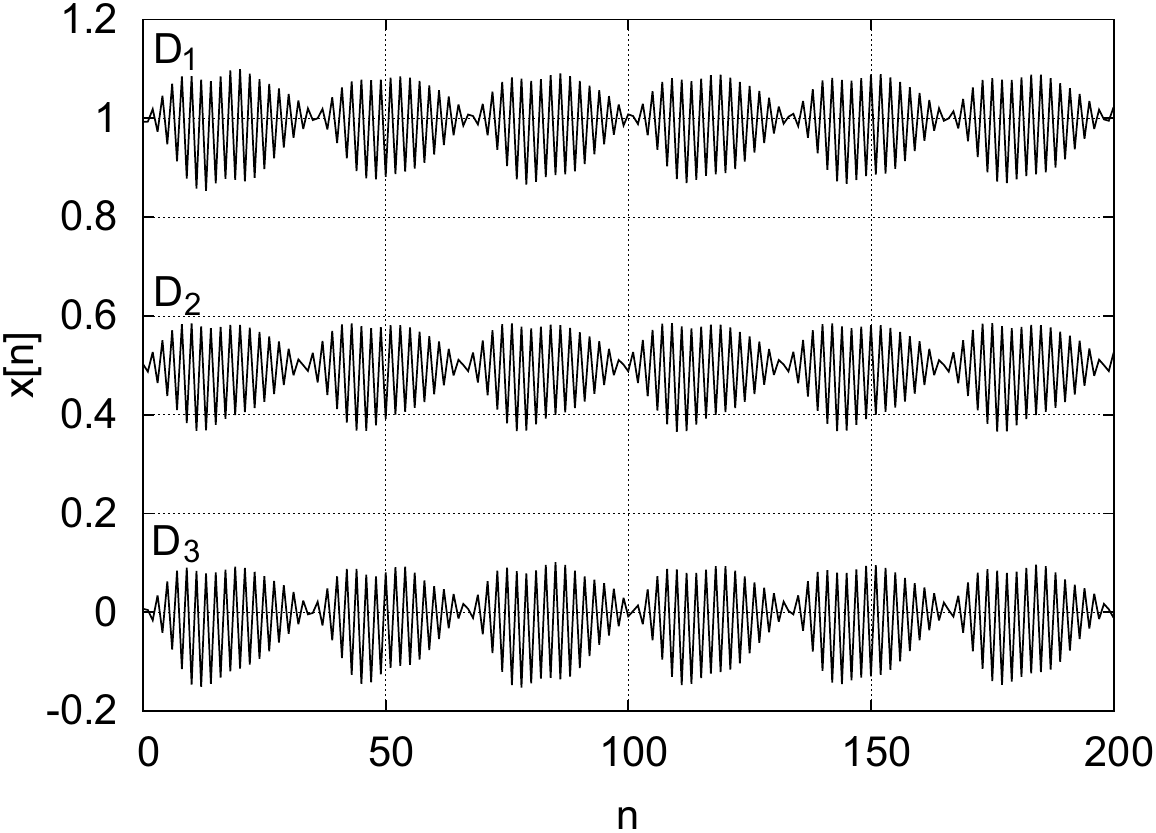}}
\caption{\small Global synchronization effects, $\alpha=3.1$, $k_2=0.7$, $x_0^i=0.1$, $k_1$ is autocalibrated before each experiment; 0.5 is added to values from $D_2$, 1.0 is added to values from $D_1$. \textbf{(a)} Behavior of coupled oscillators for $d_{12}=d_{23}=d_{13}=15$; \textbf{(b)} Behavior of coupled oscillators for $d_{12}=d_{23}=d_{13}=10$. \label{fig:fullSynchro}}
\end{figure}
We clearly see a locking of phases and amplitudes for all three oscillators. The distance between devices can be obtained by measuring the frequency of pulses.

\subsubsection{Different Number of Oscillating Devices}

As indicated in Sec.~\ref{sec:linCoupledMode}, either the number of devices or distances between them can be measured. In this case we were interested in the number of devices and repeat the experiment with a strong positive coupling coefficient $k_2$. However, in this experiment all oscillating devices were placed in a line, at distances of $d_{13}=22cm$ and $d_{12}=d_{23}=11cm$. Moreover, we slightly unbalanced oscillators 1 and 2 by varying the coefficient $k_1$. First two close oscillators 1 and 2 were turned on, and became synchronized, see Fig.~\ref{fig:inLine1}. Oscillators 1 and 3, at a distance of 22cm, were only slightly perturbed in the period-two behavior. After this, we turned on oscillator 3. The behavior of all devices changed to a mode, where synchronization and desynchronization phases exchange, see Fig.~\ref{fig:inLine2}.
\begin{figure}[ht]
\centering
\subfigure[\label{fig:inLine1}]{\includegraphics[width=0.245\textwidth]{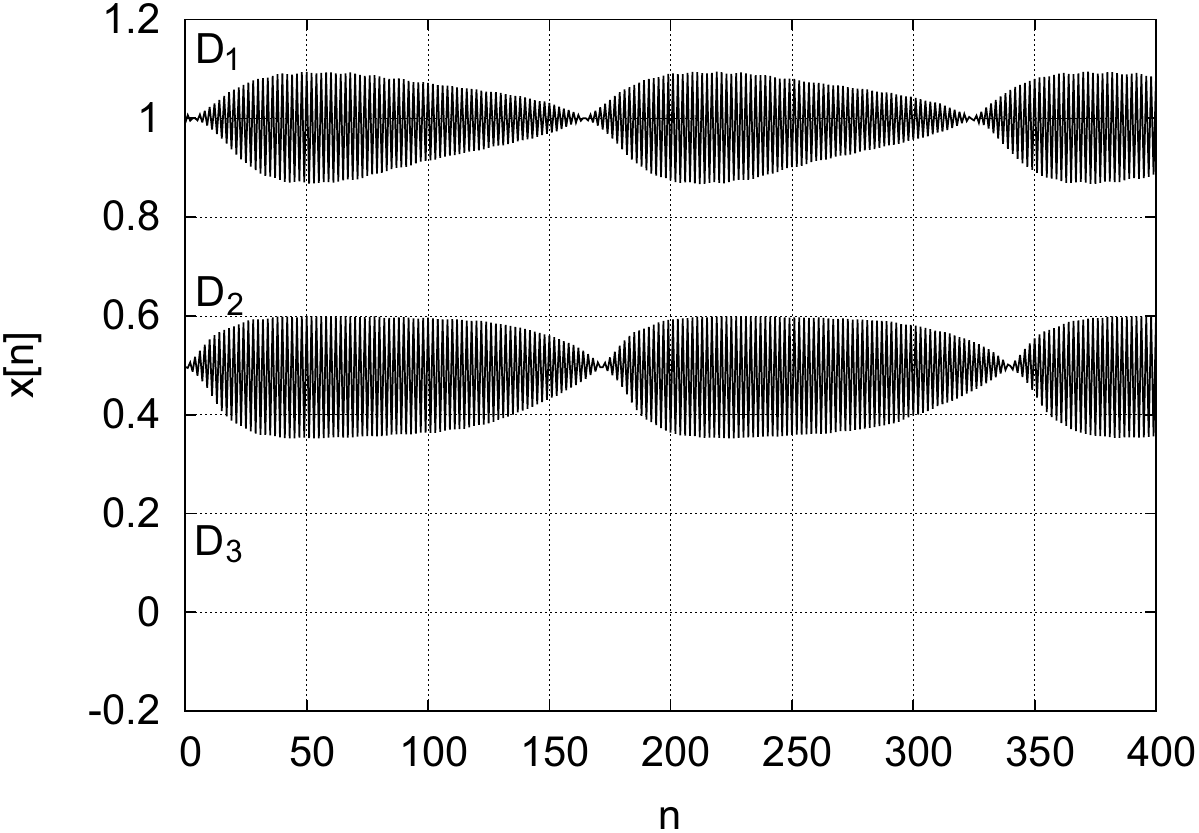}}~
\subfigure[\label{fig:inLine2}]{\includegraphics[width=0.245\textwidth]{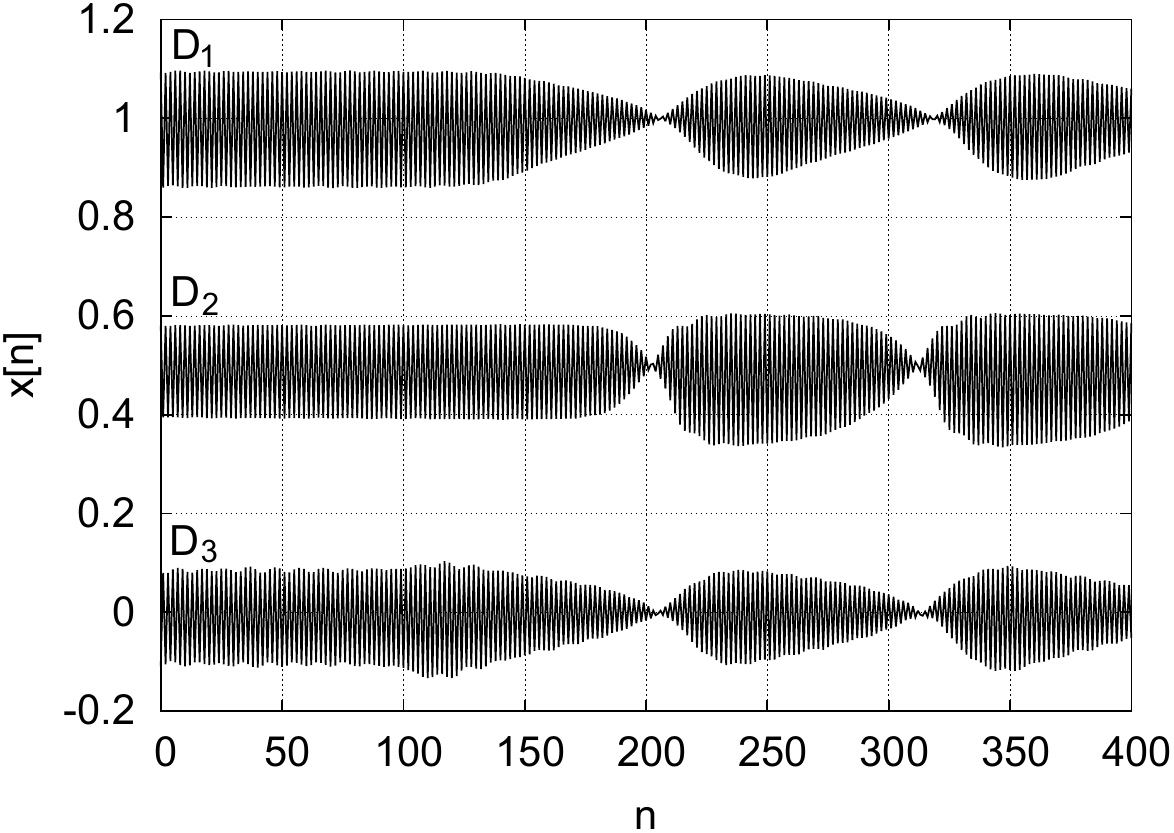}}
\caption{\small Three oscillating devices placed in a line, $d_{13}=22cm$ and $d_{12}=d_{23}=11cm$, $\alpha=3.1$, $k_2=0.75$, $x_0^i=0.1$, $k_1$ is autocalibrated before experiments, $k_{DAC}=1.4$,  0.5 is added to values from $D_2$, 1.0 is added to values from $D_1$. \textbf{(a)} Oscillator 3 is turned off; \textbf{(b)} Oscillator 3 is turned on. \label{fig:InLine}}
\end{figure}

Generally, when all devices have different oscillation parameters, caused for example by self-calibration of $k_1$ or by different distances between devices, we observe a different period of oscillation, as shown in Fig.~\ref{fig:inLine1}. This leads to an appearance of temporal desynchronization, after which all devices resynchronize. This mode can be used for detection of a group of underwater devices, even at the furthest range of sensitivity. Moreover, by measuring the degree of desynchronization, it is possible to draw conclusions about the spatial distribution of devices.

\subsection{Part 4: Electrical Mirror Mode}
\label{sec:exp4}

For experiments in electrical mirror mode, one or two devices oscillated and one device represented the electrical mirror implemented as an independent system, see Fig. \ref{fig:mirror}. All devices had the same time interval (2ms) as in other experiments. The mirror device first senses the electrical field without sending its own signal and stores the received values in  internal array. After this, the value received from the previous step (generally the value $x_{n-\gamma}$, where $\gamma$ is the time delay) was sent to the emitting electrodes. After 2ms, the emitting electrodes were switched off, the ADC read the value of electrical field, and the  cycle repeated. For this experiment it is necessary to calibrate: \textbf{(a)} a proportional coefficient between the received and emitted values, first so as not to saturate the amplifiers, and second to provide a possible large interaction radius; \textbf{(b)} time interval when the emitting electrodes are switched on and off.

\begin{figure}[ht]
\centering
\subfigure{\includegraphics[width=0.4\textwidth]{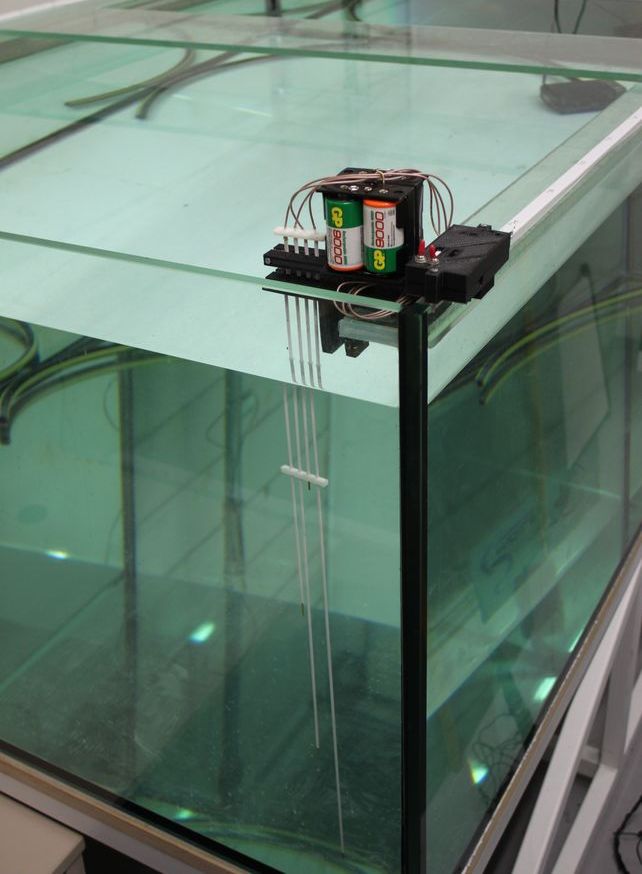}}
\caption{\small Active electric mirror in potential mode. \label{fig:mirror}}
\end{figure}

In the first experiment we intended to confirm the model (\ref{eq:coupledSystemDF}) from Sec.~\ref{sec:DelyedFeedbackMode} regarding time-delayed feedback with two devices (one oscillator and one electrical mirror). Fig.~\ref{fig:delayZ} shows values received from the oscillating device.
\begin{figure}[ht]
\centering
\subfigure[\label{fig:delayZ1}]{\includegraphics[width=0.245\textwidth]{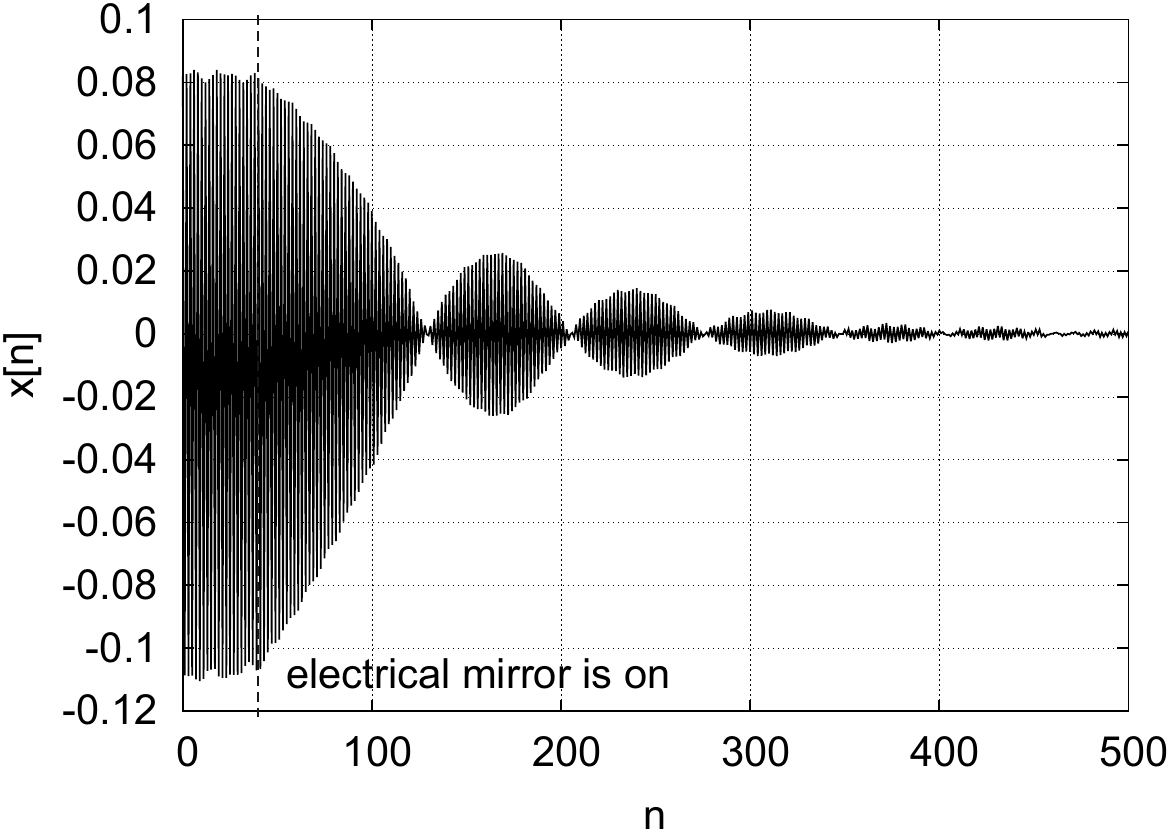}}~
\subfigure[\label{fig:delayZ2}]{\includegraphics[width=0.245\textwidth]{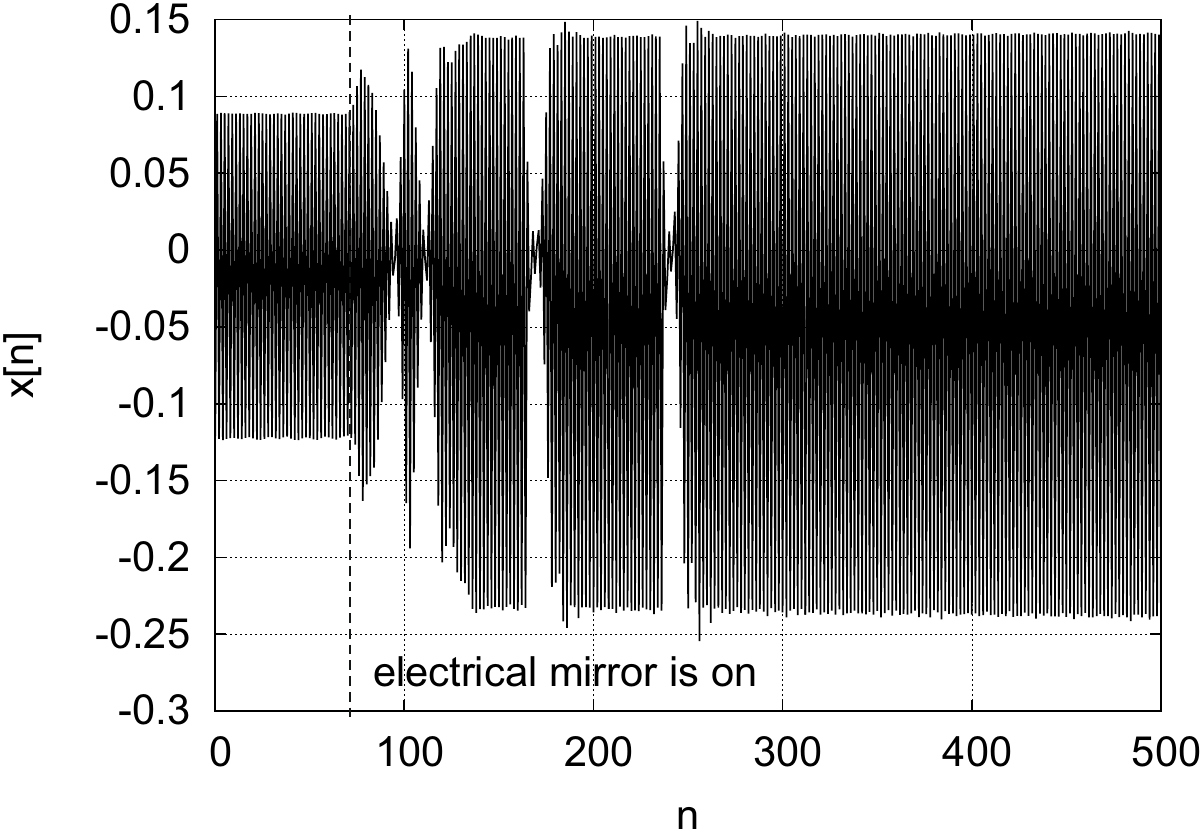}}
\caption{\small Experiment with time-delayed feedback (electrical mirror); the temporal behavior of the oscillating device is shown, $d_{13}=5cm$, $\alpha=3.1$, $x_0^i=0.1$, $k_1$ is autocalibrated before experiments, $k_{DAC}=1.4$. Mirroring device returns $-x_{n-1}$ each time. \textbf{(a)} $k_2=0.5$  \textbf{(b)} $k_2=-0.5$. \label{fig:delayZ}}
\end{figure}
The oscillating device implements Eq.~(\ref{eq:coupledSystem}). The mirroring device returns $-x_{n-1}$ each time, whereas we changed $k_2=0.5$ and $k_2=-0.5$ in the feedback term of the oscillating device. We observed a characteristic behavior in the delayed feedback (see also Fig.~\ref{fig:EigDel1}) when the signal is compensated to zero (as shown in Fig.~\ref{fig:delayZ1}) or when the amplitude of the period-two motion was increased (as shown in Fig.~\ref{fig:delayZ2}) for positive and negative values of the feedback signal respectively. The values of the coefficient $k_3$ in Eq.(\ref{eq:coupledSystemDF}) differ from those shown here, due to amplifications, losses in the water/mirror and other factors; they need to be experimentally calibrated.

In Fig.~\ref{fig:3DevMirror} two devices $D_2$ and $D_3$ oscillate and $D_1$ acts as an electrical mirror. First, $D_2$ and $D_3$ are synchronized, as shown in Fig.~\ref{fig:3DevMirror3}, then the mirror device $D_1$ is turned on; see  Fig.~\ref{fig:3DevMirror4}.
\begin{figure}[ht]
\centering
\subfigure[\label{fig:3DevMirror3}]{\includegraphics[width=0.245\textwidth]{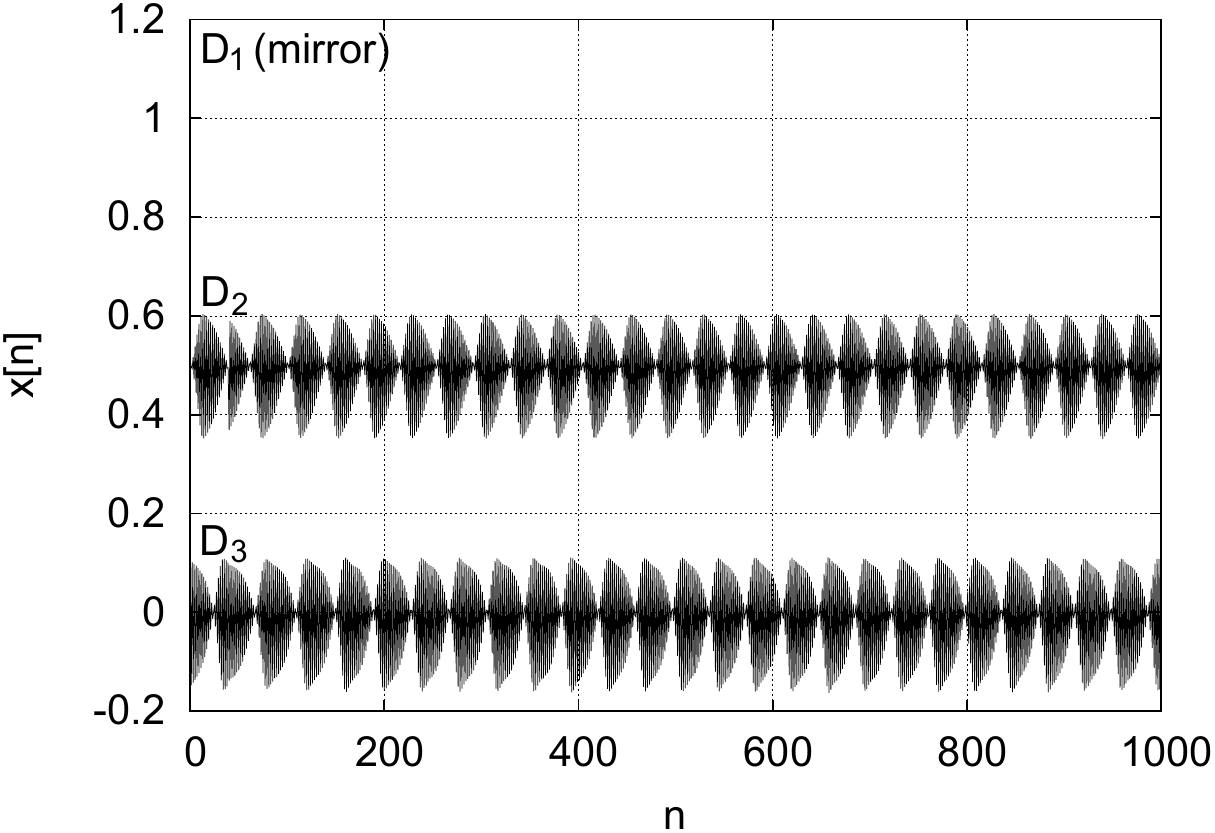}}~
\subfigure[\label{fig:3DevMirror4}]{\includegraphics[width=0.245\textwidth]{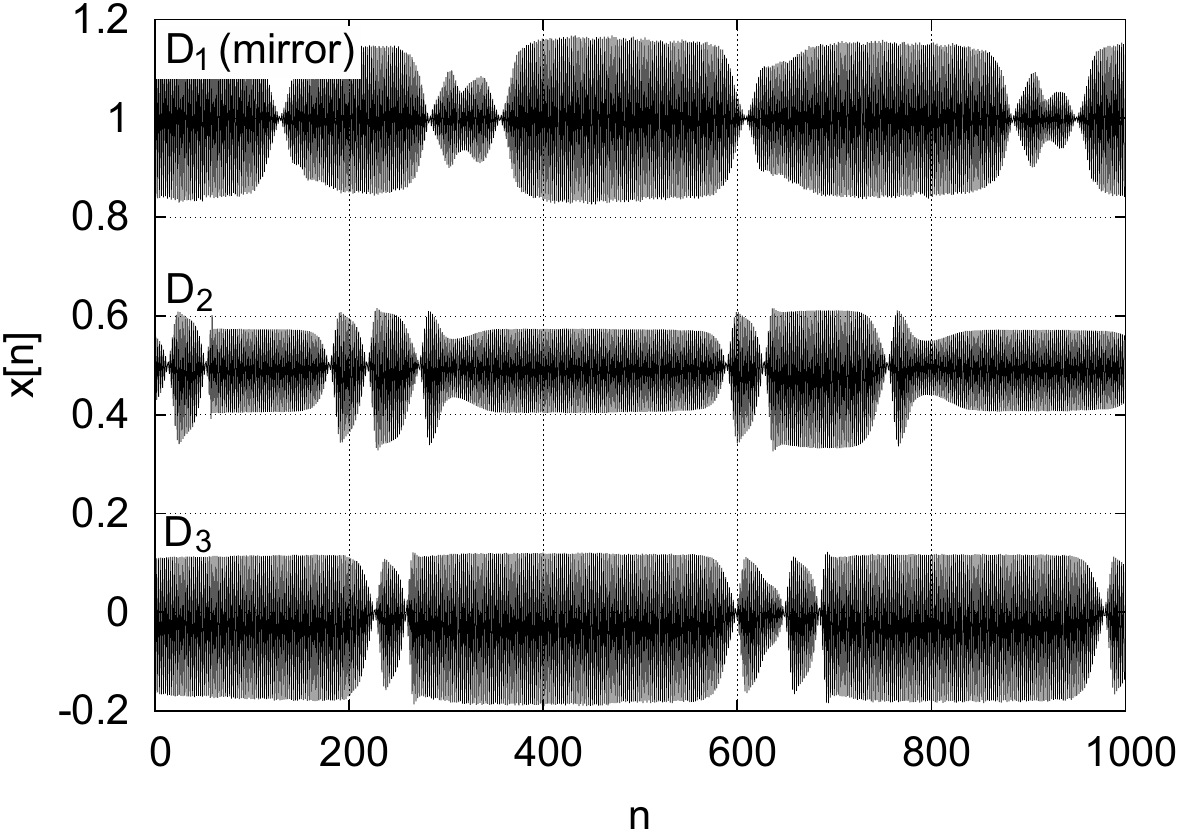}}
\caption{\small Experiment with time-delayed feedback (electrical mirror); the temporal behavior of all devices is shown, $d_{13}=d_{12}=d_{23}=5cm$, $\alpha=3.1$, $x_0^i=0.1$, $k_1$ is autocalibrated before experiments, $k_{DAC}=1.4$, 0.5 is added to values from $D_2$, 1.0 is added to values from $D_1$.
\textbf{(a)} $k_2=-0.2$, no mirror device; \textbf{(b)} $k_2=-0.2$, mirror emits $-x_{n-1}$. \label{fig:3DevMirror}}
\end{figure}
We observed a crash of global synchronization and a change in the oscillating pattern. Comparing the results with Fig.~\ref{fig:fullSynchro}, we see that the global synchronization pattern enables robots to recognize whether the signal is emitted by another swarm or their own signal is reflected by the mirror. 

The system is also sensitive to a non-delayed signal. In Fig.~\ref{fig:Non-delay} we demonstrate the case where the mirroring device returns $-x_{n}$ and the oscillating device has the coefficient $k_2=0.5$.
\begin{figure}[ht]
\centering
\subfigure[\label{fig:Non-delay2}]{\includegraphics[width=0.245\textwidth]{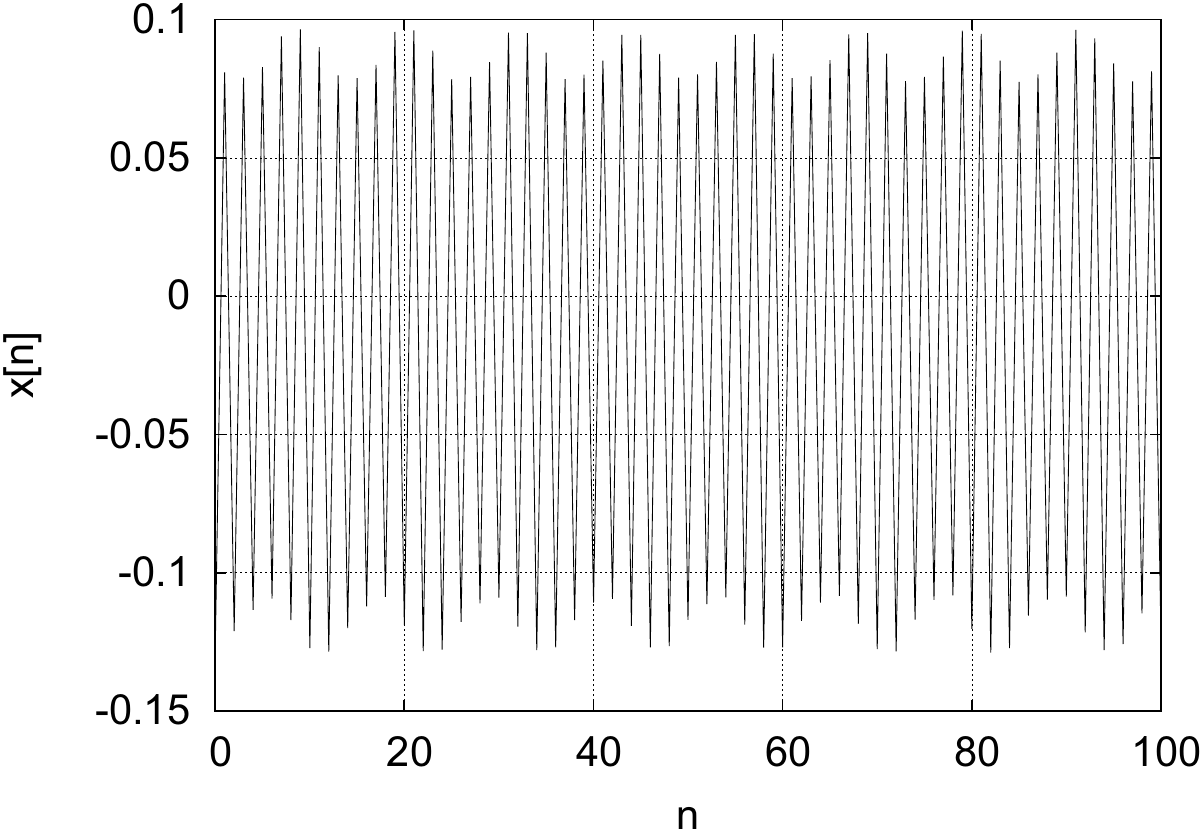}}~
\subfigure[\label{fig:Non-delay5}]{\includegraphics[width=0.245\textwidth]{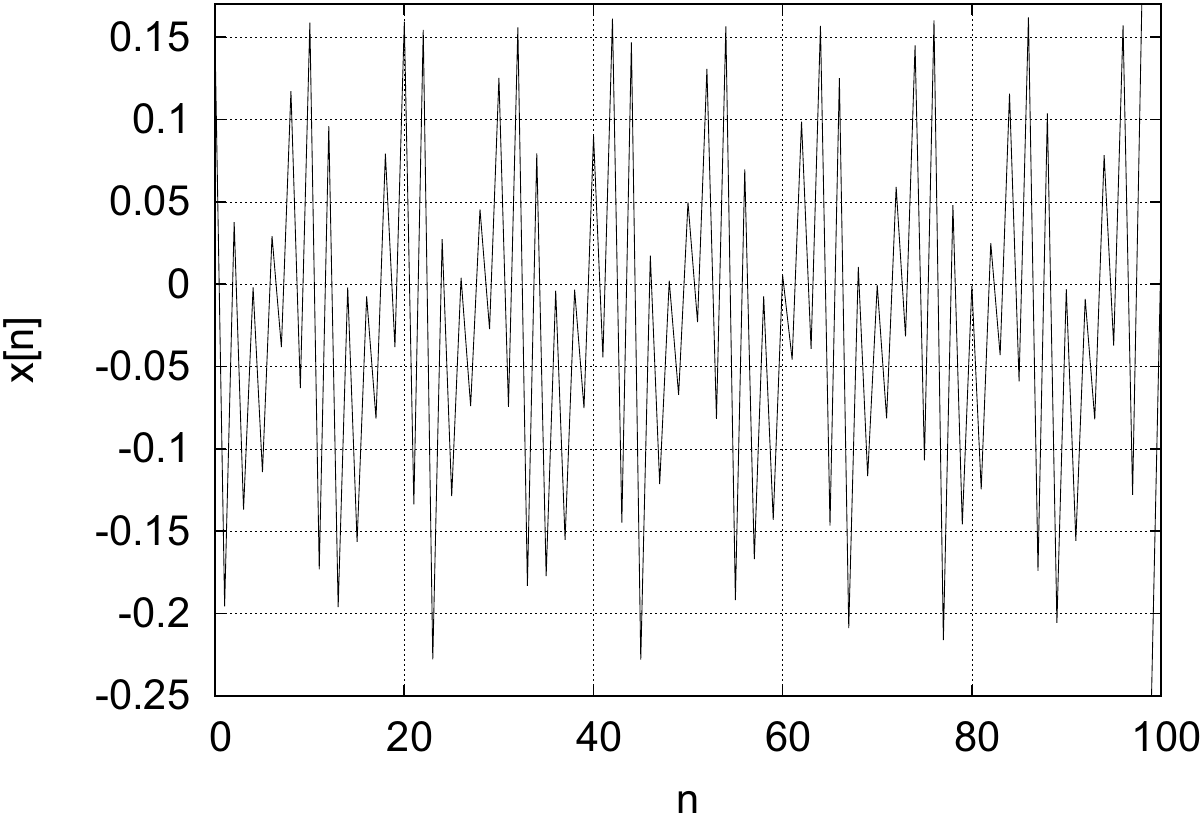}}
\caption{\small Experiment with feedback (electrical mirror); the temporal behavior of the oscillating device is shown, $d_{13}=22cm$ and $d_{12}=d_{23}=11cm$, $\alpha=3.1$, $k_2=0.5$, $x_0^i=0.1$, $k_1$ is autocalibrated before experiments, $k_{DAC}=1.4$. \textbf{(a)} $d_{12}=20cm$; \textbf{(b)} $d_{12}=5cm$. \label{fig:Non-delay}}
\end{figure}
Oscillations are sensitive to the presence of this feedback signal even at the boundary of the sensing range.

\section{Part 5: Current mode sensor}
\label{sec:currentModeSensor}

Experiments in the current mode are aimed at demonstrating coupling by physical means (by interferences of electric fields in water), and sensitivity of common electric fields to placement and movement of dipoles and dielectric objects in such a common field. General setup is shown in Fig. \ref{fig:experimentEIS} and include 8 electric dipoles combined in two groups on movable holders. Each dipole has own fixed oscillating frequency between 70 and 450 Hz. Since the coupling part of Eq.~(\ref{eq:gen3}) is implemented in 'physical way' (similar to the case $k_2=0$ in Sec. \ref{sec:linCoupledModeZero}), the current mode is simpler than the potential mode, however is also less controllable. The current between electrodes at applied potential is calculated as impedance (in RMS form for AC signals). Sensing is performed by the last pair of dipoles and is shown in Figs. \ref{fig:collectiveEIS}. Size of dipole is related to dipole-dipole distance as 1:8, to the size of the whole group as 1:24, and to size of the sensing area as 1:40. Due to larger sensing area, this setup has a smaller dipole size and is also used to study electrochemical and optical effects in fluidic media \cite{kernbach2022electrochemical} and biological organisms \cite{WatchPlant21}.

\begin{figure}[ht]
\centering
\subfigure{\includegraphics[width=0.45\textwidth]{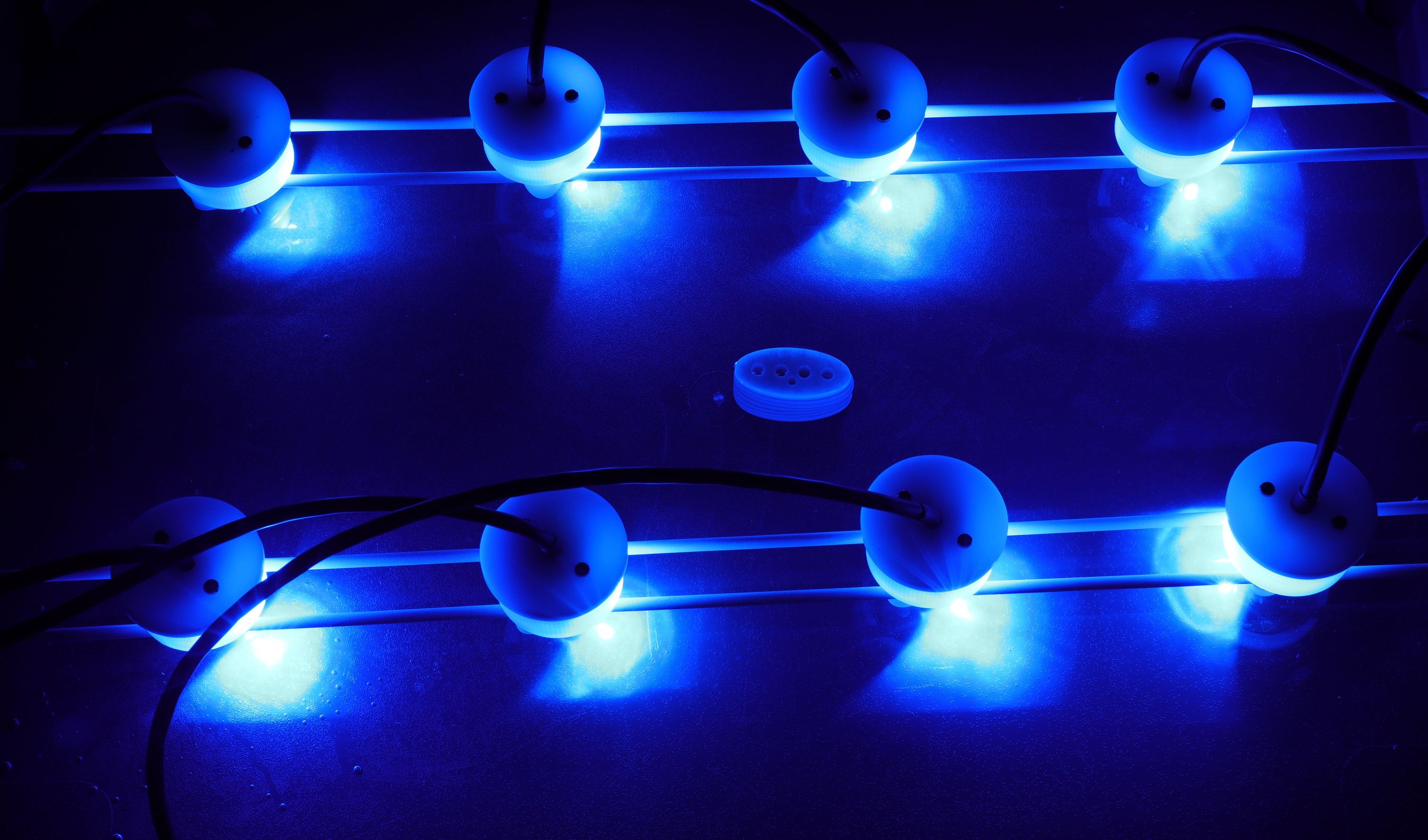}}
\caption{\small Experimental setup for current mode sensors with 8 dipoles. Dielectric object in a common field is shown. \label{fig:experimentEIS}}
\end{figure}

Since the configuration of common electric field is sensitive to positions of dipoles and geometry of field with included obstacles, this can be used for sensing purposes. Fig. \ref{fig:collectiveEIS2} demonstrates electrochemical dynamics measured in the last dipole in cases if a dielectric object or dipoles are moved in the sensing area. Influence of dipoles is larger and this allows separating signal levels for group-external objects and for group-internal dipoles.

\begin{figure}[ht]
\centering
\subfigure[\label{fig:collectiveEIS2}]{\includegraphics[width=0.49\textwidth]{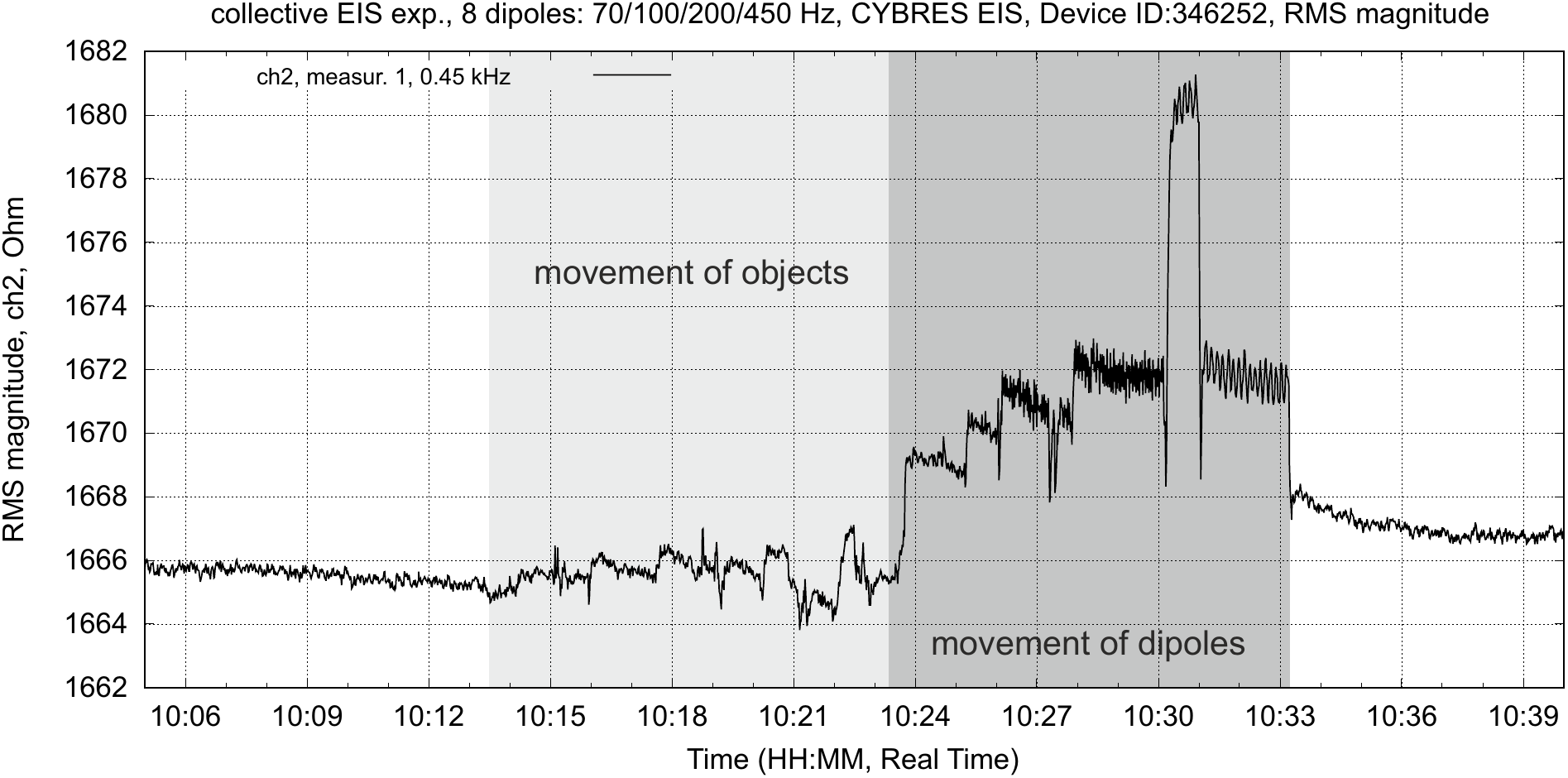}}
\subfigure[\label{fig:collectiveEIS1}]{\includegraphics[width=0.49\textwidth]{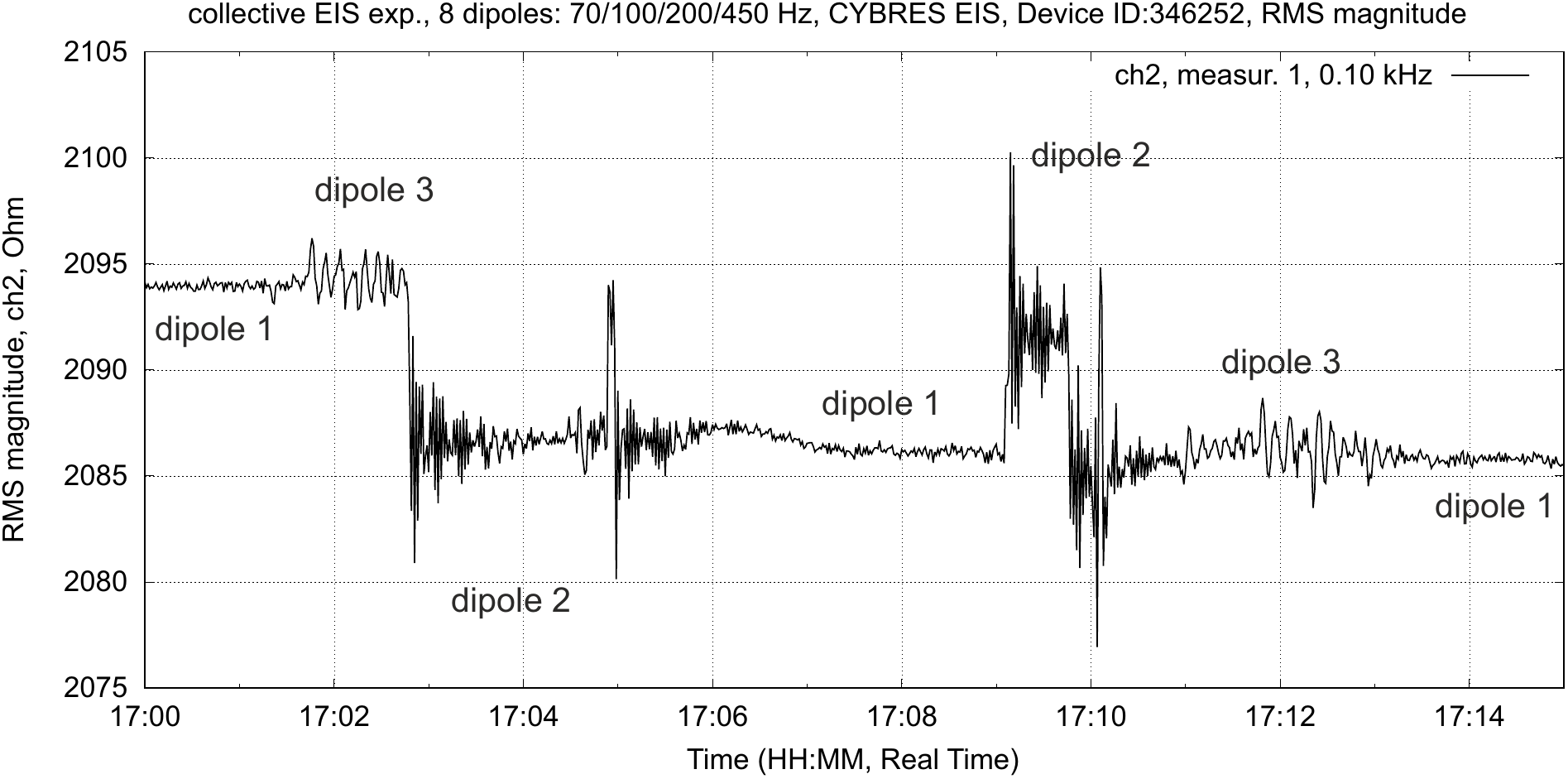}}
\caption{\small Electrochemical dynamics of a single dipole for \textbf{(a)} movement of dielectric objects and other dipoles in a sensing area of the group; \textbf{(b)} approaching other dipoles with different oscillating frequencies and appearance of interference patterns. \label{fig:collectiveEIS}}
\end{figure}

All dipoles oscillate at different frequencies and the interference pattern of electric field is specific to each of these dipoles. Fig. \ref{fig:collectiveEIS1} demonstrate examples of such patterns, detectable on the level of a single dipole. This approach can be used for different purposes, e.g. for identification of beacons and AUVs.

\section{Conclusion}
\label{sec:conclusion}

This work explores the behavior of nonlinear oscillators, coupled through an electric field in water. The $\pm5V$ driven hardware is capable of generating and direct sensing an electric field within 3x-4x body lengths (size of electric dipole -- the distance between electrodes). The coupled mode of different oscillating devices enables extending this distance up to 5x-6x in potential mode and $>$10x in current mode in fresh water. We have demonstrated five different schemes for potential and current sensing, where two effects have an impact on the behavior of nonlinear oscillators: phase desychronization of pulses and global synchronization of amplitudes. A combination of both effects enables either the sensing of distances or the number of emitting devices. Analysis of temporal patterns allows the devices to draw more complex conclusions about the spatial configurations of AUVs or dielectric objects in the sensing area. Moreover, temporal synchronization of amplitudes makes it possible to coordinate activities of AUVs without global communication.

Using one device as an electric mirror to reflect the received signals can change the qualitative behavior of the coupled system. The main argument for these experiments is that by comparing the self- and received signals, each AUV can recognize its membership in the group. This approach underlies more complex cognitive strategies of group-based identification and collective recognition of the 'self' and 'non-self'. Experiments demonstrated a good correlation with analytical results, this can be used for designing collective capabilities of larger underwater systems.

\section{Acknowledgement}

This work is partially supported by EU-H2020 Projects 'subCULTron: Submarine cultures perform long-term robotic exploration of unconventional environmental niches', grant No: 640967 and 'WATCHPLANT: Smart Biohybrid Phyto-Organisms for Environmental In Situ Monitoring', grant No: 101017899 funded by European Commission.

\small

\begin{thebibliography}{10}
\providecommand{\url}[1]{#1}
\csname url@samestyle\endcsname
\providecommand{\newblock}{\relax}
\providecommand{\bibinfo}[2]{#2}
\providecommand{\BIBentrySTDinterwordspacing}{\spaceskip=0pt\relax}
\providecommand{\BIBentryALTinterwordstretchfactor}{4}
\providecommand{\BIBentryALTinterwordspacing}{\spaceskip=\fontdimen2\font plus
\BIBentryALTinterwordstretchfactor\fontdimen3\font minus
  \fontdimen4\font\relax}
\providecommand{\BIBforeignlanguage}[2]{{%
\expandafter\ifx\csname l@#1\endcsname\relax
\typeout{** WARNING: IEEEtran.bst: No hyphenation pattern has been}%
\typeout{** loaded for the language `#1'. Using the pattern for}%
\typeout{** the default language instead.}%
\else
\language=\csname l@#1\endcsname
\fi
#2}}
\providecommand{\BIBdecl}{\relax}
\BIBdecl

\bibitem{TinyFish10}
K9keystrokes, ``Tiny robot fish may help save sea life from {BP} oil leak,''
  \emph{http://hubpages.com/hub/Tiny-Robot-Fish-may-help-save-sea-life-from-BP-Oil-Leak},
  2010.

\bibitem{KernbachDipperSutantyo11}
S.~Kernbach, T.~Dipper, and D.~Sutantyo, ``Multi-modal local sensing and
  communication for collective underwater systems,'' in \emph{Proceedings of
  the 11th International Conference on Mobile Robots and Competitions,
  {ROBOTICA11}}, 2011, pp. 96--101.

\bibitem{Bingham02}
D.~Bingham, T.~Drake, A.~Hill, and R.~Lott, ``The application of autonomous
  underwater vehicle ({AUV}) technology in the oil industry, vision and
  experiences,'' in \emph{FIG XXII International Congress Washington, D.C. USA,
  April 19-26 2002}, 2002, pp. 1--13.

\bibitem{Kernbach11-HCR}
S.~Kernbach, Ed., \emph{Handbook of Collective Robotics: Fundamentals and
  Challenges}.\hskip 1em plus 0.5em minus 0.4em\relax Singapore: Pan Stanford
  Publishing, 2012.

\bibitem{Emde98}
G.~von~der Emde, S.~Schwarz, L.~Gomez, R.~Budelli, and K.~Grant, ``Electric
  fish measure distance in the dark,'' \emph{Nature}, vol. 395, pp. 890--894,
  1998.

\bibitem{Sim01062011}
M.~Sim and D.~Kim, ``Electrolocation with an electric organ discharge waveform
  for biomimetic application,'' \emph{Adaptive Behavior}, vol.~19, no.~3, pp.
  172--186, 2011.

\bibitem{Boyer15}
F.~Boyer, V.~Lebastard, C.~Chevallereau, S.~Mintchev, and C.~Stefanini,
  ``Underwater navigation based on passive electric sense: New perspectives for
  underwater docking,'' \emph{The International Journal of Robotics Research},
  vol.~34, no.~9, pp. 1228--1250, 2015.

\bibitem{Chevallereau14}
C.~Christine, M.-R. Benachenhou, V.~Lebastard, and F.~Boyer, ``Electric
  sensor-based control of underwater robot groups,'' \emph{Robotics, IEEE
  Transactions on}, vol.~30, pp. 604--618, 06 2014.

\bibitem{Shang20}
\BIBentryALTinterwordspacing
W.~Shang, W.~Xue, Y.~Li, X.~Wu, and Y.~Xu, ``An improved underwater electric
  field-based target localization combining subspace scanning algorithm and
  meta-ep pso algorithm,'' \emph{Journal of Marine Science and Engineering},
  vol.~8, no.~4, 2020. [Online]. Available:
  \url{https://www.mdpi.com/2077-1312/8/4/232}
\BIBentrySTDinterwordspacing

\bibitem{Baffet08}
G.~Baffet, P.~Gossiaux, M.~Porez, and F.~Boyer, ``Underwater robotic:
  localization with electrolocation for collision avoidance,''
  \emph{http://hal.archives-ouvertes.fr/in2p3-00300570/}, 2008.

\bibitem{Kernbach17water}
S.~Kernbach, I.~Kuksin, and O.~Kernbach, ``On accurate differential
  measurements with electrochemical impedance spectroscopy,'' \emph{WATER},
  vol.~8, pp. 136--155, 2017.

\bibitem{ANGELS}
ANGELS, \emph{ANGuilliform robot with ELectric Sense, EU-project 231845,
  2009-2011}.\hskip 1em plus 0.5em minus 0.4em\relax European Communities,
  2011.

\bibitem{Thenius16subCULT}
R.~Thenius, D.~Moser, J.~C. Varughese, S.~Kernbach, I.~Kuksin, O.~Kernbach,
  E.~Kuksina, N.~Mi{\v{s}}kovi{\'{c}}, S.~Bogdan, T.~Petrovi{\'{c}},
  A.~Babi{\'{c}}, F.~Boyer, V.~Lebastard, S.~Bazeille, G.~W. Ferrari,
  E.~Donati, R.~Pelliccia, D.~Romano, G.~J. Van~Vuuren, C.~Stefanini,
  M.~Morgantin, A.~Campo, and T.~Schmickl, ``{subCULTron} - cultural
  development as a tool in underwater robotics,'' in \emph{Artificial Life and
  Intelligent Agents}, P.~R. Lewis, C.~J. Headleand, S.~Battle, and P.~D.
  Ritsos, Eds.\hskip 1em plus 0.5em minus 0.4em\relax Cham: Springer
  International Publishing, 2018, pp. 27--41.

\bibitem{CoCoRo}
CoCoRo, \emph{Collective Cognitive Robots, FP7}.\hskip 1em plus 0.5em minus
  0.4em\relax European Communities, 2011-2013.

\bibitem{Caputi11}
A.~A. Caputi and J.~Nogueira, ``Identifying self- and nonself- generated
  signals: Lessons from electrosensory systems,'' in \emph{Sensing Systems in
  Nature}, C.~L{\'o}pez-Larrea, Ed.\hskip 1em plus 0.5em minus 0.4em\relax
  Berlin, Heidelberg: Landes Bioscience and Springer, 2011, pp. 5--25.

\bibitem{AquaJelly}
FESTO, \emph{An artificial jellyfish with electric drive unit: an autonomously
  controlled jellyfish}.\hskip 1em plus 0.5em minus 0.4em\relax Festo AG \& Co.
  KG, 2008.

\bibitem{Atmanspachera05}
H.~Atmanspachera and H.~Scheingraber, ``Stabilization of causally and
  non-causally coupled map lattices,'' \emph{Physica A}, no. 345, pp. 435--447,
  2005.

\bibitem{Konishi99}
K.~Konishi and H.~Kokame, ``Decentralized delayed-feedback control of a one-way
  coupled ring lattice,'' \emph{Physica D}, vol. 127, pp. 1--12, 1999.

\bibitem{Chate92}
H.~Chat\'e and P.~Manneville, ``Emergence of effective low-dimensional dynamics
  in the macroscopic behavior of coupled map lattices,'' \emph{Europhysics
  Letters}, vol.~17, no.~4, pp. 291--296, 1992.

\bibitem{Maistrenko98}
Y.~Maistrenko, V.~Maistrenko, and A.~Popovich, ``Transverse instability and
  riddled basin in a system of two coupled logistic maps,'' \emph{Phys. Rev.
  E}, vol.~57, no.~3, pp. 2713--2724, 1998.

\bibitem{Levi99}
P.~Levi, M.~Schanz, S.~Kornienko, and O.~Kornienko, ``Application of order
  parameter equation for the analysis and the control of nonlinear time
  discrete dynamical systems,'' \emph{Int. J. Bifurcation and Chaos}, vol.~9,
  no.~8, pp. 1619--1634, 1999.

\bibitem{Kaneko93}
K.~Kaneko, \emph{Theory and application of coupled map lattices}.\hskip 1em
  plus 0.5em minus 0.4em\relax Chichester, New York, Brisbane, Toronto,
  Singapore: John Willey \& Sons., 1993.

\bibitem{Kernbach08}
S.~Kernbach, \emph{Structural Self-organization in Multi-Agents and
  Multi-Robotic Systems}.\hskip 1em plus 0.5em minus 0.4em\relax Logos Verlag,
  Berlin, 2008.

\bibitem{Kornienko_S04}
S.~Kornienko, O.~Kornienko, and P.~Levi, ``Multi-agent repairer of damaged
  process plans in manufacturing environment,'' in \emph{Proc. of the 8th Conf.
  on Intelligent Autonomous Systems (IAS-8), Amsterdam, NL}, 2004, pp.
  485--494.

\bibitem{Kornienko_OS01}
O.~Kornienko, S.~Kornienko, and P.~Levi, ``Collective decision making using
  natural self-organization in distributed systems,'' in \emph{Proc. of Int.
  Conf. on Computational Intelligence for Modelling, Control and Automation
  (CIMCA'2001), Las Vegas, USA}, 2001, pp. 460--471.

\bibitem{Kornienko_S03A}
S.~Kornienko, O.~Kornienko, and P.~Levi, ``Flexible manufacturing process
  planning based on the multi-agent technology,'' in \emph{Proc. of the 21st
  IASTED Int. Conf. on AI and Applications (AIA '2003), Innsbruck, Austria},
  2003, pp. 156--161.

\bibitem{Kornienko_S06b}
S.~Kernbach and O.~Kernbach, ``Collective energy homeostasis in a large-scale
  micro-robotic swarm,'' \emph{Robotics and Autonomous Systems, {DOI}
  10.1016/j.robot.2011.08.001}, vol.~59, pp. 1090--1101, 2011.

\bibitem{Christensen09}
A.~Christensen, R.~O'Grady, and M.~Dorigo, ``From fireflies to fault-tolerant
  swarms of robots,'' \emph{IEEE Transactions on Evolutionary Computation},
  vol.~13, no.~4, pp. 754 --766, aug. 2009.

\bibitem{Yuste:2005p45387}
R.~Yuste, J.~MacLean, J.~Smith, and A.~Lansner, ``The cortex as a central
  pattern generator,'' \emph{Nat Rev Neurosci}, vol.~6, no.~6, pp. 477--483,
  2005.

\bibitem{hamann2017flora}
H.~Hamann, M.~Soorati, M.~Heinrich, D.~Hofstadler, I.~Kuksin, F.~Veenstra,
  M.~Wahby, S.~Nielsen, S.~Risi, T.~Skrzypczak, P.~Zahadat, P.~Wojtaszek,
  K.~St{\o}y, T.~Schmickl, S.~Kernbach, and P.~Ayres, ``Flora robotica -- an
  architectural system combining living natural plants and distributed
  robots,'' 2017.

\bibitem{Endo08}
G.~Endo, J.~Morimoto, T.~Matsubara, J.~Nakanishi, and G.~Cheng, ``Learning
  cpg-based biped locomotion with a policy gradient method: Application to a
  humanoid robot,'' \emph{Int. J. Rob. Res.}, vol.~27, no.~2, pp. 213--228,
  2008.

\bibitem{Meister11}
E.~Meister, S.~Stepanenko, and S.~Kernbach, ``Adaptive locomotion of multibody
  snake-like robot,'' in \emph{International Conference on Multibody Dynamics,
  Brussels, Belgium}, 2011, pp. 1--8.

\bibitem{kernbach09-2adaptive-short}
S.~Kernbach, P.~Levi, E.~Meister, F.~Schlachter, and O.~Kernbach, ``Towards
  self-adaptation of robot organisms with a high developmental plasticity,'' in
  \emph{Proceedings of the 2009 Computation World: Future Computing, Service
  Computation, Cognitive, Adaptive, Content, Patterns}, ser. COMPUTATIONWORLD
  '09.\hskip 1em plus 0.5em minus 0.4em\relax Washington, DC, USA: IEEE
  Computer Society, 2009, pp. 180--187.

\bibitem{Nayfeh93}
A.~Nayfeh, \emph{Method of normal forms}.\hskip 1em plus 0.5em minus
  0.4em\relax New York: John Wiley \& Sohn, 1993.

\bibitem{Sandefur90}
J.~Sandefur, \emph{Discrete dynamical systems. Theory and Application}.\hskip
  1em plus 0.5em minus 0.4em\relax Calarendon Press, Oxford, 1990.

\bibitem{Alamir10}
M.~Alamir, O.~Omar, N.~Servagent, A.~Girin, P.~Bellemain, V.~Lebastard,
  P.~Gossiaux, F.~Boyer, and S.~Bouvier, ``On solving inverse problems for
  electric fish like robots,'' in \emph{Robotics and Biomimetics (ROBIO), 2010
  IEEE International Conference on}, dec. 2010, pp. 1081 --1086.

\bibitem{Wu98}
C.~W. Wu, ``Global synchronization in coupled map lattices,'' in \emph{Circuits
  and Systems, 1998. ISCAS '98. Proceedings of the 1998 IEEE International
  Symposium on}, vol.~3, may-3 jun 1998, pp. 302 --305 vol.3.

\bibitem{Lu07}
W.~Lu and T.~Chen, ``Global synchronization of discrete-time dynamical network
  with a directed graph,'' \emph{Circuits and Systems II: Express Briefs, IEEE
  Transactions on}, vol.~54, no.~2, pp. 136 --140, feb. 2007.

\bibitem{mehta:350}
M.~Mehta and S.~Sinha, ``Asynchronous updating of coupled maps leads to
  synchronization,'' \emph{Chaos: An Interdisciplinary Journal of Nonlinear
  Science}, vol.~10, no.~2, pp. 350--358, 2000.

\bibitem{Loncar19}
I.~Loncar, A.~Babic, B.~Arbanas~Ferreira, G.~Vasiljevic, T.~Petrovic,
  S.~Bogdan, and N.~Miskovic, ``A heterogeneous robotic swarm for long-term
  monitoring of marine environments,'' \emph{Applied Sciences}, vol.~9, p.
  1388, 04 2019.

\bibitem{kernbach2022electrochemical}
S.~Kernbach, ``Electrochemical characterisation of ionic dynamics resulting
  from spin conversion of water isomers,'' \emph{arXiv preprint
  arXiv:2202.00526}, 2022.

\bibitem{WatchPlant21}
H.~Hamann, S.~Bogdan, A.~Diaz-Espejo, L.~Garcia~Carmona, V.~Hernandez-Santana,
  S.~Kernbach, A.~Kernbach, A.~Quijano-Lopez, B.~Salamat, and M.~Wahby,
  ``Watchplant: Networked bio-hybrid systems for pollution monitoring of urban
  areas.''\hskip 1em plus 0.5em minus 0.4em\relax ALIFE 2021: The 2021
  Conference on Artificial Life, 2021.

\end{thebibliography}

\end{document}